\documentclass[twoside,11pt]{article}

\usepackage{hyperref}
\usepackage{jair, theapa, rawfonts}

\usepackage{graphicx}
\usepackage{amsmath}
\usepackage{amsthm}
\usepackage{booktabs}
\usepackage{algorithm}
\usepackage[noend]{algpseudocode}
\usepackage{enumitem}
\usepackage{stix2}

\usepackage{thmtools, thm-restate}

\newcommand{\citet}[1]
{\citeauthor{#1}~\shortcite{#1}}
\newcommand{\citep}{\cite}

\usepackage{textcomp}

\usepackage{tikz}
\usetikzlibrary{arrows}
\usetikzlibrary{calc}
\usetikzlibrary{shapes.geometric}
\usepackage{environ}

\usepackage{filecontents,catchfile}

\makeatletter
\pgfdeclareshape{datastore}{
 \inheritsavedanchors[from=rectangle]
 \inheritanchorborder[from=rectangle]
 \inheritanchor[from=rectangle]{center}
 \inheritanchor[from=rectangle]{base}
 \inheritanchor[from=rectangle]{north}
 \inheritanchor[from=rectangle]{north east}
 \inheritanchor[from=rectangle]{east}
 \inheritanchor[from=rectangle]{south east}
 \inheritanchor[from=rectangle]{south}
 \inheritanchor[from=rectangle]{south west}
 \inheritanchor[from=rectangle]{west}
 \inheritanchor[from=rectangle]{north west}
 \backgroundpath{
 \southwest \pgf@xa=\pgf@x \pgf@ya=\pgf@y
 \northeast \pgf@xb=\pgf@x \pgf@yb=\pgf@y
 \pgfpathmoveto{\pgfpoint{\pgf@xa}{\pgf@ya}}
 \pgfpathlineto{\pgfpoint{\pgf@xb}{\pgf@ya}}
 \pgfpathmoveto{\pgfpoint{\pgf@xa}{\pgf@yb}}
 \pgfpathlineto{\pgfpoint{\pgf@xb}{\pgf@yb}}
 }
}
\makeatother

\usepackage{amsmath,amsfonts,amsthm}
\usepackage{epsfig,epstopdf,url,array}

\usepackage[markup=underlined]{changes}
\definechangesauthor[color=red]{Kerstin}
\definechangesauthor[color=blue]{Anas}
\definechangesauthor[color= teal]{Derek}
\definechangesauthor[color=brown]{draft}

\usepackage{cleveref}
\usepackage{tcolorbox}
\usepackage{fancyvrb}
\newcommand\userinput[1]{\textbf{#1}}
\usepackage{tgcursor}

\algnewcommand\algorithmicforeach{\textbf{for each}}
\algdef{S}[FOR]{ForEach}[1]{\algorithmicforeach\ #1\ \algorithmicdo}

\theoremstyle{plain}
\newtheorem{thm}{Theorem}[section]

\newtheorem{task}[thm]{Search task}

\theoremstyle{definition}
\newtheorem{defn}{Definition}[section]

\theoremstyle{remark}

\jairheading{1}{1993}{1-15}{6/91}{9/91}
\ShortHeadings{D-VAL: planning domains functional equivalence validation tool}
{Shrinah, Long, \& Eder}
\firstpageno{1}

\newcommand{\PipesworldResults}{ 
\begin{table}[H] 
\begin{center} 
\resizebox{\textwidth}{!}{ 
\begin{tabular}{|p{0.40\linewidth}|M{0.1\linewidth}|M{0.05\linewidth}|M{0.08\linewidth}|M{0.13\linewidth}|M{0.08\linewidth}|M{0.08\linewidth}|M{0.1\linewidth}|} 
\hline 
\textbf{Domain version}  & \textbf{Expected impact} & \textbf{FE} & \textbf{Reason} &  \textbf{Simple domains} & \textbf{Total time (s)}  & \textbf{SMT time (s)}  & \textbf{Planning time (s)}    \\ \hline  
Pipesworld with crafted valid macro & Yes & FE & 12 & No & 16.77 & 16.57 & 0.2 \\  \hline
Pipesworld with random valid macro & Yes & FE & 12 & No & 17.39 & 17.24 & 0.15 \\  \hline
Pipesworld with random invalid macro & No & NCV & 17 & No & 0.21 & 0 & 0.21 \\  \hline
Pipesworld with swapped variables & No & NCV & 15 & No & 0.15 & 0 & 0.15 \\  \hline
Pipesworld with swapped atoms & No & NCV & 13 & No & 4.81 & 4.64 & 0.17 \\  \hline
Pipesworld with deleted operator & No & NCV & 16 & No & 0.17 & 0 & 0.16 \\  \hline
\end{tabular}} 
\caption{The results of validating the functional equivalence between the Pipesworld domain and its modified versions. These modifications are detailed in \Cref{tab:PipesworldModifications}. The description of the reported values in this table is available in the caption of  \Cref{tab:GripperResults}.} 
\label{tab:PipesworldResults} 
\end{center} 
\end{table}} 

\newcommand{\BlocksworldResults}{ 
\begin{table}[H] 
\begin{center} 
\resizebox{\textwidth}{!}{ 
\begin{tabular}{|p{0.40\linewidth}|M{0.1\linewidth}|M{0.05\linewidth}|M{0.08\linewidth}|M{0.13\linewidth}|M{0.08\linewidth}|M{0.08\linewidth}|M{0.1\linewidth}|} 
\hline 
\textbf{Domain version}  & \textbf{Expected impact} & \textbf{FE} & \textbf{Reason} &  \textbf{Simple domains} & \textbf{Total time (s)}  & \textbf{SMT time (s)}  & \textbf{Planning time (s)}    \\ \hline  
Blocksworld with crafted valid macro & Yes & FE & 1 & Yes & 0.31 & 0.19 & 0.11 \\  \hline
Blocksworld with random valid macro & Yes & FE & 1 & Yes & 0.29 & 0.19 & 0.1 \\  \hline
Blocksworld with random invalid macro & No & NFE & 6 & Yes & 0.1 & 0 & 0.1 \\  \hline
Blocksworld with swapped variables & No & NFE & 2 & Yes & 0.25 & 0.15 & 0.1 \\  \hline
Blocksworld with swapped atoms & No & NFE & 4 & Yes & 0.1 & 0 & 0.1 \\  \hline
Blocksworld with deleted operator & No & NFE & 8 & Yes & 0.1 & 0 & 0.1 \\  \hline
\end{tabular}} 
\caption{The results of validating the functional equivalence between the Blocksworld domain and its modified versions. These modifications are detailed in \Cref{tab:BlocksworldModifications}. The description of the reported values in this table is available in the caption of  \Cref{tab:GripperResults}.} 
\label{tab:BlocksworldResults} 
\end{center} 
\end{table}} 

\newcommand{\CavedivingResults}{ 
\begin{table}[H] 
\begin{center} 
\resizebox{\textwidth}{!}{ 
\begin{tabular}{|p{0.40\linewidth}|M{0.1\linewidth}|M{0.05\linewidth}|M{0.08\linewidth}|M{0.13\linewidth}|M{0.08\linewidth}|M{0.08\linewidth}|M{0.1\linewidth}|} 
\hline 
\textbf{Domain version}  & \textbf{Expected impact} & \textbf{FE} & \textbf{Reason} &  \textbf{Simple domains} & \textbf{Total time (s)}  & \textbf{SMT time (s)}  & \textbf{Planning time (s)}    \\ \hline  
Cave-diving with crafted valid macro & Yes & FE & 1 & Yes & 2.51 & 2.31 & 0.2 \\  \hline
Cave-diving with random valid macro & Yes & FE & 1 & Yes & 2.51 & 2.3 & 0.21 \\  \hline
Cave-diving with random invalid macro & No & NFE & 6 & Yes & 0.23 & 0 & 0.23 \\  \hline
Cave-diving with swapped variables & No & NFE & 2 & Yes & 2.29 & 2.1 & 0.18 \\  \hline
Cave-diving with swapped atoms & No & NFE & 2 & Yes & 1.75 & 1.54 & 0.21 \\  \hline
Cave-diving with deleted operator & No & NFE & 8 & Yes & 0.16 & 0 & 0.15 \\  \hline
\end{tabular}} 
\caption{The results of validating the functional equivalence between the Cave-diving domain and its modified versions. These modifications are detailed in \Cref{tab:CavedivingModifications}. The description of the reported values in this table is available in the caption of  \Cref{tab:GripperResults}.} 
\label{tab:CavedivingResults} 
\end{center} 
\end{table}} 

\newcommand{\ChildsnackResults}{ 
\begin{table}[H] 
\begin{center} 
\resizebox{\textwidth}{!}{ 
\begin{tabular}{|p{0.40\linewidth}|M{0.1\linewidth}|M{0.05\linewidth}|M{0.08\linewidth}|M{0.13\linewidth}|M{0.08\linewidth}|M{0.08\linewidth}|M{0.1\linewidth}|} 
\hline 
\textbf{Domain version}  & \textbf{Expected impact} & \textbf{FE} & \textbf{Reason} &  \textbf{Simple domains} & \textbf{Total time (s)}  & \textbf{SMT time (s)}  & \textbf{Planning time (s)}    \\ \hline  
Child-snack with crafted valid macro & Yes & FE & 12 & No & 1.04 & 0.87 & 0.17 \\  \hline
Child-snack with random valid macro & Yes & FE & 12 & No & 1.05 & 0.9 & 0.14 \\  \hline
Child-snack with random invalid macro & No & NCV & 17 & No & 0.14 & 0 & 0.14 \\  \hline
Child-snack with swapped atoms & No & NCV & 14 & No & 0.13 & 0 & 0.13 \\  \hline
Child-snack with deleted operator & No & NCV & 16 & No & 0.15 & 0 & 0.14 \\  \hline
\end{tabular}} 
\caption{The results of validating the functional equivalence between the Child-snack domain and its modified versions. These modifications are detailed in \Cref{tab:ChildsnackModifications}. The description of the reported values in this table is available in the caption of  \Cref{tab:GripperResults}.} 
\label{tab:ChildsnackResults} 
\end{center} 
\end{table}} 

\newcommand{\FloortileResults}{ 
\begin{table}[H] 
\begin{center} 
\resizebox{\textwidth}{!}{ 
\begin{tabular}{|p{0.40\linewidth}|M{0.1\linewidth}|M{0.05\linewidth}|M{0.08\linewidth}|M{0.13\linewidth}|M{0.08\linewidth}|M{0.08\linewidth}|M{0.1\linewidth}|} 
\hline 
\textbf{Domain version}  & \textbf{Expected impact} & \textbf{FE} & \textbf{Reason} &  \textbf{Simple domains} & \textbf{Total time (s)}  & \textbf{SMT time (s)}  & \textbf{Planning time (s)}    \\ \hline  
Floor-tile with crafted valid macro & Yes & FE & 12 & No & 2.58 & 2.38 & 0.19 \\  \hline
Floor-tile with random valid macro & Yes & FE & 12 & No & 2.52 & 2.32 & 0.2 \\  \hline
Floor-tile with random invalid macro & No & NCV & 17 & No & 0.2 & 0 & 0.2 \\  \hline
Floor-tile with swapped variables & No & NCV & 13 & No & 2.32 & 2.17 & 0.15 \\  \hline
Floor-tile with swapped atoms & No & NCV & 13 & No & 2.23 & 2.04 & 0.18 \\  \hline
Floor-tile with deleted operator & No & NCV & 16 & No & 0.17 & 0 & 0.17 \\  \hline
\end{tabular}} 
\caption{The results of validating the functional equivalence between the Floor-tile domain and its modified versions. These modifications are detailed in \Cref{tab:FloortileModifications}. The description of the reported values in this table is available in the caption of  \Cref{tab:GripperResults}.} 
\label{tab:FloortileResults} 
\end{center} 
\end{table}} 

\newcommand{\GripperResults}{ 
\begin{table}[H] 
\begin{center} 
\resizebox{\textwidth}{!}{ 
\begin{tabular}{|p{0.40\linewidth}|M{0.1\linewidth}|M{0.05\linewidth}|M{0.08\linewidth}|M{0.13\linewidth}|M{0.08\linewidth}|M{0.08\linewidth}|M{0.1\linewidth}|} 
\hline 
\textbf{Domain version}  & \textbf{Expected impact} & \textbf{FE} & \textbf{Reason} &  \textbf{Simple domains} & \textbf{Total time (s)}  & \textbf{SMT time (s)}  & \textbf{Planning time (s)}    \\ \hline  
Gripper with crafted valid macro & Yes & FE & 1 & Yes & 0.2 & 0.14 & 0.05 \\  \hline
Gripper with random valid macro & Yes & FE & 1 & Yes & 0.18 & 0.13 & 0.05 \\  \hline
Gripper with random invalid macro & No & NFE & 6 & Yes & 0.06 & 0 & 0.05 \\  \hline
Gripper with swapped atoms & No & NFE & 4 & Yes & 0.05 & 0 & 0.05 \\  \hline
Gripper with deleted operator & No & NFE & 8 & Yes & 0.04 & 0 & 0.04 \\  \hline
\end{tabular}} 
\caption{The results of validating the functional equivalence between the Gripper domain and its modified versions. Expected impact: ``yes'' means the introduced modification on the original Gripper domain is expected to produce a version that is functionally equivalent to the original domain. These modifications are detailed in \Cref{tab:GripperModifications}. The FE column reports the verdict of our tool on the functional equivalence of each validation task: FE: functionally equivalent; NFE: not functionally equivalent; and NCV: no conclusive verdict. The reason column provides the justification of the decision of our tool. The reported numbers can be decoded with the help of \Cref{tab:ReasonsDecoder}. Simple domains: ``Yes'' means both the original domain and its modified version are simple domains, ``No'' means both domains are not simple domains,~i.e. either one or both are complex domains. Total time is the time taken by our tool to complete the validation task; it equals the sum of SMT and Planning times. SMT time is the time taken by the Z3 solver to check the existence of a suitable mapping between the predicates of the original domain and its modified version. Planning time is the time taken by the FF planner to find macro operators in the original domain and its modified version and to check operators are primitive in the case of simple domains.} 
\label{tab:GripperResults} 
\end{center} 
\end{table}} 

\newcommand{\HikingResults}{ 
\begin{table}[H] 
\begin{center} 
\resizebox{\textwidth}{!}{ 
\begin{tabular}{|p{0.40\linewidth}|M{0.1\linewidth}|M{0.05\linewidth}|M{0.08\linewidth}|M{0.13\linewidth}|M{0.08\linewidth}|M{0.08\linewidth}|M{0.1\linewidth}|} 
\hline 
\textbf{Domain version}  & \textbf{Expected impact} & \textbf{FE} & \textbf{Reason} &  \textbf{Simple domains} & \textbf{Total time (s)}  & \textbf{SMT time (s)}  & \textbf{Planning time (s)}    \\ \hline  
Hiking with crafted valid macro & Yes & FE & 1 & Yes & 3.83 & 3.58 & 0.25 \\  \hline
Hiking with random valid macro & Yes & FE & 1 & Yes & 3.81 & 3.59 & 0.22 \\  \hline
Hiking with random invalid macro & No & NFE & 6 & Yes & 0.26 & 0 & 0.26 \\  \hline
Hiking with swapped variables & No & FE & 1 & Yes & 3.82 & 3.62 & 0.2 \\  \hline
Hiking with swapped atoms & No & NFE & 2 & Yes & 1.91 & 1.71 & 0.2 \\  \hline
Hiking with deleted operator & No & FE & 1 & Yes & 3.81 & 3.63 & 0.17 \\  \hline
\end{tabular}} 
\caption{The results of validating the functional equivalence between the Hiking domain and its modified versions. These modifications are detailed in \Cref{tab:HikingModifications}. The description of the reported values in this table is available in the caption of  \Cref{tab:GripperResults}.} 
\label{tab:HikingResults} 
\end{center} 
\end{table}} 

\newcommand{\LogisticsResults}{ 
\begin{table}[H] 
\begin{center} 
\resizebox{\textwidth}{!}{ 
\begin{tabular}{|p{0.40\linewidth}|M{0.1\linewidth}|M{0.05\linewidth}|M{0.08\linewidth}|M{0.13\linewidth}|M{0.08\linewidth}|M{0.08\linewidth}|M{0.1\linewidth}|} 
\hline 
\textbf{Domain version}  & \textbf{Expected impact} & \textbf{FE} & \textbf{Reason} &  \textbf{Simple domains} & \textbf{Total time (s)}  & \textbf{SMT time (s)}  & \textbf{Planning time (s)}    \\ \hline  
Logistics with crafted valid macro & Yes & FE & 1 & Yes & 4.82 & 4.51 & 0.31 \\  \hline
Logistics with random valid macro & Yes & FE & 1 & Yes & 4.95 & 4.58 & 0.36 \\  \hline
Logistics with random invalid macro & No & NFE & 6 & Yes & 0.36 & 0 & 0.36 \\  \hline
Logistics with swapped atoms & No & NFE & 2 & Yes & 3.97 & 3.55 & 0.42 \\  \hline
Logistics with deleted operator & No & NFE & 8 & Yes & 0.34 & 0 & 0.33 \\  \hline
\end{tabular}} 
\caption{The results of validating the functional equivalence between the Logistics domain and its modified versions. These modifications are detailed in \Cref{tab:LogisticsModifications}. The description of the reported values in this table is available in the caption of  \Cref{tab:GripperResults}.} 
\label{tab:LogisticsResults} 
\end{center} 
\end{table}} 

\newcommand{\ElevatorResults}{ 
\begin{table}[H] 
\begin{center} 
\resizebox{\textwidth}{!}{ 
\begin{tabular}{|p{0.40\linewidth}|M{0.1\linewidth}|M{0.05\linewidth}|M{0.08\linewidth}|M{0.13\linewidth}|M{0.08\linewidth}|M{0.08\linewidth}|M{0.1\linewidth}|} 
\hline 
\textbf{Domain version}  & \textbf{Expected impact} & \textbf{FE} & \textbf{Reason} &  \textbf{Simple domains} & \textbf{Total time (s)}  & \textbf{SMT time (s)}  & \textbf{Planning time (s)}    \\ \hline  
Elevator with crafted valid macro & Yes & FE & 12 & No & 0.31 & 0.21 & 0.09 \\  \hline
Elevator with random valid macro & Yes & FE & 12 & No & 0.31 & 0.21 & 0.1 \\  \hline
Elevator with random invalid macro & No & NCV & 17 & No & 0.13 & 0 & 0.12 \\  \hline
Elevator with swapped variables & No & NCV & 13 & No & 0.28 & 0.17 & 0.1 \\  \hline
Elevator with swapped atoms & No & NCV & 13 & No & 0.28 & 0.17 & 0.11 \\  \hline
Elevator with deleted operator & No & NCV & 16 & No & 0.08 & 0 & 0.08 \\  \hline
\end{tabular}} 
\caption{The results of validating the functional equivalence between the Elevator domain and its modified versions. These modifications are detailed in \Cref{tab:ElevatorModifications}. The description of the reported values in this table is available in the caption of  \Cref{tab:GripperResults}.} 
\label{tab:ElevatorResults} 
\end{center} 
\end{table}} 

\newcommand{\ParkingResults}{ 
\begin{table}[H] 
\begin{center} 
\resizebox{\textwidth}{!}{ 
\begin{tabular}{|p{0.40\linewidth}|M{0.1\linewidth}|M{0.05\linewidth}|M{0.08\linewidth}|M{0.13\linewidth}|M{0.08\linewidth}|M{0.08\linewidth}|M{0.1\linewidth}|} 
\hline 
\textbf{Domain version}  & \textbf{Expected impact} & \textbf{FE} & \textbf{Reason} &  \textbf{Simple domains} & \textbf{Total time (s)}  & \textbf{SMT time (s)}  & \textbf{Planning time (s)}    \\ \hline  
Parking with crafted valid macro & Yes & FE & 1 & Yes & 0.63 & 0.53 & 0.1 \\  \hline
Parking with random valid macro & Yes & FE & 1 & Yes & 0.65 & 0.53 & 0.12 \\  \hline
Parking with random invalid macro & No & NFE & 6 & Yes & 0.11 & 0 & 0.11 \\  \hline
Parking with swapped variables & No & NFE & 4 & Yes & 0.1 & 0 & 0.09 \\  \hline
Parking with swapped atoms & No & NFE & 4 & Yes & 0.11 & 0 & 0.11 \\  \hline
Parking with deleted operator & No & NFE & 8 & Yes & 0.09 & 0 & 0.09 \\  \hline
\end{tabular}} 
\caption{The results of validating the functional equivalence between the Parking domain and its modified versions. These modifications are detailed in \Cref{tab:ParkingModifications}. The description of the reported values in this table is available in the caption of  \Cref{tab:GripperResults}.} 
\label{tab:ParkingResults} 
\end{center} 
\end{table}} 

\newcommand{\RoverResults}{ 
\begin{table}[H] 
\begin{center} 
\resizebox{\textwidth}{!}{ 
\begin{tabular}{|p{0.40\linewidth}|M{0.1\linewidth}|M{0.05\linewidth}|M{0.08\linewidth}|M{0.13\linewidth}|M{0.08\linewidth}|M{0.08\linewidth}|M{0.1\linewidth}|} 
\hline 
\textbf{Domain version}  & \textbf{Expected impact} & \textbf{FE} & \textbf{Reason} &  \textbf{Simple domains} & \textbf{Total time (s)}  & \textbf{SMT time (s)}  & \textbf{Planning time (s)}    \\ \hline  
Rover with crafted valid macro & Yes & FE & 1 & Yes & 6.26 & 6.06 & 0.19 \\  \hline
Rover with random valid macro & Yes & FE & 1 & Yes & 6.37 & 6.13 & 0.23 \\  \hline
Rover with random invalid macro & No & NFE & 6 & Yes & 0.22 & 0 & 0.22 \\  \hline
Rover with swapped variables & No & NFE & 2 & Yes & 3.73 & 3.54 & 0.19 \\  \hline
Rover with swapped atoms & No & NFE & 2 & Yes & 3.77 & 3.53 & 0.23 \\  \hline
Rover with deleted operator & No & NFE & 8 & Yes & 0.2 & 0 & 0.19 \\  \hline
\end{tabular}} 
\caption{The results of validating the functional equivalence between the Rover domain and its modified versions. These modifications are detailed in \Cref{tab:RoverModifications}. The description of the reported values in this table is available in the caption of  \Cref{tab:GripperResults}.} 
\label{tab:RoverResults} 
\end{center} 
\end{table}} 

\newcommand{\FreecellResults}{ 
\begin{table}[H] 
\begin{center} 
\resizebox{\textwidth}{!}{ 
\begin{tabular}{|p{0.40\linewidth}|M{0.1\linewidth}|M{0.05\linewidth}|M{0.08\linewidth}|M{0.13\linewidth}|M{0.08\linewidth}|M{0.08\linewidth}|M{0.1\linewidth}|} 
\hline 
\textbf{Domain version}  & \textbf{Expected impact} & \textbf{FE} & \textbf{Reason} &  \textbf{Simple domains} & \textbf{Total time (s)}  & \textbf{SMT time (s)}  & \textbf{Planning time (s)}    \\ \hline  
Freecell with crafted valid macro & Yes & FE & 1 & Yes & 42.89 & 42.57 & 0.32 \\  \hline
Freecell with random valid macro & Yes & FE & 1 & Yes & 41.57 & 41.24 & 0.33 \\  \hline
Freecell with random invalid macro & No & NFE & 6 & Yes & 0.36 & 0 & 0.36 \\  \hline
Freecell with swapped variables & No & NFE & 2 & Yes & 5.72 & 5.37 & 0.35 \\  \hline
Freecell with swapped atoms & No & NFE & 2 & Yes & 5.75 & 5.4 & 0.35 \\  \hline
Freecell with deleted operator & No & NFE & 8 & Yes & 0.33 & 0 & 0.32 \\  \hline
\end{tabular}} 
\caption{The results of validating the functional equivalence between the Freecell domain and its modified versions. These modifications are detailed in \Cref{tab:FreecellModifications}. The description of the reported values in this table is available in the caption of  \Cref{tab:GripperResults}.} 
\label{tab:FreecellResults} 
\end{center} 
\end{table}} 

\newcommand{\ScanalyzerResults}{ 
\begin{table}[H] 
\begin{center} 
\resizebox{\textwidth}{!}{ 
\begin{tabular}{|p{0.40\linewidth}|M{0.1\linewidth}|M{0.05\linewidth}|M{0.08\linewidth}|M{0.13\linewidth}|M{0.08\linewidth}|M{0.08\linewidth}|M{0.1\linewidth}|} 
\hline 
\textbf{Domain version}  & \textbf{Expected impact} & \textbf{FE} & \textbf{Reason} &  \textbf{Simple domains} & \textbf{Total time (s)}  & \textbf{SMT time (s)}  & \textbf{Planning time (s)}    \\ \hline  
Scanalyzer with crafted valid macro & Yes & FE & 1 & Yes & 25.38 & 25.27 & 0.11 \\  \hline
Scanalyzer with random valid macro & Yes & FE & 1 & Yes & 20.92 & 20.77 & 0.15 \\  \hline
Scanalyzer with random invalid macro & No & NFE & 6 & Yes & 0.14 & 0 & 0.14 \\  \hline
Scanalyzer with swapped variables & No & FE & 1 & Yes & 24.65 & 24.55 & 0.1 \\  \hline
Scanalyzer with swapped atoms & No & NFE & 4 & Yes & 0.11 & 0 & 0.11 \\  \hline
Scanalyzer with deleted operator & No & NFE & 8 & Yes & 0.1 & 0 & 0.1 \\  \hline
\end{tabular}} 
\caption{The results of validating the functional equivalence between the Scanalyzer domain and its modified versions. These modifications are detailed in \Cref{tab:ScanalyzerModifications}. The description of the reported values in this table is available in the caption of  \Cref{tab:GripperResults}.} 
\label{tab:ScanalyzerResults} 
\end{center} 
\end{table}} 

\newcommand{\PipesworldModifications}{ 
\begin{table}[H] 
\begin{center} 
\resizebox{\textwidth}{!}{ 
\begin{tabular}{|M{0.2\linewidth}|M{0.1\linewidth}|M{0.70\linewidth}|} 
\hline 
\textbf{Domain version}  & \textbf{Expected impact} & \textbf{Modification description}     \\ \hline  
Pipesworld with crafted valid macro & Yes & The valid macro operator ``pop-start'' is handcrafted and added to this modified version.
 \\  \hline
Pipesworld with random valid macro & Yes & The valid macro operator ``push-start-push-end-pop-start'' is randomly created from the operators of the original domain and added to this modified version.
 \\  \hline
Pipesworld with random invalid macro & No & The invalid macro operator ``push-start-push-end-pop-start'' is added to this modified version. This invalid macro is produced from the valid macro that is explained in the previous entry of this table by swapping the add effect (pop-updating ?x1) with the precondition (last ?x2 ?x1) in the original valid macro.
 \\  \hline
Pipesworld with swapped variables & No & The variable ?last-batch-atom in the parameters of the delete effect (follow-M ?next-last-batch-atom ?last-batch-atom) in the operator ``push-end'' in the original domain is swapped with the variable ?next-last-batch-atom from the parameters of the same delete effect of the same operator in this modified version.
 \\  \hline
Pipesworld with swapped atoms & No & The atom (last ?batch-atom-in ?pipe) in the add effects of the operator ``pop-unitarypipe'' in the original domain is changed to a precondition in this modified version, and the atom (connect ?from-area ?to-area ?pipe ) from the preconditions of the same operator in the original domain is changed to an add effect in this modified version.
 \\  \hline
Pipesworld with deleted operator & No & The operator ``push-unitarypipe'' is removed from this modified domain, this operator exists in the original domain.
 \\  \hline
\end{tabular}}
\caption{The description of the modifications applied to the Pipesworld domain to produce its modified versions. Expected impact: ``yes'' means the introduced modification to the original Pipesworld domain is expected to produce a version that is functionally equivalent to the original domain. } 
\label{tab:PipesworldModifications} 
\end{center} 
\end{table}} 

\newcommand{\BlocksworldModifications}{ 
\begin{table}[H] 
\begin{center} 
\resizebox{\textwidth}{!}{ 
\begin{tabular}{|M{0.2\linewidth}|M{0.1\linewidth}|M{0.70\linewidth}|} 
\hline 
\textbf{Domain version}  & \textbf{Expected impact} & \textbf{Modification description}     \\ \hline  
Blocksworld with crafted valid macro & Yes & The valid macro operator ``pick-up-stack'' is handcrafted and added to this modified version.
 \\  \hline
Blocksworld with random valid macro & Yes & The valid macro operator ``pick-up-put-down-stack'' is randomly created from the operators of the original domain and added to this modified version.
 \\  \hline
Blocksworld with random invalid macro & No & The invalid macro operator ``pick-up-put-down-stack'' is added to this modified version. This invalid macro is produced from the valid macro that is explained in the previous entry of this table by swapping the add effect (clear ?x2) with the precondition (ontable-M ?x1 ) in the original valid macro.
 \\  \hline
Blocksworld with swapped variables & No & The variable ?x in the parameters of the add effect (on ?y ?x) in the operator ``stack'' in the original domain is swapped with the variable ?y from the parameters of the same add effect of the same operator in this modified version.
 \\  \hline
Blocksworld with swapped atoms & No & The atom (clear ?y) in the add effects of the operator ``unstack'' in the original domain is changed to a precondition in this modified version, and the atom (clear ?x) from the preconditions of the same operator in the original domain is changed to an add effect in this modified version.
 \\  \hline
Blocksworld with deleted operator & No & The operator ``unstack'' is removed from this modified domain, this operator exists in the original domain.
 \\  \hline
\end{tabular}}
\caption{The description of the modifications applied to the Blocksworld domain to produce its modified versions. Expected impact: ``yes'' means the introduced modification to the original Blocksworld domain is expected to produce a version that is functionally equivalent to the original domain. } 
\label{tab:BlocksworldModifications} 
\end{center} 
\end{table}} 

\newcommand{\CavedivingModifications}{ 
\begin{table}[H] 
\begin{center} 
\resizebox{\textwidth}{!}{ 
\begin{tabular}{|M{0.2\linewidth}|M{0.1\linewidth}|M{0.70\linewidth}|} 
\hline 
\textbf{Domain version}  & \textbf{Expected impact} & \textbf{Modification description}     \\ \hline  
Cave-diving with crafted valid macro & Yes & The valid macro operator ``swim-photograph'' is handcrafted and added to this modified version.
 \\  \hline
Cave-diving with random valid macro & Yes & The valid macro operator ``prepare-tank-enter-water-pickup-tank'' is randomly created from the operators of the original domain and added to this modified version.
 \\  \hline
Cave-diving with random invalid macro & No & The invalid macro operator ``prepare-tank-enter-water-pickup-tank'' is added to this modified version. This invalid macro is produced from the valid macro that is explained in the previous entry of this table by swapping the add effect (holding ?x1 ?x2) with the precondition (capacity ?x1 ?x5) in the original valid macro.
 \\  \hline
Cave-diving with swapped variables & No & The variable ?q1 in the parameters of the precondition (next-quantity-M ?q1 ?q2 ) in the operator ``drop-tank'' in the original domain is swapped with the variable ?q2 from the parameters of the same precondition of the same operator in this modified version.
 \\  \hline
Cave-diving with swapped atoms & No & The atom (at-diver ?d ?l) in the add effects of the operator ``enter-water'' in the original domain is changed to a precondition in this modified version, and the atom (cave-entrance ?l) from the preconditions of the same operator in the original domain is changed to an add effect in this modified version.
 \\  \hline
Cave-diving with deleted operator & No & The operator ``prepare-tank'' is removed from this modified domain, this operator exists in the original domain.
 \\  \hline
\end{tabular}}
\caption{The description of the modifications applied to the Cave-diving domain to produce its modified versions. Expected impact: ``yes'' means the introduced modification to the original Cave-diving domain is expected to produce a version that is functionally equivalent to the original domain. } 
\label{tab:CavedivingModifications} 
\end{center} 
\end{table}} 

\newcommand{\ChildsnackModifications}{ 
\begin{table}[H] 
\begin{center} 
\resizebox{\textwidth}{!}{ 
\begin{tabular}{|M{0.2\linewidth}|M{0.1\linewidth}|M{0.70\linewidth}|} 
\hline 
\textbf{Domain version}  & \textbf{Expected impact} & \textbf{Modification description}     \\ \hline  
Child-snack with crafted valid macro & Yes & The valid macro operator ``make-sandwich-put-on-tray'' is handcrafted and added to this modified version.
 \\  \hline
Child-snack with random valid macro & Yes & The valid macro operator ``make-sandwich-no-gluten-put-on-tray-serve-sandwich-no-gluten'' is randomly created from the operators of the original domain and added to this modified version.
 \\  \hline
Child-snack with random invalid macro & No & The invalid macro operator ``make-sandwich-no-gluten-put-on-tray-serve-sandwich-no-gluten'' is added to this modified version. This invalid macro is produced from the valid macro that is explained in the previous entry of this table by swapping the add effect (no-gluten-sandwich ?x3) with the precondition (allergic-gluten-M ?x7) in the original valid macro.
 \\  \hline
Child-snack with swapped atoms & No & The atom (at ?t ?p2) in the add effects of the operator ``move-tray'' in the original domain is changed to a precondition in this modified version, and the atom (at ?t ?p1) from the preconditions of the same operator in the original domain is changed to an add effect in this modified version.
 \\  \hline
Child-snack with deleted operator & No & The operator ``put-on-tray'' is removed from this modified domain, this operator exists in the original domain.
 \\  \hline
\end{tabular}}
\caption{The description of the modifications applied to the Child-snack domain to produce its modified versions. Expected impact: ``yes'' means the introduced modification to the original Child-snack domain is expected to produce a version that is functionally equivalent to the original domain. } 
\label{tab:ChildsnackModifications} 
\end{center} 
\end{table}} 

\newcommand{\FloortileModifications}{ 
\begin{table}[H] 
\begin{center} 
\resizebox{\textwidth}{!}{ 
\begin{tabular}{|M{0.2\linewidth}|M{0.1\linewidth}|M{0.70\linewidth}|} 
\hline 
\textbf{Domain version}  & \textbf{Expected impact} & \textbf{Modification description}     \\ \hline  
Floor-tile with crafted valid macro & Yes & The valid macro operator ``up-right'' is handcrafted and added to this modified version.
 \\  \hline
Floor-tile with random valid macro & Yes & The valid macro operator ``change-color-paint-up-paint-down'' is randomly created from the operators of the original domain and added to this modified version.
 \\  \hline
Floor-tile with random invalid macro & No & The invalid macro operator ``change-color-paint-up-paint-down'' is added to this modified version. This invalid macro is produced from the valid macro that is explained in the previous entry of this table by swapping the add effect (robot-has ?x1 ?x3) with the precondition (available-color ?x3) in the original valid macro.
 \\  \hline
Floor-tile with swapped variables & No & The variable ?x in the parameters of the precondition (right-M ?y ?x ) in the operator ``right'' in the original domain is swapped with the variable ?y from the parameters of the same precondition of the same operator in this modified version.
 \\  \hline
Floor-tile with swapped atoms & No & The atom (clear ?x) in the add effects of the operator ``down'' in the original domain is changed to a precondition in this modified version, and the atom (down ?y ?x) from the preconditions of the same operator in the original domain is changed to an add effect in this modified version.
 \\  \hline
Floor-tile with deleted operator & No & The operator ``right'' is removed from this modified domain, this operator exists in the original domain.
 \\  \hline
\end{tabular}}
\caption{The description of the modifications applied to the Floor-tile domain to produce its modified versions. Expected impact: ``yes'' means the introduced modification to the original Floor-tile domain is expected to produce a version that is functionally equivalent to the original domain. } 
\label{tab:FloortileModifications} 
\end{center} 
\end{table}} 

\newcommand{\GripperModifications}{ 
\begin{table}[H] 
\begin{center} 
\resizebox{\textwidth}{!}{ 
\begin{tabular}{|M{0.2\linewidth}|M{0.1\linewidth}|M{0.70\linewidth}|} 
\hline 
\textbf{Domain version}  & \textbf{Expected impact} & \textbf{Modification description}     \\ \hline  
Gripper with crafted valid macro & Yes & The valid macro operator ``pick-move-drop'' is handcrafted and added to this modified version.
 \\  \hline
Gripper with random valid macro & Yes & The valid macro operator ``move-pick-drop'' is randomly created from the operators of the original domain and added to this modified version.
 \\  \hline
Gripper with random invalid macro & No & The invalid macro operator ``move-pick-drop'' is added to this modified version. This invalid macro is produced from the valid macro that is explained in the previous entry of this table by swapping the add effect (at-robby ?x2) with the precondition (at ?x3 ?x2) in the original valid macro.
 \\  \hline
Gripper with swapped atoms & No & The atom (at-robby ?to) in the add effects of the operator ``move'' in the original domain is changed to a precondition in this modified version, and the atom (at-robby ?from) from the preconditions of the same operator in the original domain is changed to an add effect in this modified version.
 \\  \hline
Gripper with deleted operator & No & The operator ``drop'' is removed from this modified domain, this operator exists in the original domain.
 \\  \hline
\end{tabular}}
\caption{The description of the modifications applied to the Gripper domain to produce its modified versions. Expected impact: ``yes'' means the introduced modification to the original Gripper domain is expected to produce a version that is functionally equivalent to the original domain. } 
\label{tab:GripperModifications} 
\end{center} 
\end{table}} 

\newcommand{\HikingModifications}{ 
\begin{table}[H] 
\begin{center} 
\resizebox{\textwidth}{!}{ 
\begin{tabular}{|M{0.2\linewidth}|M{0.1\linewidth}|M{0.70\linewidth}|} 
\hline 
\textbf{Domain version}  & \textbf{Expected impact} & \textbf{Modification description}     \\ \hline  
Hiking with crafted valid macro & Yes & The valid macro operator ``drive-tent-put-up'' is handcrafted and added to this modified version.
 \\  \hline
Hiking with random valid macro & Yes & The valid macro operator ``put-down-put-up-drive-passenger'' is randomly created from the operators of the original domain and added to this modified version.
 \\  \hline
Hiking with random invalid macro & No & The invalid macro operator ``put-down-put-up-drive-passenger'' is added to this modified version. This invalid macro is produced from the valid macro that is explained in the previous entry of this table by swapping the add effect (at-person ?x1 ?x4) with the precondition (at-tent ?x5 ?x3) in the original valid macro.
 \\  \hline
Hiking with swapped variables & No & The variable ?x3 in the parameters of the precondition (partners-M ?x6 ?x3 ?x5 ) in the operator ``walk-together'' in the original domain is swapped with the variable ?x5 from the parameters of the same precondition of the same operator in this modified version.
 \\  \hline
Hiking with swapped atoms & No & The atom (up ?x3) in the add effects of the operator ``put-up'' in the original domain is changed to a precondition in this modified version, and the atom (at-tent ?x3 ?x2) from the preconditions of the same operator in the original domain is changed to an add effect in this modified version.
 \\  \hline
Hiking with deleted operator & No & The operator ``drive-tent-passenger'' is removed from this modified domain, this operator exists in the original domain.
 \\  \hline
\end{tabular}}
\caption{The description of the modifications applied to the Hiking domain to produce its modified versions. Expected impact: ``yes'' means the introduced modification to the original Hiking domain is expected to produce a version that is functionally equivalent to the original domain. } 
\label{tab:HikingModifications} 
\end{center} 
\end{table}} 

\newcommand{\LogisticsModifications}{ 
\begin{table}[H] 
\begin{center} 
\resizebox{\textwidth}{!}{ 
\begin{tabular}{|M{0.2\linewidth}|M{0.1\linewidth}|M{0.70\linewidth}|} 
\hline 
\textbf{Domain version}  & \textbf{Expected impact} & \textbf{Modification description}     \\ \hline  
Logistics with crafted valid macro & Yes & The valid macro operator ``load-fly-unload-airplane'' is handcrafted and added to this modified version.
 \\  \hline
Logistics with random valid macro & Yes & The valid macro operator ``load-truck-location-load-truck-airport-load-airplane'' is randomly created from the operators of the original domain and added to this modified version.
 \\  \hline
Logistics with random invalid macro & No & The invalid macro operator ``load-truck-location-load-truck-airport-load-airplane'' is added to this modified version. This invalid macro is produced from the valid macro that is explained in the previous entry of this table by swapping the add effect (in-truck ?x1 ?x4) with the precondition (at-package-location ?x1 ?x2) in the original valid macro.
 \\  \hline
Logistics with swapped atoms & No & The atom (at-truck-location ?truck ?loc-to) in the add effects of the operator ``drive-truck-airport-location'' in the original domain is changed to a precondition in this modified version, and the atom (airport-in-city ?loc-from ?city) from the preconditions of the same operator in the original domain is changed to an add effect in this modified version.
 \\  \hline
Logistics with deleted operator & No & The operator ``fly-airplane'' is removed from this modified domain, this operator exists in the original domain.
 \\  \hline
\end{tabular}}
\caption{The description of the modifications applied to the Logistics domain to produce its modified versions. Expected impact: ``yes'' means the introduced modification to the original Logistics domain is expected to produce a version that is functionally equivalent to the original domain. } 
\label{tab:LogisticsModifications} 
\end{center} 
\end{table}} 

\newcommand{\ElevatorModifications}{ 
\begin{table}[H] 
\begin{center} 
\resizebox{\textwidth}{!}{ 
\begin{tabular}{|M{0.2\linewidth}|M{0.1\linewidth}|M{0.70\linewidth}|} 
\hline 
\textbf{Domain version}  & \textbf{Expected impact} & \textbf{Modification description}     \\ \hline  
Elevator with crafted valid macro & Yes & The valid macro operator ``board-up-depart'' is handcrafted and added to this modified version.
 \\  \hline
Elevator with random valid macro & Yes & The valid macro operator ``board-depart-up'' is randomly created from the operators of the original domain and added to this modified version.
 \\  \hline
Elevator with random invalid macro & No & The invalid macro operator ``board-depart-up'' is added to this modified version. This invalid macro is produced from the valid macro that is explained in the previous entry of this table by swapping the add effect (served ?x3) with the precondition (origin ?x3 ?x1) in the original valid macro.
 \\  \hline
Elevator with swapped variables & No & The variable ?f1 in the parameters of the add effect (above ?f1 ?f2) in the operator ``down'' in the original domain is swapped with the variable ?f2 from the parameters of the same add effect of the same operator in this modified version.
 \\  \hline
Elevator with swapped atoms & No & The atom (lift-at ?f2) in the add effects of the operator ``down'' in the original domain is changed to a precondition in this modified version, and the atom (above ?f2 ?f1) from the preconditions of the same operator in the original domain is changed to an add effect in this modified version.
 \\  \hline
Elevator with deleted operator & No & The operator ``up'' is removed from this modified domain, this operator exists in the original domain.
 \\  \hline
\end{tabular}}
\caption{The description of the modifications applied to the Elevator domain to produce its modified versions. Expected impact: ``yes'' means the introduced modification to the original Elevator domain is expected to produce a version that is functionally equivalent to the original domain. } 
\label{tab:ElevatorModifications} 
\end{center} 
\end{table}} 

\newcommand{\ParkingModifications}{ 
\begin{table}[H] 
\begin{center} 
\resizebox{\textwidth}{!}{ 
\begin{tabular}{|M{0.2\linewidth}|M{0.1\linewidth}|M{0.70\linewidth}|} 
\hline 
\textbf{Domain version}  & \textbf{Expected impact} & \textbf{Modification description}     \\ \hline  
Parking with crafted valid macro & Yes & The valid macro operator ``move-car-to-curb-move-car-to-car'' is handcrafted and added to this modified version.
 \\  \hline
Parking with random valid macro & Yes & The valid macro operator ``move-curb-to-curb-move-curb-to-car-move-car-to-curb'' is randomly created from the operators of the original domain and added to this modified version.
 \\  \hline
Parking with random invalid macro & No & The invalid macro operator ``move-curb-to-curb-move-curb-to-car-move-car-to-curb'' is added to this modified version. This invalid macro is produced from the valid macro that is explained in the previous entry of this table by swapping the add effect (at-curb ?x1) with the precondition (at-curb-num ?x1 ?x4) in the original valid macro.
 \\  \hline
Parking with swapped variables & No & The variable ?car in the parameters of the add effect (behind-car ?cardest ?car) in the operator ``move-curb-to-car'' in the original domain is swapped with the variable ?cardest from the parameters of the same add effect of the same operator in this modified version.
 \\  \hline
Parking with swapped atoms & No & The atom (at-curb ?car) in the add effects of the operator ``move-car-to-curb'' in the original domain is changed to a precondition in this modified version, and the atom (behind-car ?car ?carsrc) from the preconditions of the same operator in the original domain is changed to an add effect in this modified version.
 \\  \hline
Parking with deleted operator & No & The operator ``move-curb-to-car'' is removed from this modified domain, this operator exists in the original domain.
 \\  \hline
\end{tabular}}
\caption{The description of the modifications applied to the Parking domain to produce its modified versions. Expected impact: ``yes'' means the introduced modification to the original Parking domain is expected to produce a version that is functionally equivalent to the original domain. } 
\label{tab:ParkingModifications} 
\end{center} 
\end{table}} 

\newcommand{\RoverModifications}{ 
\begin{table}[H] 
\begin{center} 
\resizebox{\textwidth}{!}{ 
\begin{tabular}{|M{0.2\linewidth}|M{0.1\linewidth}|M{0.70\linewidth}|} 
\hline 
\textbf{Domain version}  & \textbf{Expected impact} & \textbf{Modification description}     \\ \hline  
Rover with crafted valid macro & Yes & The valid macro operator ``calibrate-take-image'' is handcrafted and added to this modified version.
 \\  \hline
Rover with random valid macro & Yes & The valid macro operator ``navigate-sample-soil-drop'' is randomly created from the operators of the original domain and added to this modified version.
 \\  \hline
Rover with random invalid macro & No & The invalid macro operator ``navigate-sample-soil-drop'' is added to this modified version. This invalid macro is produced from the valid macro that is explained in the previous entry of this table by swapping the add effect (have-soil-analysis ?x1 ?x3) with the precondition (available ?x1) in the original valid macro.
 \\  \hline
Rover with swapped variables & No & The variable ?y in the parameters of the precondition (can-traverse ?x ?y ?z ) in the operator ``navigate'' in the original domain is swapped with the variable ?z from the parameters of the same precondition of the same operator in this modified version.
 \\  \hline
Rover with swapped atoms & No & The atom (calibrated ?i ?r) in the add effects of the operator ``calibrate'' in the original domain is changed to a precondition in this modified version, and the atom (at ?r ?w) from the preconditions of the same operator in the original domain is changed to an add effect in this modified version.
 \\  \hline
Rover with deleted operator & No & The operator ``communicate-rock-data'' is removed from this modified domain, this operator exists in the original domain.
 \\  \hline
\end{tabular}}
\caption{The description of the modifications applied to the Rover domain to produce its modified versions. Expected impact: ``yes'' means the introduced modification to the original Rover domain is expected to produce a version that is functionally equivalent to the original domain. } 
\label{tab:RoverModifications} 
\end{center} 
\end{table}} 

\newcommand{\FreecellModifications}{ 
\begin{table}[H] 
\begin{center} 
\resizebox{\textwidth}{!}{ 
\begin{tabular}{|M{0.2\linewidth}|M{0.1\linewidth}|M{0.70\linewidth}|} 
\hline 
\textbf{Domain version}  & \textbf{Expected impact} & \textbf{Modification description}     \\ \hline  
Freecell with crafted valid macro & Yes & The valid macro operator ``move-sendtohome'' is handcrafted and added to this modified version.
 \\  \hline
Freecell with random valid macro & Yes & The valid macro operator ``move-move-b-sendtofree'' is randomly created from the operators of the original domain and added to this modified version.
 \\  \hline
Freecell with random invalid macro & No & The invalid macro operator ``move-move-b-sendtofree'' is added to this modified version. This invalid macro is produced from the valid macro that is explained in the previous entry of this table by swapping the add effect (cellspace ?x5) with the precondition (canstack ?x3 ?x2) in the original valid macro.
 \\  \hline
Freecell with swapped variables & No & The variable ?cols in the parameters of the precondition (successor ?cols ?ncols) in the operator ``sendtohome-b'' in the original domain is swapped with the variable ?ncols from the parameters of the same precondition of the same operator in this modified version.
 \\  \hline
Freecell with swapped atoms & No & The atom (cellspace ?ncells) in the add effects of the operator ``colfromfreecell'' in the original domain is changed to a precondition in this modified version, and the atom (successor ?ncells ?cells) from the preconditions of the same operator in the original domain is changed to an add effect in this modified version.
 \\  \hline
Freecell with deleted operator & No & The operator ``sendtofree'' is removed from this modified domain, this operator exists in the original domain.
 \\  \hline
\end{tabular}}
\caption{The description of the modifications applied to the Freecell domain to produce its modified versions. Expected impact: ``yes'' means the introduced modification to the original Freecell domain is expected to produce a version that is functionally equivalent to the original domain. } 
\label{tab:FreecellModifications} 
\end{center} 
\end{table}} 

\newcommand{\ScanalyzerModifications}{ 
\begin{table}[H] 
\begin{center} 
\resizebox{\textwidth}{!}{ 
\begin{tabular}{|M{0.2\linewidth}|M{0.1\linewidth}|M{0.70\linewidth}|} 
\hline 
\textbf{Domain version}  & \textbf{Expected impact} & \textbf{Modification description}     \\ \hline  
Scanalyzer with crafted valid macro & Yes & The valid macro operator ``analyze-2-rotate-2'' is handcrafted and added to this modified version.
 \\  \hline
Scanalyzer with random valid macro & Yes & The valid macro operator ``analyze-2-analyze-4-rotate-2'' is randomly created from the operators of the original domain and added to this modified version.
 \\  \hline
Scanalyzer with random invalid macro & No & The invalid macro operator ``analyze-2-analyze-4-rotate-2'' is added to this modified version. This invalid macro is produced from the valid macro that is explained in the previous entry of this table by swapping the add effect (analyzed ?x5) with the precondition (on ?x6 ?x2) in the original valid macro.
 \\  \hline
Scanalyzer with swapped variables & No & The variable ?s1 in the parameters of the precondition (cycle-2 ?s1 ?s2 ) in the operator ``rotate-2'' in the original domain is swapped with the variable ?s2 from the parameters of the same precondition of the same operator in this modified version.
 \\  \hline
Scanalyzer with swapped atoms & No & The atom (on ?c4 ?s3) in the add effects of the operator ``analyze-4'' in the original domain is changed to a precondition in this modified version, and the atom (on ?c2 ?s2) from the preconditions of the same operator in the original domain is changed to an add effect in this modified version.
 \\  \hline
Scanalyzer with deleted operator & No & The operator ``analyze-2'' is removed from this modified domain, this operator exists in the original domain.
 \\  \hline
\end{tabular}}
\caption{The description of the modifications applied to the Scanalyzer domain to produce its modified versions. Expected impact: ``yes'' means the introduced modification to the original Scanalyzer domain is expected to produce a version that is functionally equivalent to the original domain. } 
\label{tab:ScanalyzerModifications} 
\end{center} 
\end{table}} 

\newcommand{\MysteryVariableMapping}{ 
\begin{table}[H] 
\begin{center} 
\resizebox{\textwidth}{!}{ 
\begin{tabular}{|l|l|}
\hline
\multicolumn{1}{|c|}{\textbf{Transportation variable signatures}} & \multicolumn{1}{c|}{\textbf{Mystery variable signatures}} \\ \hline
?v 2 has-space load ?v?s1 1                                       & ?v 2 harmony overcome ?v?s2 1                             \\ \hline
?v 2 has-space unload ?v?s2 1                                     & ?v 2 harmony succumb ?v?s2 1                              \\ \hline
?v 2 at\_v move ?v?l1 1                                           & ?v 2 craves\_v feast ?v?n1 1                              \\ \hline
?v 2 has-space unload ?v?s1 1                                     & ?v 2 harmony succumb ?v?s1 1                              \\ \hline
?v 2 in load ?c?v 2                                               & ?v 2 fears overcome ?c?v 2                                \\ \hline
?v 2 in unload ?c?v 2                                             & ?v 2 fears succumb ?c?v 2                                 \\ \hline
?v 2 at\_v unload ?v?l 1                                          & ?v 2 craves\_v succumb ?v?n 1                             \\ \hline
?v 2 has-space load ?v?s2 1                                       & ?v 2 harmony overcome ?v?s1 1                             \\ \hline
?v 2 at\_v load ?v?l 1                                            & ?v 2 craves\_v overcome ?v?n 1                            \\ \hline
?v 2 at\_v move ?v?l2 1                                           & ?v 2 craves\_v feast ?v?n2 1                              \\ \hline
?c 2 in load ?c?v 1                                               & ?c 2 fears overcome ?c?v 1                                \\ \hline
?c 2 in unload ?c?v 1                                             & ?c 2 fears succumb ?c?v 1                                 \\ \hline
?c 2 at\_c load ?c?l 1                                            & ?c 2 craves\_c overcome ?c?n 1                            \\ \hline
?c 2 at\_c unload ?c?l 1                                          & ?c 2 craves\_c succumb ?c?n 1                             \\ \hline
?l 2 at\_c unload ?c?l 2                                          & ?n 2 craves\_c succumb ?c?n 2                             \\ \hline
?l 2 at\_v unload ?v?l 2                                          & ?n 2 craves\_v succumb ?v?n 2                             \\ \hline
?l 2 at\_c load ?c?l 2                                            & ?n 2 craves\_c overcome ?c?n 2                            \\ \hline
?l 2 at\_v load ?v?l 2                                            & ?n 2 craves\_v overcome ?v?n 2                            \\ \hline
?l1 2 conn move ?l1?l2 1                                          & ?n1 2 eats feast ?n1?n2 1                                 \\ \hline
?l1 2 at\_v move ?v?l1 2                                          & ?n1 2 craves\_v feast ?v?n1 2                             \\ \hline
?l1 2 has-fuel move ?l1?f1 1                                      & ?n1 2 locale feast ?n1?l2 1                               \\ \hline
?l1 2 has-fuel move ?l1?f2 1                                      & ?n1 2 locale feast ?n1?l1 1                               \\ \hline
?s1 2 space-neighbor load ?s2?s1 2                                & ?s2 2 orbits overcome ?s1?s2 2                            \\ \hline
?s1 2 space-neighbor unload ?s1?s2 1                              & ?s1 2 orbits succumb ?s1?s2 1                             \\ \hline
?s1 2 has-space unload ?v?s1 2                                    & ?s1 2 harmony succumb ?v?s1 2                             \\ \hline
?s1 2 has-space load ?v?s1 2                                      & ?s2 2 harmony overcome ?v?s2 2                            \\ \hline
?s2 2 space-neighbor load ?s2?s1 1                                & ?s1 2 orbits overcome ?s1?s2 1                            \\ \hline
?s2 2 has-space load ?v?s2 2                                      & ?s1 2 harmony overcome ?v?s1 2                            \\ \hline
?s2 2 has-space unload ?v?s2 2                                    & ?s2 2 harmony succumb ?v?s2 2                             \\ \hline
?s2 2 space-neighbor unload ?s1?s2 2                              & ?s2 2 orbits succumb ?s1?s2 2                             \\ \hline
?f1 2 has-fuel move ?l1?f1 2                                      & ?l2 2 locale feast ?n1?l2 2                               \\ \hline
?f1 2 fuel-neighbor move ?f2?f1 2                                 & ?l2 2 attacks feast ?l1?l2 2                              \\ \hline
?f2 2 fuel-neighbor move ?f2?f1 1                                 & ?l1 2 attacks feast ?l1?l2 1                              \\ \hline
?f2 2 has-fuel move ?l1?f2 2                                      & ?l1 2 locale feast ?n1?l1 2                               \\ \hline
?l2 2 conn move ?l1?l2 2                                          & ?n2 2 eats feast ?n1?n2 2                                 \\ \hline
?l2 2 at\_v move ?v?l2 2                                          & ?n2 2 craves\_v feast ?v?n2 2                             \\ \hline
\end{tabular}}
\caption{The mapping from the signatures of the variables of the Transportation domain to the signatures of the variables of the Mystery domain.} 
\label{tab:MysteryVariableMapping} 
\end{center} 
\end{table}} 

\newcommand{\MysteryPredicateMapping}{
\begin{figure}

\begin{center}

\tikzset{every picture/.style={line width=0.75pt}} 

\begin{tikzpicture}[x=0.75pt,y=0.75pt,yscale=-1,xscale=1]

\draw   (284.5,145.6) .. controls (284.5,88.94) and (317.74,43) .. (358.75,43) .. controls (399.76,43) and (433,88.94) .. (433,145.6) .. controls (433,202.26) and (399.76,248.2) .. (358.75,248.2) .. controls (317.74,248.2) and (284.5,202.26) .. (284.5,145.6) -- cycle ;
\draw    (150.4,77.5) -- (338.18,77.12) ;
\draw [shift={(340.18,77.11)}, rotate = 179.88] [color={rgb, 255:red, 0; green, 0; blue, 0 }  ][line width=0.75]    (10.93,-3.29) .. controls (6.95,-1.4) and (3.31,-0.3) .. (0,0) .. controls (3.31,0.3) and (6.95,1.4) .. (10.93,3.29)   ;

\draw    (150.56,97.34) -- (338.11,96.9) ;
\draw [shift={(340.11,96.9)}, rotate = 179.87] [color={rgb, 255:red, 0; green, 0; blue, 0 }  ][line width=0.75]    (10.93,-3.29) .. controls (6.95,-1.4) and (3.31,-0.3) .. (0,0) .. controls (3.31,0.3) and (6.95,1.4) .. (10.93,3.29)   ;

\draw    (151.22,117.61) -- (339.89,116.51) ;
\draw [shift={(341.89,116.5)}, rotate = 179.67] [color={rgb, 255:red, 0; green, 0; blue, 0 }  ][line width=0.75]    (10.93,-3.29) .. controls (6.95,-1.4) and (3.31,-0.3) .. (0,0) .. controls (3.31,0.3) and (6.95,1.4) .. (10.93,3.29)   ;

\draw    (150.33,136.5) -- (338.78,136.72) ;
\draw [shift={(340.78,136.72)}, rotate = 180.07] [color={rgb, 255:red, 0; green, 0; blue, 0 }  ][line width=0.75]    (10.93,-3.29) .. controls (6.95,-1.4) and (3.31,-0.3) .. (0,0) .. controls (3.31,0.3) and (6.95,1.4) .. (10.93,3.29)   ;

\draw    (150.33,156.94) -- (339.44,156.5) ;
\draw [shift={(341.44,156.5)}, rotate = 179.87] [color={rgb, 255:red, 0; green, 0; blue, 0 }  ][line width=0.75]    (10.93,-3.29) .. controls (6.95,-1.4) and (3.31,-0.3) .. (0,0) .. controls (3.31,0.3) and (6.95,1.4) .. (10.93,3.29)   ;

\draw    (151.22,176.72) -- (338.11,176.5) ;
\draw [shift={(340.11,176.5)}, rotate = 179.93] [color={rgb, 255:red, 0; green, 0; blue, 0 }  ][line width=0.75]    (10.93,-3.29) .. controls (6.95,-1.4) and (3.31,-0.3) .. (0,0) .. controls (3.31,0.3) and (6.95,1.4) .. (10.93,3.29)   ;

\draw    (149.89,196.72) -- (338.56,196.5) ;
\draw [shift={(340.56,196.5)}, rotate = 179.93] [color={rgb, 255:red, 0; green, 0; blue, 0 }  ][line width=0.75]    (10.93,-3.29) .. controls (6.95,-1.4) and (3.31,-0.3) .. (0,0) .. controls (3.31,0.3) and (6.95,1.4) .. (10.93,3.29)   ;

\draw    (150.11,218.06) -- (337.89,216.52) ;
\draw [shift={(339.89,216.5)}, rotate = 179.53] [color={rgb, 255:red, 0; green, 0; blue, 0 }  ][line width=0.75]    (10.93,-3.29) .. controls (6.95,-1.4) and (3.31,-0.3) .. (0,0) .. controls (3.31,0.3) and (6.95,1.4) .. (10.93,3.29)   ;

\draw   (37.7,145.6) .. controls (37.7,88.94) and (70.94,43) .. (111.95,43) .. controls (152.96,43) and (186.2,88.94) .. (186.2,145.6) .. controls (186.2,202.26) and (152.96,248.2) .. (111.95,248.2) .. controls (70.94,248.2) and (37.7,202.26) .. (37.7,145.6) -- cycle ;

\draw (31.3,257) node [anchor=north west][inner sep=0.75pt]   [align=left] {Trasportation predicates};
\draw (295.95,257) node [anchor=north west][inner sep=0.75pt]   [align=left] {Mystery predicates};
\draw (343.86,76.5) node [anchor=west] [inner sep=0.75pt]   [align=left] {harmony};
\draw (148.4,77.5) node [anchor=east] [inner sep=0.75pt]   [align=left] {has-space};
\draw (342.11,96.5) node [anchor=west] [inner sep=0.75pt]   [align=left] {craves\_v};
\draw (148.56,96.94) node [anchor=east] [inner sep=0.75pt]   [align=left] { at\_v};
\draw (343.89,116.5) node [anchor=west] [inner sep=0.75pt]   [align=left] {craves\_c};
\draw (149.22,117.61) node [anchor=east] [inner sep=0.75pt]   [align=left] {at\_c};
\draw (342.78,136.72) node [anchor=west] [inner sep=0.75pt]   [align=left] {orbits};
\draw (148.33,136.5) node [anchor=east] [inner sep=0.75pt]   [align=left] {space-neighbor};
\draw (343.44,156.5) node [anchor=west] [inner sep=0.75pt]   [align=left] {attacks};
\draw (148.33,156.94) node [anchor=east] [inner sep=0.75pt]   [align=left] {fuel-neighbor};
\draw (342.11,176.5) node [anchor=west] [inner sep=0.75pt]   [align=left] {locale};
\draw (149.22,176.72) node [anchor=east] [inner sep=0.75pt]   [align=left] {has-fuel};
\draw (342.56,196.5) node [anchor=west] [inner sep=0.75pt]   [align=left] {eates};
\draw (147.89,196.72) node [anchor=east] [inner sep=0.75pt]   [align=left] {conn};
\draw (341.89,216.5) node [anchor=west] [inner sep=0.75pt]   [align=left] {fears};
\draw (148.11,218.06) node [anchor=east] [inner sep=0.75pt]   [align=left] {in};

\end{tikzpicture}
\end{center}
\caption{The mapping from the predicates of the Transportation domain to the predicates of the Mystery domain that makes the reach set of the former equal to the reach set of the latter.}
\label{fig:MysteryPredicateMapping}
\end{figure}}

\newcommand{\MysteryOperatorMapping}{
\begin{figure}

\begin{center}

\tikzset{every picture/.style={line width=0.75pt}} 

\begin{tikzpicture}[x=0.75pt,y=0.75pt,yscale=-1,xscale=1]

\draw    (147,108) -- (285.4,107.7) ;
\draw [shift={(287.4,107.7)}, rotate = 179.88] [color={rgb, 255:red, 0; green, 0; blue, 0 }  ][line width=0.75]    (10.93,-3.29) .. controls (6.95,-1.4) and (3.31,-0.3) .. (0,0) .. controls (3.31,0.3) and (6.95,1.4) .. (10.93,3.29)   ;
\draw   (77,134.5) .. controls (77,101.64) and (99.39,75) .. (127,75) .. controls (154.61,75) and (177,101.64) .. (177,134.5) .. controls (177,167.36) and (154.61,194) .. (127,194) .. controls (99.39,194) and (77,167.36) .. (77,134.5) -- cycle ;
\draw    (147,130) -- (285.4,129.7) ;
\draw [shift={(287.4,129.7)}, rotate = 179.88] [color={rgb, 255:red, 0; green, 0; blue, 0 }  ][line width=0.75]    (10.93,-3.29) .. controls (6.95,-1.4) and (3.31,-0.3) .. (0,0) .. controls (3.31,0.3) and (6.95,1.4) .. (10.93,3.29)   ;
\draw    (147,152) -- (285.4,151.7) ;
\draw [shift={(287.4,151.7)}, rotate = 179.88] [color={rgb, 255:red, 0; green, 0; blue, 0 }  ][line width=0.75]    (10.93,-3.29) .. controls (6.95,-1.4) and (3.31,-0.3) .. (0,0) .. controls (3.31,0.3) and (6.95,1.4) .. (10.93,3.29)   ;
\draw   (268.2,134.5) .. controls (268.2,101.64) and (290.59,75) .. (318.2,75) .. controls (345.81,75) and (368.2,101.64) .. (368.2,134.5) .. controls (368.2,167.36) and (345.81,194) .. (318.2,194) .. controls (290.59,194) and (268.2,167.36) .. (268.2,134.5) -- cycle ;

\draw (291,107) node [anchor=west] [inner sep=0.75pt]   [align=left] {feast};
\draw (145,108) node [anchor=east] [inner sep=0.75pt]   [align=left] {move};
\draw (289.4,129.7) node [anchor=west] [inner sep=0.75pt]   [align=left] {overcome};
\draw (145,130) node [anchor=east] [inner sep=0.75pt]   [align=left] {load};
\draw (289.4,151.7) node [anchor=west] [inner sep=0.75pt]   [align=left] {succumb};
\draw (145,152) node [anchor=east] [inner sep=0.75pt]   [align=left] {unload};
\draw (46.4,205) node [anchor=north west][inner sep=0.75pt]   [align=left] {\begin{minipage}[lt]{114.04pt}\setlength\topsep{0pt}
\begin{center}
Transportation operators
\end{center}

\end{minipage}};
\draw (261.4,206) node [anchor=north west][inner sep=0.75pt]   [align=left] {\begin{minipage}[lt]{84.35pt}\setlength\topsep{0pt}
\begin{center}
Mystery operators
\end{center}

\end{minipage}};

\end{tikzpicture}
\end{center}
\caption{The mapping from the operators the Transportation domain to the operators of the Mystery domain.}
\label{fig:MysteryOperatorMapping}
\end{figure}}

\newcommand{\ReasonsDecoder}{ 
\begin{table}[H] 
\begin{center} 
\resizebox{\textwidth}{!}{ 
\begin{tabular}{|M{0.1\linewidth}|P{0.9\linewidth}|}
\hline
\textbf{Reason Number} & \textbf{Justification}                                                                                                                                                                                                                                    \\ \hline
1                      & There is a predicate consistent mapping from the atoms of the primitive operators of D1 to the atoms of the primitive operators of D2 as per the operator-structure theorem.                                                                              \\ \hline
2                      & There is no predicate consistent mapping from the atoms of the primitive operators of D1 to the atoms of the primitive operators of D2 as per the operator-structure theorem.                                                                             \\ \hline
3                      & If all non-macro operators in D1 and D2 are confirmed to be primitive, then there is no predicate consistent mapping from the atoms of the primitive operators of D1 to the atoms of the primitive operators of D2 as per the operator-structure theorem. \\ \hline
4                      & At least one of D1 primitive operators does not have any PFEPO from D2.                                                                                                                                                                                   \\ \hline
5                      & If all non-macro operators in D1 and D2 are confirmed to be primitive,  then at least one of D1 primitive operators does not have any PFEPO from D2.                                                                                                      \\ \hline
6                      & At least one of D2 primitive operators does not have any PFEPO from D1.                                                                                                                                                                                   \\ \hline
7                      & If all non-macro operators in D1 and D2 are confirmed to be primitive,  then at least one of D2 primitive operators does not have any PFEPO from D1.                                                                                                      \\ \hline
8                      & D1 has more primitive operators than D2.                                                                                                                                                                                                                  \\ \hline
9                      & If all non-macro operators in D1 and D2 are confirmed to be primitive,  then D1 has more primitive operators than D2.                                                                                                                                     \\ \hline
10                     & D2 has more primitive operators than D1.                                                                                                                                                                                                                  \\ \hline
11                     & If all non-macro operators in D1 and D2 are confirmed to be primitive,  then D2 has more primitive operators than D1.                                                                                                                                     \\ \hline
12                     & There is a mapping from the predicates of D1 to the predicates of D2 that makes the reach set of D1 a subset of the reach set of D2 and D1 and D2 have an equal number of operators.                                                                      \\ \hline
13                     & There is no mapping from the predicates of D1 to the predicates of D2 can make the reach set of D1 a subset of the reach set of D2.                                                                                                                       \\ \hline
14                     & At least one of D1 operators does not have any PFEPO from D2.                                                                                                                                                                                             \\ \hline
15                     & At least one of D2 operators does not have any PFEPO from D1.                                                                                                                                                                                             \\ \hline
16                     & D1 has more operators than D2.                                                                                                                                                                                                                            \\ \hline
17                     & D2 has more operators than D1.                                                                                                                                                                                                                            \\ \hline
18                     & The two domains have different number of predicates.  Hence they are considered not functionally equivalent by our method.                                                                                                                                \\ \hline
\end{tabular}}
\caption{The description of the reasons behind the decisions returnd by D-VAL.} 
\label{tab:ReasonsDecoder} 
\end{center} 
\end{table}} 

\usepackage{array}
\newcolumntype{P}[1]{>{\centering\arraybackslash}p{#1}}

\newcolumntype{M}[1]{>{\centering\arraybackslash}m{#1}}

\newcommand{\SE}{\textit{SE }}

\newcommand{\D}[1]{$D_#1$}

\begin{document}

\title{D-VAL: An automatic functional equivalence validation tool for planning domain models \thanks{\ \ Supported by EPSRC grant EP/P510427/1 in collaboration with Schlumberger.}}

\author{\name Anas Shrinah \email anas.shrinah@bristol.ac.uk \\
 \addr University of Bristol, Bristol BS8 1UB, UK
 \AND
 \name Derek Long \email dlong6@slb.com \\
 \addr Schlumberger Cambridge Research, Cambridge CB3 0EL, UK
 \AND
 \name Kerstin Eder \email kerstin.eder@bristol.ac.uk \\
 \addr University of Bristol, Bristol BS8 1UB, UK
 }

\maketitle

\begin{abstract}
This paper introduces an approach to validate the functional equivalence of planning domain models.
Validating the functional equivalence of planning domain models is the problem of formally confirming that two planning domain models can be used to solve the same set of problems for any set of objects.
The need for techniques to validate the functional equivalence of planning domain models has been highlighted in previous research and has applications in model learning, development and extension.
We prove the soundness and completeness of our method.
We also develop D-VAL, an automatic functional equivalence validation tool for planning domain models.
Empirical evaluation shows that D-VAL validates the functional equivalence of all examined domains in less than 43 seconds.
Additionally, we provide a benchmark to evaluate the feasibility and performance of this and future related work.
\end{abstract}


\section{Introduction}
\label{Introduction}
Validation of planning domain models is a key challenge in Knowledge Engineering in Planning and Scheduling (KEPS).
Among other tasks, this activity is concerned with checking the correctness of a planning domain model with respect to its specification, requirements or any other reference.
If the reference is described informally, then the process of validating the correctness of domain models is also informal~\cite{mccluskey2017engineering}.
On the other hand, when the requirements are described formally, it is feasible to perform formal validation and to automate this process, as we will show in this paper.

Our research focuses on validating the functional equivalence of planning domain models.
Two planning domain models are functionally equivalent if both can be used to solve the same set of planning problems.
The need for a technique to analyse planning domain models for functional equivalence has been highlighted in the literature~\cite{shoeeb2011comparing,Vallati2013TowardsAP,mccluskey2017engineering}. 

One example application is the evaluation of the quality of planning model learning algorithms~\cite{aineto2018learning,zhuo2013action}.
This evaluation process can be achieved by using hand-crafted models to generate a number of plans which are then fed to the model learning method to produce the learnt planning domain models.
The functional equivalence of the original and learnt models is then checked to evaluate the quality of the learning algorithm.

Another application is to validate that the modifications performed by optimisation methods that are applied to planning domain models do not alter the functionality of the original domain.
This includes reformulation, re-representation~\cite{vallati2015effective} and tuning~\cite{hutter2009paramils} of domain models in order to increase the efficiency of planners using them. Examples of domain reformulation include macro-learning~\cite{newton2007learning,botea2005macro}, action schema splitting~\cite{areces2014optimizing} and entanglements~\cite{chrpa2012exploiting}.

The idea of proving functional equivalence between two artifacts is not new; this concept is used in proving program equivalence in software development~\cite{godlin2013regression,godlin2009regression,felsing2014automating}, and in checking circuit equivalence in hardware design~\cite{van2000sequential,kuehlmann2002combinational,mishchenko2006improvements}. 

To prove the functional equivalence of two planning domains, we introduce a method that uses a planner to find and remove any redundant operator from the given domains.
From the perspective of the functionality of planning domain models, a redundant operator is any operator that can be removed without changing the functionality of its domain.
After that, a Satisfiability Modulo Theories (SMT) solver is used to find a consistent and legal mapping between the predicates of the two domains.
Thus, it (constructively) proves the functional equivalence of two planning domain models.
Our algorithm is the VAL~\cite{howey2004val} equivalent for validating the functional equivalence of planning domain models. Hence, we refer to it as D-VAL. 

In this paper, we first formally define the functional equivalence of two planning domain models.
Then, we classify domains into two types: simple and complex.
Complex domains have two or more operators that have the same effects.
Conversely, simple domains are any domain that is not complex.
We then prove that if two simple domains are functionally equivalent, our method can find a bijective predicate mapping that makes the two domains functionally redundant.
Hence, our method is complete with regard to simple domains.
On the other hand, our method cannot disprove the functional equivalence for complex planning domain models.
Thus, our method cannot always give a conclusive verdict when validating complex domains. 

Furthermore, we introduce a test benchmark consisting of 75 functional equivalence validation tasks produced from 13 planning domain models from the International Planning Competition (IPC) \cite{IPC2014} and perform empirical experiments on this benchmark to demonstrate the feasibility of our method.
This work considers typed domains STRIPS subsets of PDDL without implicit equality predicates.

The scope of our method is limited to planning domain models with an equal number of atoms. 
This restriction causes the following limitations.
Reformulations that produce domains with supplementary atoms such as action schema splitting~\cite{areces2014optimizing} and entanglements~\cite{chrpa2012exploiting} are currently not covered in the scope of this paper.
Similarly, functionally equivalent domains with atoms in one domain derived from two or more atoms in another domain are also out of the scope of this paper. 
Nevertheless, this paper opens new research avenues by formalising and solving a slightly restricted version of the problem of validating the functional equivalence of planning domain models.
Furthermore, erroneous domains that contain invalid operators or redundant atoms require some form of preprocessing, which is not in the scope of this paper.
Moreover, symmetry reduction methods, like bagged representation~\cite{riddle2016improving}, produce planning domain models that are functionally equivalent to the original domains with respect to specific planning problems. Thus, such methods are not considered to produce functionally equivalent domains in general.
\subsection{Paper overview}\label{sec:overview}
This paper is organised as follows.
\Cref{sec:related_work} contrasts our method with related work.
\Cref{sec:preliminaries} presents the planning theory concepts used in this paper.
\Cref{sec:Functional_equivalence_domain_models} defines the functional equivalence of planning domain models and its supporting concepts: the reach set of operators, the reach set of sequence of operators, and the reach set of planning domain models.
\Cref{sec:DomainsTypes} explains the difference between simple and complex domains.
The logical steps proposed by our approach to validating the functional equivalence of two simple planning domain models are outlined in \Cref{sec:roadmap}.
In addition to that, this section also introduces the theorems that form the theoretical foundation of our method.
These theorems and their lemmas are listed in \Cref{Theorems}. 
Additionally, this section points toward the proofs of these theorems in the appendices.
Our method for validating the functional equivalence of simple domains is explained in \Cref{sec:ValidatingFESimpleDomains}.
The core of our method, the variable mapping SMT problem, is introduced in \Cref{subsec:FindingConsistentAtomMappings} and then detailed in \Cref{sec:SMT}.
\Cref{sec:checkMacro,sec:checkPrimitive,sec:findSOSOsForO,sec:ComparingReachsetsOperatorAndUion} describes the tasks that need to be performed by our method to remove non-primitive operators before the variable mapping SMT problem can be compiled and solved. 
\Cref{sec:example} provides a concrete example that illustrates the outputs of our method for validating the functional equivalence between the Mystery domain and its undisguised version.
\Cref{sec:ValidatingFEComplexDomains} explains our approach for proving the functional equivalence of complex models.
\Cref{sec:experiment} reports and discusses the experimental results.
Finally, \Cref{sec:conclusion_and_future_work} concludes the paper and suggests future work.

\section{Related work} %
\label{sec:related_work}

Our definition of planning domain model functional equivalence is a relaxed version of \citeauthor{shoeeb2011comparing}\textquotesingle s~\citeyear{shoeeb2011comparing} weak equivalence definition.
In their work, \citeauthor{shoeeb2011comparing} defined two flavours of domain equivalences.
They first defined the planning domain model strong equivalence as a one-to-one mapping that maps the names of predicates, variables, operator schema and types from one domain to another.
They also defined the planning domain model weak equivalence as a relation between two domains when both can represent the same set of planning problems, and every valid plan that can be produced from the first domain is also a valid plan in the second domain under a certain bijective mapping of the components of the two domains.

We relax their definition in two ways.
Firstly, we require a bijective mapping between the predicates and variables, but not necessarily between every operator schema of functionally equivalent domains.
Nevertheless, in our definition, we request a bijective mapping between primitive operators, particular type of operator schemata, in the domains under validation.
Secondly, we consider planning domain models functionally equivalent if they can solve the same set of planning problems, even if the solutions differ. 

In fact, it can be argued that the weak and strong equivalence relations presented in the authors' research are identical.
For two domains to produce the same set of plans for the same set of problems, the directed graphs representing the reachable state space of both domains must be isomorphic.
Isomorphism is a strong bi-simulation relation.
For the directed graphs of the reachable state space of two domains to be isomorphic, both domains should be identical apart from the names of the domains\textquotesingle \ components. 
Hence, the requirement to generate the same plans for every planning problem can be achieved only through strong equivalence, not by a weak equivalence as proposed by \citeauthor{shoeeb2011comparing}~\citeyear{shoeeb2011comparing}.

The logic behind our definition is that if two methods are proven to always find valid answers for any set of problems, not necessarily the same answers, then we say these two methods can do the same job regardless of the steps taken by each method.
Thus, we consider them functionally equivalent.
With this mindset, we propose our planning domain model functional equivalence definition. 
If two domains can be used to solve the same set of problems (under a certain mapping) regardless of the actions taken, then we call them functionally equivalent domains. 
This paper extends \citeauthor{shoeeb2011comparing}\textquotesingle s~\citeyear{shoeeb2011comparing} work by formally defining planning domain model functional equivalence and proposing a method to prove this equivalence. 
Moreover, we support our method with a sound and proven set of theorems along with their proofs, which form the theoretical basis of our method.

\citeauthor{mccluskey2017engineering}~\citeyear{mccluskey2017engineering} argue that the correctness of domain models is an essential factor in the overall quality of the planning function. 
They consider a knowledge model to be a planning domain model and a planning problem instance and suggest translating the components of the knowledge model into assertions.
A knowledge model is then said to accurately capture its requirements if the interpretation given by the requirements satisfies the assertions in the knowledge model.

Our notion of ``functional equivalence'' is closely related to their notion of ``accuracy''.
If two planning domain models are functionally equivalent, both domain models satisfy the same functional requirements and represent these requirements to the same degree of accuracy.
Note that our notion of ``functional equivalence'' does not consider the ``operationality'' criterion described by \citeauthor{mccluskey2017engineering}~\citeyear{mccluskey2017engineering}.

In addition to that, the authors outline an approach to checking the accuracy of operators from a planning domain model with the help of a single planning instance.
Their approach is limited to individual planning instances, whereas our proposed method is independent of the planning problems.
Besides that, this paper describes a working implementation of our method based on proven theoretical bases.
Moreover, our research reports on the feasibility of our method by means of empirical evaluation. 

\citeauthor{cresswell2013acquiring}~\citeyear{cresswell2013acquiring} define the equivalence of two domains with regards to a planning instance using graph isomorphism. According to their definition, two domains are equivalent if the directed graphs representing the reachable state space of both domains are isomorphic.
This criterion proves the equivalence between the two domains for individual planning instances.
 On the contrary, our method proves the functional equivalence for any set of problems. 
 Another subtle difference is that our method is invariant to the domains\textquotesingle \ state space paths as it is a weak equivalence relation, whereas their approach checks strong equivalence. 
 Hence, their approach is sensitive to any variation in the possible state space paths between the two domains. The path sensitivity in their method is due to the fact that their definition is based on graph isomorphism. 
 Furthermore, proving isomorphism between two graphs is computationally dependent on the size of the planning problems, whereas the computational cost of our method is dependent only on the size of the given domains.

The planning domain models functional equivalence problem is also related to model reconciliation and maintenance research. 
The following paragraphs contrasts the models functional equivalence problem and our validation approach to some of the prominent research in these areas.

The model reconciliation problem introduced in~\cite{chakraborti2017plan} is concerned with changing one model to make it closer to another model with respect to the cost of a given plan.    
Given two different planning domain models $M_1$ and $M_2$, the model reconciliation task aims to change $M_1$ to $\widehat{M_1}$ such as an optimal plan that is produced using $M_2$ is also optimal when interpreted using the modified model $\widehat{M_1}$.
Note that two reconciled planning domain models are not necessarily functionally equivalent as the reconciliation process is performed with respect to individual planning instances.
Moreover, unlike our method, the model reconciliation approach proposed in~\cite{chakraborti2017plan} assumes both models $M_1$ and $M_2$ have the same set of predicates and operators. Our approach relaxes this condition and aims to find a mapping $F_p$ between the predicates of the two domains that makes the two domains functionally equivalent. 

\citeauthor{ijcai2019p83}~\citeyear{ijcai2019p83} generalise the work in~\cite{chakraborti2017plan} by learning a model approximation of the mental model of the human user through interacting with the user.
The user is asked whether transitions from the Markov Decision Process (MDP) that represents the robot model are explainable.
After the human model is learnt, this method reconciles the user approximated model with the robot model with respect to some execution trace.
This research discusses the difference between model reconciliation processes that are intended to to explain an optimal policy or an execution trace of of an MDP.
However, the reconciliation method proposed in this work is limited to explaining individual execution traces of an MDP; thus, this method also cannot be used to prove the functional equivalence of two models.
Moreover, the transitions of the MDP model, including the states and the action labels, are subset of the transitions of the learned model which is then reconciled with the MDP model.
In contrast, as explained earlier, our method is able to prove functional equivalence of models with different states and transition labels.

The model maintenance problem addresses the challenge of updating a planning domain model $M_1$ to match the always evolving mental model of the the user $M_2$. 
In dynamic environments, a users must update their understanding to reflect changes in the environment.
As a result of these changes, the two models $M_1$ and $M_2$ drifts apart.
A system called Marshal that solves this problem is proposed in~\cite{bryce2016maintaining}. 
Marshal interacts with human users through queries to provides observations to a particle filter which anticipates the model that  most closely resembles the mental model of the user.
This method uses stochastic process to learn a model through interaction with a user, thus it is not automated and cannot be guaranteed to learn the actual mental model of the user. 

Macro generation or macro learning is a technique that helps increasing the speed of the planning process~\cite{botea2005macro,newton2007learning,coles2007marvin,chrpa2015online}.
Macro-learning methods augment planning domain models with macro operators. 
These macro operators create shortcuts in the state spaces of planning problems, hence improving the search for goal states.
The ultimate goal of macro-learning methods is to find the most effective macros in the least amount of time, so these macros can be added to a given planning domain model.
While macro-learning methods add shortcuts to the state spaces of planning problems, our approach removes such shortcuts during the process of validating the functional equivalence of planning domain models. 
Our method tests if any of the operators of a given domain is a macro operator. An operator is considered as a macro operator if there is a sequence of operators from the same domain that has the same add and delete effects as the tested operator. 
After macros are identified, we remove them from the given domain.
Removing macro operators from a planning domain model does not affect the reachability of its planning problem ~\cite{newton2007learning}, but it essential part of the process of proving function equivalence between planning domain models. 
Other distinctive difference is that, unlike macro-learning problem, our validation approach must find any possible macros in the given planning domain models.
On the other hand, macro-learning methods does not have to be exhaustive.

Domains functional equivalence and model recognition solve different problems. However, the model recognition method proposed in~\cite{aineto2019model} and our approach both compile a search problem into a classical planning problem.
The model recognition problem is concerned with identifying the model that better explains a partially observed plan execution.
Given a set of models $\mathbb{M} = \{M_1,\dots,M_n\}$ and a partially observed plan execution $\mathcal{O}$, the method proposed in~\cite{aineto2019model} recognises the model $M_i$ that best explains the given observation $\mathcal{O}$. 
This method compares between the models in $\mathbb{M}$ based on the number of atoms insertions and deletions that need to be applied on the operators of each model so that the modified models satisfy the observation $\mathcal{O}$.
This number of modified atoms is called the edit distance between a model $M$ and its modified model $M'$.
The model that requires the minimum edit distance to produce a modified model that satisfies $\mathcal{O}$ is more likely to be the model that best explains the observation $\mathcal{O}$.
The authors compile the problem of calculating the edit distant needed to modify a model to satisfy a given observation $\mathcal{O}$ as a meta-planning problem.
The solution to this problem is a plan that consists of meta-actions.
Each meta-action inserts or deletes an atom to one of the action schemata of the model $M$, or validates the application of a modified action schema to the observation $\mathcal{O}$.
The initial state of this meta-planning problem is a model $M$ and observation $\mathcal{O}$. The goal is met when the planner finds the set of meta-actions that transform $M$ into $M'$ which satisfies $\mathcal{O}$.

Our method also compiles a search problem into a classical planning problem.
The approach proposed in this paper formulates the problem of constructing a macro operator that has the same add and delete effects as a given operator from the same domain as a planning problem.
This task is needed in the process of checking whether an operator is a primitive or a macro operator as part of validating the functional equivalence of planning domain models.

\section{Preliminaries} %
\label{sec:preliminaries}
In classical planning, world objects are represented by unique constants which are the elements of a finite set {\it Obj}.
The properties of the objects are described using atoms.
Atoms consist of predicate letters that are applied to a set of variables.
A ground atom has its variables assigned to object constants from {\it Obj}.
Let $P$ be a finite set of the predicates that appears in the atoms that describe the status of all the objects of a world. 

A state of the world, $s$, is defined by the status of all objects in this world.
The status of an object is defined by the truth evaluation of its atoms.
The state space of a world is the set $S$ of all states, where $|S| = 2^{\{\text{number of all ground atoms}\}}$.
False atoms are not included in the state definition and a closed-world assumption is made.

A planning domain model is a tuple $D = (P,O)$, where $P$ is a set of predicates and $O$ is a set of planning operators. 
A planning operator, $o$, is a tuple $\mathit{o = (name(o), Pre(o), Add(o), Del(o))}$
where $name(o) = op\ name(x_1,\hdots, x_k)$, ``$op\ name$'' is a unique operator name and $x_1,\hdots,x_k$ are variable symbols (arguments or parameters) appearing in the operator\textquotesingle s atoms, and $Pre(o), Add(o)$ and $Del(o)$ are sets of ungrounded atoms, specifying preconditions, add effects and delete effects of the operator $o$, respectively.
The intersection of the precondition and add effects of an operator must be empty.
The intersection of the add effects and delete effects of an operator must be empty.
The delete effects of an operator must be a subset of the preconditions of that operator.
The atoms in operators consist of predicate letters from $P$ that are applied to variables taken only from $x_1,\hdots,x_k$.

An action is a ground instance of an operator.
Actions are instantiated from operators by grounding the atoms of the operators,~i.e. by substituting their variables with objects.
An action name is the same as the name of the operator from which the action has been instantiated.
An action\textquotesingle s preconditions, add effects and delete effects are sets of ground atoms.
Action $a$ is applicable in state $s$ if and only if $Pre(a) \subseteq s$. The application of $a$ in $s$ results in the successor state $ \gamma(s,a) = (\mathit{s \setminus Del(a)) \cup Add(a)}$ provided that $a$ is applicable in $s$.

\subsection{Functions definitions}
In this section, we well define the functions used in the proof of this theorem.
We will use the operator $navigate$ from the rover domain to provide examples of these functions.

\begin{verbatim}
(:action navigate
:parameters (?x - rover ?y - waypoint ?z - waypoint) 
:precondition (and (can_traverse ?x ?y ?z) (available ?x) 
   (at ?x ?y) (visible ?y ?z))
:effect (and (not (at ?x ?y)) (at ?x ?z) ) )
\end{verbatim}

\begin{itemize}
 \item The function $Pred$ returns the predicate name of a given atom. $Pred$: atom $\rightarrow$ predicate name. For instance, $ Pred((at\ ?x\ ?y)) = at$.

 \item The function $Arity$ returns the arity of a given atom. $Arity$: atom $\rightarrow$ $\mathbb{N}$. For instance, $ Arity((at\ ?x\ ?y)) = 2$.

 \item The function $Var$ returns the set of variables of a given atom. $Var$: atom $\rightarrow$ set of variables. For instance, $ Var((at\ ?x\ ?y)) = \{?x, ?y\}$.

 \item The function $Position$ returns the position of a given variable in a given atom. $Position$: (variable,atom) $\rightarrow$ $\mathbb{N}$. For instance, $ Position(?x, (at\ ?x\ ?y)) = 1 $.
 
 \sloppy  \item The function $Predicates$ returns the set of predicates of a given operator. $Predicates$: operator $\rightarrow$ set of predicates. For instance, $ Predicates(\text{navigate}) = \{ \text{can\_traverse}^3,\text{available}^1,\text{at}^2,\text{visible}^2\}$.
 The superscript of the predicate name is its arity.

 \item The function $Atoms$ returns the set of atoms of a given operator. $Atoms$: operator $\rightarrow$ set of atoms. For instance, $ Atoms(\text{navigate}) = \{(can\_traverse\ ?x\ ?y\ ?z), \allowbreak (available\ ?x),\allowbreak (at\ ?x\ ?y), \allowbreak (visible\ ?y\ ?z), \allowbreak (at\ ?x\ ?z)\}$.

 \item The functions $Domain(f)$ and $Range(f)$ return the domain and range of the function $f$, respectively.

\end{itemize}

\subsection{Sequence of operators consolidation}\label{sec:sequenceOfOperatorsConsolidation}
In the description of our approach, we refer to the terms sequence of operators, configured sequence of operators and consolidated sequence of operators.
This section explains these terms and the process of consolidating a sequence of operators.

Consolidating the operators of a sequence of operators is an iterative process.
The procedure starts by consolidating the first two operators, then it consolidates the next operator with the outcome of the previous consolidation step.
The consolidation process continues until all operators are considered.
The result of each consolidation process is a single operator.
The produced operator is a macro operator build from the consolidated operators. 
If a sequence of operators consists of a single operator $o$, then the product of consolidating this sequence of operators is the operator $o$. 

The consolidation procedure unifies or unites the parameters of the two given operators based on the types of their parameters.
If the type of the considered parameter is in the types of the parameters of the other operator, then the consolidation procedure can either unifies or unites this parameter with some of the parameters of the other operator.
The unification process unifies a parameter from one operator with a parameter from the other operator if both parameters are of the same type.
On the other hand, the unionisation process adds the considered parameter to the parameters from the other operator with the same type.
However, if the type of the considered parameter is not in the types of the other operator, then the consolidation procedure must perform a unionisation process to add the considered parameter along with its type to the parameters and types of the produced macro.

For a given unification and unionisation configuration of the parameters of the two operators $o_1$ and $o_2$, the consolidation process produces a macro $m$ if none of the atoms in the preconditions of $o_2$ are in the delete effects of $o_1$.
Otherwise, we say consolidating the two operators $o_1$ and $o_2$ is an invalid process.
The preconditions of $m$ are the preconditions of $o_1$ and the preconditions of $o_2$ that are not supported by add effects from $o_1$.
The add effects of $m$ are the add effects of $o_2$ and the add effects of $o_1$ that are not in the delete effects of $o_2$.
The delete effects of $m$ are the delete effects of $o_2$ and the delete effects of $o_1$ that are not in the add effects of $o_2$.

So, we say a macro is made of a sequence of operators with a specific unification and unionisation configuration of the parameters of each two consecutive operators such that the consolidation of this sequence of operators is a valid process.
We call such sequence of operators ``configured sequence of operators''. 
Thus, we say a macro is made from a configured sequence of operators.
For brevity, from here onwards we will refer to ``configured sequence of operators'' just as ``sequence of operators''. 
This does not create any ambiguity because we do not use the concept of unconfigured sequence of operators.

In the next section, we define the reach sets of operators, sequence of operators and planning domain models for any set of objects.
In addition to that we define the functional equivalence of planning domain models.
\section{The definition of the functional equivalence of planning domain models} %
\label{sec:Functional_equivalence_domain_models}
The functionality of a planning domain model is characterised by the set of planning problems that can be solved using this domain.
For a domain to support solving a planning problem, a planner has to be able to use this domain to produce the required transitions in the state space from the initial state to one of the goal states.
Thus, for a given set of objects, we call the set of all tuples of the start and end states of all transitions that can be produced using the operators of a planning domain model as the reach set of this domain.
Note that these transitions can result from applying any legal sequence of actions to any state that satisfies the preconditions of the first action in the sequence of actions.
A legal sequence of actions means a sequence of actions such that all preconditions of each action are satisfied in the state that results from applying the previous action in the sequence.
Therefore, two planning domain models with equal reach sets are considered functionally equivalent as they can solve the same set of problems.
Defining the reach set of a planning domain model requires the following definitions.
\begin{defn}\label{Def:reach_set_operator}
\emph{The reach set of a primitive operator} $o$ over a set of objects {\it Obj} is defined using the set of actions $A_{\{o,Obj\}}$ which is instantiated from the operator $o$ with the set of objects {\it Obj} as
$\Gamma(o,Obj) = \{ (s,\gamma(s,a)) \mid s \in S, \ a \in A_{\{o,Obj\}} \text{ and } a \text{ applicable in } s\}$.
\end{defn}
\begin{defn}
\emph{The reach set of a sequence of operators} $seq$ over a set of objects {\it Obj} is defined using the set of action sequences $\Pi_{\{seq,Obj\}}$ which are instantiated from the sequence of operators $seq$ with the set of objects {\it Obj} as
\sloppy $\Gamma(seq,Obj) = \{ (s,\gamma(s,\pi)) \mid s \in S, \ \pi \in \Pi_{\{seq,Obj\}} \text{ and } \pi \text{ applicable in } s\}$.
Where $\gamma(s,\pi)$ is the successor state of a state $s$ when the sequence $\pi$ of actions is applied.
A sequence of operators could consist of a single or many operators where some items might be repeated.
\end{defn}
\begin{defn}\label{PlanningDomainModelReachability}
The \emph{reach set of a planning domain model} $D$ over a set of objects {\it Obj} is then defined as
$\Gamma(D,Obj) = \bigcup_{seq \in SEQ} \Gamma(seq,Obj)$, where $SEQ$ is the set of all possible sequences of operators generated from $D$.
\end{defn}
For two planning domain models to have equal reach sets, both domains must have the same predicates.
We can relax this condition by requesting only a bijective mapping between the predicates of equal arities in the two domains. 
Under such mapping, two domains with different predicates but with an equal number of predicates with equal arities can be used to represent the same state space.
Thus if these two domains are functionally equivalent, they can have the same reach sets under this mapping for any set of objects.

Now we will informally define the planning domain model functional equivalence. Two planning domains $D_1$ and $D_2$ are \textbf{\emph{functionally equivalent}} if and only if there is a bijective mapping $F_p$ from the predicates of $D_1$ to the predicates of $D_2$ with equal arities such that when the predicates of $D_1$ are substituted with the predicates from $D_2$ using the predicate mapping $F_p$, the reach sets of the produced domain $F_p(D_1)$ is equal to the reach set of $D_2$ for any set of objects.

\begin{defn} \label{Def:functional-equivalence}
Two planning domain models $D_1 = (P_1,O_1)$ and $D_2 = (P_2,O_2)$ are \textbf{\emph{functionally equivalent}}, $D_1 \equiv_{\mathit{Func}} D_2$, \textbf{iff}:
\begin{enumerate} 
 \item $ \exists F_p: P_1 \twoheadrightarrowtail P_2 \text{ where } \operatorname{graph}(F_p) = \{ (p,F_p(p)) \in P_1 \times P_2 : p \in P_1 \text{ and } arity(p) = arity(F_p(p)) \}$
 \item $F_p(D_1)$ is the image of $D_1$ using $F_p$ to substitute its predicates with those of $D_2$.
 \item $\forall Obj : \ \Gamma(F_p(D_1),Obj) = \Gamma(D_2,Obj)$. 

\end{enumerate}
\end{defn}
Note that the functional equivalence of planning domain models is independent of the planning tasks for which these domains might be used.
This relation depends only on the set of predicates and operator schemata in the given domains. Therefore, this relation needs to be proven with regards to the planning domain models regardless of any set of objects, initial states or goal states.

The functional equivalence of planning domain models is a reflexive, symmetric and transitive relation; thus, the functional equivalence of planning domain models is an equivalence relation.
According to this definition, the functional equivalence between two planning domain models can be proven by showing that, for any set of objects, the reach set of one domain is equal to the reach set of an image of the other domain under a certain predicate mapping. 

\subsection{Operators types}
We classify operators into four categories based on the relations between their reach sets and the reach sets of other operators and sequences of operators from the same domain.
The main type is primitive.
\begin{defn}\label{def:primitieOperator}
An operator $o$ is a \textbf{\emph{primitive operator}} if the reach set of $o$ is not a subset of the union of the members of any subset of the set of all sequences of operators in its domain without $o$.
$$o \in Primitive(O) \rightarrow \forall Seq \subseteq SEQ(O \setminus o) \ (\Gamma(o) \not \subseteq \bigcup_{seq \in Seq}(\Gamma(seq))$$
\end{defn}
In the special case when the set $Seq$ has just one sequence, and this sequence has just one primitive operator, we can infer that the reach set of a primitive operator is not a subset of the reach set of any other primitive operator in its domain.
Thus, primitive operators are not redundant by definition.
\begin{defn}
An operator $o$ is a \textbf{\emph{macro operator}} if the reach set of $o$ equals the reach set of a sequence of primitive or macro operators from its domain without $o$. 
$$o \in Macro(O) \rightarrow \exists seq \subseteq SEQ(Primitive(O) \cup Macro(O \setminus o))\ (\Gamma(o) = \Gamma(seq)$$
\end{defn}
\begin{defn}
An operator $o$ is a \textbf{\emph{split operator}} if the reach set of $o$ equals the union of the reach sets of two or more primitive or split operators from its domain without $o$.
$$o \in Split(O) \rightarrow \exists O_o \subseteq Primitive(O) \cup Split( O \setminus o)\ (\Gamma(o) = \bigcup_{o_o \in O_o}(\Gamma(o_o))$$
\end{defn}
\begin{defn}
An operator $o$ is a \textbf{\emph{split macro}} if the reach set of $o$ is equal to the union of the reach sets of two or more macros or at least one macro and one primitive operator from its domain.
\begin{multline}
o \in Split\_macro(O) \rightarrow \exists M \subseteq Macro(O) : |M| >1\ (\Gamma(o) = \bigcup_{m \in M}(\Gamma(m)) \ \vee \\
\exists M \subseteq Macro(O) : |M| >0, \exists O_p \subseteq Primitive(O): |O_p| >0 \ ( \Gamma(o) = ( \bigcup_{o_p \in O_p}(\Gamma(o_p ))) \cup (\bigcup_{m \in M}(\Gamma(m)))) \nonumber
\end{multline}
\end{defn}

The primitive operators of a domain decide which end states are reachable from which initial states.
On the other hand, the macro operators of a domain only introduce a shorter transition between two states if a sequence of primitive operators connects them.
In addition, the split operators do not add any new transitions to the reach sets of planning domain models;  their transitions must be covered by the transitions of two or more primitive operators.
Therefore, primitive operators are the only source of functionality for their domains. 
\section{The types of planning domain models }\label{sec:DomainsTypes}
According to \Cref{Def:functional-equivalence}, functionally equivalent planning domain models must have equal reach sets for any set of objects.
Hence, we can prove or disprove the functional equivalence of two planning domain models by comparing their reach sets.
This paper shows that different types of planning domain models require different methods to validate the equality of their reach sets.
We differentiate between two types of planning domain models; complex and simple.

We say a domain is complex if it has at least two operators with similar effects.
On the other hand, a domain is said to be simple if none of its operators shares the same effects.
We say two operators have similar effects if they have similar add and delete effects. 
For an effect in one operator to be similar to an effect in another operator, the two effects must have the same predicate and the same types. 
Additionally, both effects must have an equal number of parameters of each type.

Simple domains are more common than complex ones.
Out of the 13 IPC domains used in the empirical evaluation of our method in \Cref{sec:experiment}, we found only four domains are complex: Elevator, Floor-tile, Child-snack and Pipesworld.
Each of these domains has two or more operators that share the same effects.
For instance the operators ``up'' presented in \Cref{fig:up} and the operator ``down'' depicted in \Cref{fig:down} from the Elevator domain both have (lift-at ?f2) as their add effect and (not (lift-at ?f1)) as their delete effect.
\begin{figure}
\begin{center}
\begin{Verbatim}[commandchars=\\\{\}]
(:action up
 :parameters (?f1 - floor ?f2 - floor)
 :precondition (and (lift-at ?f1) (above ?f1 ?f2))
 :effect (and (lift-at ?f2) (not (lift-at ?f1))))
\end{Verbatim}
\end{center}
\caption{The operator ``up'' from the Elevator domain.}
\label{fig:up}
\end{figure}

\begin{figure}
\begin{center}
\begin{Verbatim}[commandchars=\\\{\}]
(:action down
 :parameters (?f1 - floor ?f2 - floor)
 :precondition (and (lift-at ?f1) (above ?f2 ?f1))
 :effect (and (lift-at ?f2) (not (lift-at ?f1))))
\end{Verbatim}
\end{center}
\caption{The operator ``down'' from the Elevator domain.}
\label{fig:down}
\end{figure}

Each set of operators that have similar effects forms an equivalence class. 
We call such equivalence classes the sets of operators with Similar Effects (\textit{SE}).
We consider the effects similarity relation to be not reflexive; hence, an \SE set cannot have a single operator. 
Therefore, an \SE set exists in a domain if this domain has at least two operators with similar effects.
So, for a domain to be complex, it has to have at least one \textit{SE} set.

Since the methods of validating the functional equivalence of simple and complex domains are different, the first step in any validation task is to decide whether the given domains are simple or complex. 
To simplify the implementation of D-VAL, we assume domains do not have predicates that share the same name but have different arities or different types.

To test if a domain is simple or complex, D-VAL creates a unique identifier for each operator from the given domain.
The unique identifier of an operator $o$ is the concatenation of the predicates of the add effects followed by the predicates of the delete effects of the operator $o$.
To distinguish between the predicates of the add effects and the delete effects, we suffix the latter with the word ``Del-''.
After that, D-VAL groups operators of similar identifiers, if they are two or more, into an \SE set.
If a domain has one \SE set at least, then this domain is complex; otherwise, it is simple.

Validating the functional equivalence of simple planning domain models is explained in \Cref{sec:roadmap,Theorems,sec:ValidatingFESimpleDomains,sec:SMT,sec:checkMacro,sec:checkPrimitive,sec:findSOSOsForO,sec:ComparingReachsetsOperatorAndUion,sec:example}. 
Our approach can only prove the functional equivalence of complex planning domain modes in some cases; thus, it cannot disprove that two domains are functionally equivalent.
\Cref{sec:ValidatingFEComplexDomains} explains why we need a different method to validate the equality of the reach set of complex models, and it clarifies why we cannot always prove the functional equivalence of complex domains.
Furthermore, it discusses the case when our method can prove the functional equivalence of complex models and demonstrates how to prove the functional equivalence of such domains. 

The following section outlines the proposed steps to validate the functional equivalence of two simple planning domain models.
\section{The road map of validating the functional equivalence of two simple planning domain models}\label{sec:roadmap}
This section presents the logical steps required to prove and disprove the functional equivalence of planning domain models without split operators. 
This explanation introduces our approach and theorems that support the soundness and completeness of our method.

Assume two functionally equivalent planning domain models $D_1$ and $D_2$ and a bijective function $F_p: P_1 \twoheadrightarrowtail P_2$.
Let $F_p(D_1)$ be the image of $D_1$ using $f$ to substitute its predicates with those of $D_2$ with equal arity. 
To prove $D_1$ and $D_2$ are functionally equivalent, we have to prove the reach sets of $F_p(D_1)$ and $D_2$ are equal for any set of objects, as per \Cref{Def:functional-equivalence}.
Of course, we could also prove the functional equivalence between $D_1$ and $D_2$ by proving the reach sets of $D_1$ and $F_p^{-1}(D_2)$ are equal for any set of objects because the functional equivalence relation is an equivalence relation as per \Cref{Def:functional-equivalence}.
However, in this paper, we will focus on proving that the reach sets of $F_p(D_1)$ and $D_2$ are equal for any set of objects.

So, the question is how to prove $\Gamma (F_p(D_1),Obj) = \Gamma (D_2,Obj)$?
To address this question, we have taken a divide and conquer approach. We have reduced the relation between the reach sets of two domains to relations between their operators.
This simplification is captured in \Cref{th:SimpleDomainsReachabilityTheorem} in \Cref{Theorems} and proven in \Cref{ProofOfTheoremRT}.
This theorem is based on the observation that the reach set of any simple domain is produced exclusively by its primitive operators, not by its macro operators.
Thus, for two planning domain models $D_1$ and $D_2$ to have equal reach sets for any set of objects, this theorem requests the image of the reach set of each primitive operator $o$ in \D1 to be equal to the reach set of a primitive operator $o'$ from \D2 under a bijective predicate mapping $f_p$ from the predicates of $o$ to the predicates of $o'$ for any set of objects as well.

Throughout this paper, we will compare the reach sets of operators from two different domains for any set of objects.
For brevity, we will not repeat the phrase ``for any set of objects'' unless we need to re-emphasise the scope of our method.
Furthermore, since the operators of two different domains have different predicates, their reach sets will always be different unless they are compared through a predicate mapping.
Thus, whenever we say the reach set of an operator $o$ from a domain \D1 is equal to the reach set of an operator $o'$ from another domain \D2, we mean the image of the reach set of $o$ under a predicate mapping is equal to the reach set of the operator $o'$.
In addition to that, when we say the reach set of $o$ is not equal to the reach set of $o'$, we mean the image of the reach set of $o$ under any predicate mapping is not equal to the reach set of $o'$. 
Nevertheless, we will clarify what predicate mapping is used in some comparisons when omitting its mention can cause unambiguity.
Moreover, the predicate mappings to which we refer implicitly or explicitly when we compare the reach sets of planning domain models and operators are always bijective mapping between predicates of equal arities. 

By definition, primitive operators in one domain have different reach sets.
Therefore, under a predicate mapping $F_p: P_1 \twoheadrightarrowtail P_2$, the images of the reach sets of any two primitive operators in $D_1$ cannot be equal to the reach set of a single primitive operator in $D_2$.
It follows that if $D_1$ and $D_2$ have an unequal number of primitive operators, then the image of the reach set of each primitive operator in one domain cannot be equal to the reach sets of a primitive operator in the other domain.
Hence, for $D_1$ and $D_2$ to have an equal reach set, \Cref{th:SimpleDomainsReachabilityTheorem} also mandates that the two domains must have an equal number of primitive operators. 

Note that the mapping $f_p$ that makes the reach set of a primitive operator from $D_1$ equal to the reach set of a primitive operator from $D_2$ is limited to the predicates of these two operators.
However, for $D_1$ and $D_2$ to be functionally equivalent, \Cref{Def:functional-equivalence} requires the existence of an overall predicate mapping $F_p$ from the predicates of $D_1$ to the predicates of $D_2$ such that the reach set of $F_p(D_1)$ is equal to the reach set of $D_2$.
Therefore, \Cref{th:SimpleDomainsReachabilityTheorem} dictates that the union of the individual predicate mappings $f_p$ between the predicates of the primitive operators of the two domains must be a well-defined function.

According to \Cref{th:SimpleDomainsReachabilityTheorem}, to prove $\Gamma (F_p(D_1),Obj) = \Gamma (D_2,Obj)$, we have to prove the following three conditions.
\paragraph{ First condition} \textit{$D_1$ and $D_2$ must have equal number of primitive operators.}

To prove this condition, we must calculate the set of primitive operators in each domain and show that these two sets have equal cardinality.
This task is further explained in \Cref{sec:ValidatingFESimpleDomains}.
\paragraph{ Second condition} \textit{The reach set of an image of each primitive operator $o$ in $D_1$ must be equal to the reach set of a primitive operator $o'$ in $D_2$.}

The image of $o$ is produced using a predicate mapping $f_p$ that substitutes the predicates of $o$ with predicates from $o'$ with equal arity.
To prove that the reach set of an image of one operator is equal to the reach set of a primitive operator from another domain without reasoning about individual elements of the reach sets, we have to compare the structures of operators.
\Cref{th:operatorsStructurReachSet} relates the equality of the reach sets of two operators $o$ and $o'$ to the existence of a mapping between the atoms of the two operators, such that this mapping satisfies a set of constraints that guarantees the two operators have the same structure. 
\Cref{th:operatorsStructurReachSet} is formalised in \Cref{Theorems} and proven in \Cref{ProofOfTheoremOSRST}.

To satisfy the second condition of \Cref{th:SimpleDomainsReachabilityTheorem}, we have to find one atom mapping from the atoms of each primitive operator in $D_1$ to the atoms of a primitive operator in $D_2$, such that each of these atom mappings satisfies the constraints of \Cref{th:operatorsStructurReachSet}.
The existence of such atom mapping from the atoms of an operator $o$ from \D1 to the atoms of an operator $o'$ from \D2 proves the existence of a bijective predicate mapping $f_p$ from the predicates of $o$ to the predicates of $o'$ such that the reach set of $f_p(o)$ is equal to the reach set of $o'$.
The task of finding such mappings is explained in \Cref{subsec:FindingPFEPOs} and \Cref{subsec:FindingConsistentAtomMappings}.

\paragraph{ Third condition} \textit{The individual mappings $f_p$ from the predicates of each primitive operator in $D_1$ to the predicates of each primitive operator in $D_2$ must be consistent.}

By consistent, we mean the union of these mappings is a well-defined function.
This condition can be satisfied by imposing a predicate consistency constraint that ensures the atoms from $D_1$ with the same predicate $p$ are mapped to atoms in $D_2$ with one predicate $p'$. 
This constraint differ from the constraints of \Cref{th:operatorsStructurReachSet} by the scope.
The scope of the constraints of \Cref{th:operatorsStructurReachSet} is limited to individual operators.
On the other hand, the scope of this constraint includes all atoms in all operators of both domains. 
This consistency constraint is explained in \Cref{subsec:SMTThirdCondition} and formalised in \Cref{eq:SPC}.

We have shown that proving the functional equivalence of functionally equivalent domains depends on satisfying the premises of \Cref{th:operatorsStructurReachSet} and \Cref{th:SimpleDomainsReachabilityTheorem}.
Since these theorems are proven as logical equivalences, we can safely infer that two planning domain models are not functionally equivalent if these domains do not satisfy any of the antecedents of these theorems.

\begin{figure}
\caption{The road map to prove the functional equivalence of two simple planning domain models.}
\begin{center}
\begin{tikzpicture}[
 font=\sffamily,
 every matrix/.style={ampersand replacement=\&,column sep=2cm,row sep=2cm},
 source/.style={draw,thick,rounded corners,fill=yellow!20,inner sep=.3cm},
 process/.style={draw,thick,circle,fill=blue!20},
 sink/.style={source,fill=green!20},
 datastore/.style={draw,very thick,shape=datastore,inner sep=.3cm},
 dots/.style={gray,scale=2},
 to/.style={->,>=stealth',shorten >=1pt,semithick,font=\sffamily\footnotesize},
 every node/.style={align=center}]

 \node[source] (FindMapping) {$\forall o \in Primitive(O_1), \exists o' \in Primitive(O_2),$ \\ $ \exists f_t: Atoms(o) \twoheadrightarrowtail Atoms(o')$ as per \Cref{th:operatorsStructurReachSet}};

 \node[sink, below=of FindMapping] (THoperatorsStructurReachSet) { \Cref{th:operatorsStructurReachSet}};

 \node[source, below=of THoperatorsStructurReachSet] (equal) {$\forall o \in Primitive(O_1), \exists o' \in Primitive(O_2),$ \\ $ \exists f_p: Pred(o) \twoheadrightarrowtail Pred(o') ( \Gamma(f_p(o)) = \Gamma(o') )$};

 \node[sink, below=of equal] (THreachabilityTheorem) { \Cref{th:SimpleDomainsReachabilityTheorem} \\ $(F_p = \bigcup\limits_{f_p \in Range(R^{\prime}_{om}) } f_p )\text{ is a well-defined function}$ \\ $\wedge$\\ $|Primitive(O_1)| = |Primitive(O_2)|$};

 \node[source, below=of THreachabilityTheorem] (d1Eqd2) {$\exists F_p: Pred(D_1) \twoheadrightarrowtail Pred(D_2) $ \\ $(F_p(\Gamma(D_1)) = \Gamma(D_2))$}; 

 \node[sink, below=of d1Eqd2] (DEFfunctionalEquivalence) { \Cref{Def:functional-equivalence}};

 \node[source,below=of DEFfunctionalEquivalence] (FuncEq) {$D_1 \equiv_{\mathit{Func}} D_2$};

 \draw[to] (FindMapping) -- (THoperatorsStructurReachSet);
 
 \draw[to] (THoperatorsStructurReachSet) -- (equal);
 
 \draw[to] (equal) -- (THreachabilityTheorem);

 \draw[to] (THreachabilityTheorem) -- (d1Eqd2);
 \draw[to] (d1Eqd2) -- (DEFfunctionalEquivalence);
 \draw[to] (DEFfunctionalEquivalence) -- (FuncEq);

\end{tikzpicture}
\end{center}
\label{fig:roadmap_fun_equi}
\end{figure}
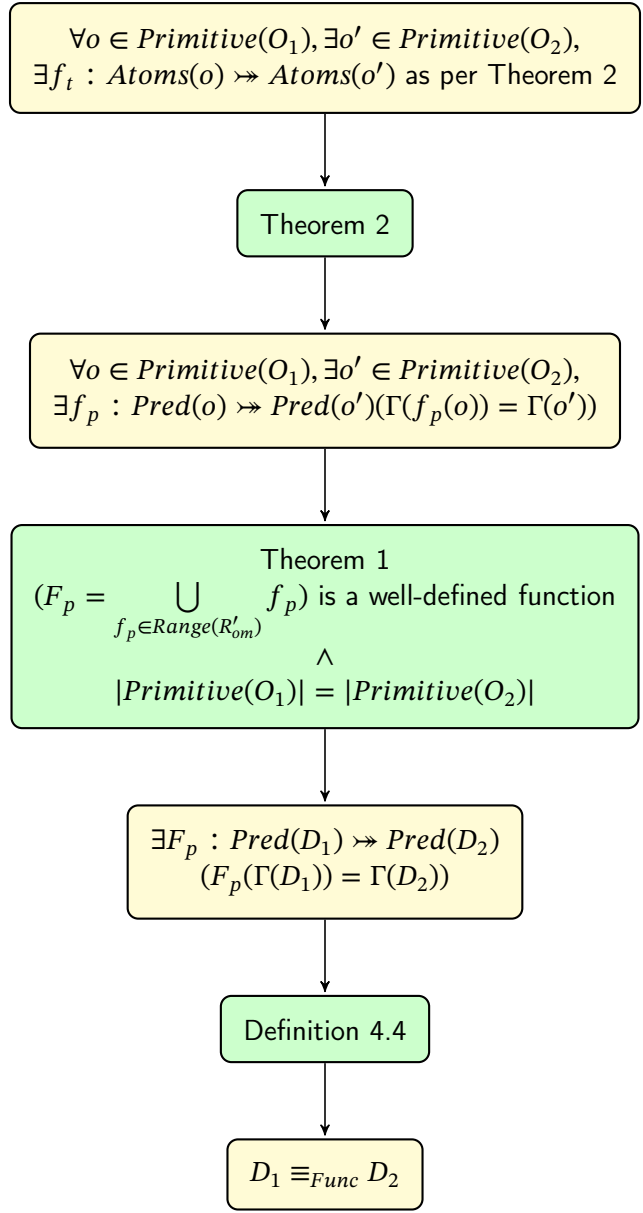

A flow chart that illustrates the proposed logical steps to prove the functional equivalence of two  simple planning domain models is depicted in \Cref{fig:roadmap_fun_equi}.

This section explained the required logical steps to prove or disprove the functional equivalence of simple planning domain models.
The following section lists theorems that form the theoretical foundation of our method.

\section{Theorems}\label{Theorems}

\subsection{Simple Domains Reachability Theorem}\label{sec:SimpleDomainsReachabilityTheorem}
TThe Domain Reachability Theorem reduces the relation between the reach sets of two simple domains \D1 and \D2 to relations between their primitive operators.
To prove the reach set of an image of \D1 is equal to the reach set of \D2, we have to prove that the reach set of an image of each primitive operator $o$ in $D_1$ is equal to the reach set of a primitive operator $o'$ in $D_2$ under a bijective mapping from the predicates of $o$ to the predicates of $o'$; both domains have an equal number of primitive operators, and all the mappings from the predicates of each primitive operator in $D_1$ to the predicates of each primitive operator in $D_2$ are consistent. 
We will define two additional concepts to simplify the explanation of this theorem. 

Let $\mathbb{F}$ be the set of all bijective mappings between predicates of equal arities from the predicates of every primitive operator in $D_1$ to the predicates of every primitive operator in $D_2$.
\begin{alignat}{1}
\mathbb{F} = \{ f_p | f_p: Predicates(o) \twoheadrightarrowtail Predicates(o') \ &\text{ where } o \in Primitive(O_1), o' \in Primitive(O_2) \nonumber \\
\text { and if } f_p(p) = p' \text{ then } Arity(p) = Arity(p') \}\nonumber 
\end{alignat}

Let $R_{OM}$ be a relation between primitive operators from $D_1$ and predicate mappings from $\mathbb{F}$. A primitive operator $o$ from $D_1$ is related to a mapping $f_p$ from $\mathbb{F}$ by $R_{OM}$ means there exists a primitive operator $o'$ from $D_2$ such that the reach set of $f_p(o)$ is equal to the reach set of $o'$.
$$ R_{OM} = \{ (o,f_p) \in Primitive(O_1) \times \mathbb{F} \ | \ \exists o' \in Primitive(O_2), \ \Gamma(f_p(o),Obj) = \Gamma(o',Obj) \}$$
This theorem is formalised as follows.

\begin{restatable}[Simple Domains Reachability Theorem]{theorem}{RT}
\label{th:SimpleDomainsReachabilityTheorem}
Consider a set of objects \textit{Obj}, two simple planning domain models, $D_1$ and $D_2$, a bijective function $F_p$ from the predicates of $D_1$ to the predicates of $D_2$ with equal arities, and the relation $R_{OM}$ that relates each primitive operator $o$ in $O_1$ to a bijective predicate mapping $f_p$ that makes the reach set of $f_p(o)$ equals to the reach set of a primitive operator from $O_2$. We have:
\begin{multline}
\Gamma(F_p(D_1),Obj) = \Gamma(D_2,Obj) \textbf{ iff } \exists R'_{om} \subseteq R_{OM}( Domain(R'_{om}) = primitive(O_1) \wedge \\ (F = \bigcup\limits_{f_p \in Range(R'_{om}) } f_p )\text{ is a well-defined function} \wedge |Primitive(O_1)| = |Primitive(O_2) |) 
\end{multline}
\end{restatable}
To simplify the proof of this theorem, we formalise the following lemmas.
\begin{restatable}{lem}{ReachSetOneOperatorSubsetUnionSEOperator}
\label{ReachSetOneOperatorSubsetUnionSEOperator}
Consider a set of objects \textit{Obj}, two simple planning domain models, $D_1$ and $D_2$, and a bijective function $F_p: P_1 \twoheadrightarrowtail P_2$. we have
$$\exists o \in Primitive(O_1), \forall O' \subseteq O_2(\Gamma(F_p(o)) \subseteq \bigcup _{o' \in O'}( o') \implies \exists se \in SE(O_2), \forall o' \in O' (o' \in se))$$
\end{restatable}

\begin{restatable}{lem}{PrimitiveNoOperatorNFELemma}
\label{PrimitiveNoOperatorNFELemma}
Consider a set of objects \textit{Obj}, two simple planning domain models, $D_1$ and $D_2$, and a bijective function $F_p: P_1 \twoheadrightarrowtail P_2$.
We have

\begin{equation}
\exists o \in Primitive(O_1), \forall Seq' \subseteq SEQ_2(\Gamma(F_p(o)) \not = \bigcup_{ seq' \in Seq'}(\Gamma(seq'))) \implies \Gamma(F_p(D_1),Obj) \not= \Gamma(D_2,Obj) \nonumber
\end{equation}

\end{restatable}

\begin{restatable}[Simple Domain Reachability Lemma]{lem}{RL}
\label{reachabilityLemma}
Consider a set of objects \textit{Obj}, two simple planning domain models, $D_1$ and $D_2$, and a bijective function $F_p: P_1 \twoheadrightarrowtail P_2$.
We have
\begin{multline}
\forall o \in Primitive(O_1), \exists o' \in Primitive(O_2) ( \Gamma(F_p(o),Obj) = \Gamma(o',Obj) \wedge \\
|Primitive(O_1)| = |Primitive(O_2) |) \iff \Gamma(F_p(D_1),Obj) = \Gamma(D_2,Obj)
\end{multline}
\end{restatable}

\Cref{th:SimpleDomainsReachabilityTheorem} is proven in \Cref{ProofOfTheoremRT} with the help of \Cref{PrimitiveNoOperatorNFELemma} and \Cref{reachabilityLemma}.
The proofs of these lemmas are provided in \Cref{ProofOfLemmas}.

\subsection{Operators Structure Reach Set Theorem}
Operators Structure Reach Set Theorem relates the reach sets equality of two operators to the condition that both should have the same structure.
This condition is represented by the existence of a special mapping between the atoms of the two operators.
This theorem is fundamental as it enables us to evaluate the equality of the reach sets of two operators without enumerating the elements of these two sets.
Note that it is impossible to enumerate the reach sets of two operators for any set of objects, as there are infinitely many sets of objects. This theorem is formalised as follows.
 
\begin{restatable}[Operators Structure Reach Set Theorem]{theorem}{OSRST}
\label{th:operatorsStructurReachSet}

Consider a set of objects \textit{Obj}, two operators $o$ and $o'$, and a bijective function $f_p: Predicates(o) \twoheadrightarrowtail Predicates(o')$ such that $f_p$ maps the predicates of $o$ with those of $o'$ with equal arities. 
Let $f_p(o)$ be the image of $o$ using $f_p$ to substitute the predicates of $o$ with those of $o'$ with equal arities.
For any set of objects, the reach set of $f_p(o)$ is equal to the reach set of $o'$ \textbf{iff} there exists a bijective mapping $f_t$ from the atoms of $o$ to the atoms of $o'$ such that $f_t$ and $f_t^{-1}$ satisfy the following conditions:

\begin{enumerate}

 \item Atoms in the preconditions, delete effects and add effects of one operator must be mapped to atoms in the preconditions, delete effects and add effects of the other operator, respectively;

 \item Atoms in one operator must be mapped to atoms in the other operator with equal arity;
 
 \item Atoms with the same predicate $p$ in one operator must be mapped to atoms with some predicate $p'$ in the other operator; and

 \item Atoms with a shared variable $v$ in one operator must be mapped to atoms with some shared variable $v'$ in the other operator such that the positions of $v$ and $v'$ in the parameters of the mapped atoms are equal.
\end{enumerate}
\normalfont Here we provide the formal form of the theorem. The part related to $f_t^{-1}$ has been omitted to avoid repetition.
$$ \forall Obj, \exists f_p: Predicates(o) \twoheadrightarrowtail Predicates(o') \ \text { where if } f_p(p) = p' \text{ then } Arity(p) = Arity(p') \ ($$
$$\ \Gamma(f_p(o),Obj) = \Gamma(o',Obj) \iff $$
$$\exists f_t: Atoms(o) \twoheadrightarrowtail Atoms(o') \ ( $$
$$ \forall t \in Pre(o) \ (\exists t' \in Pre(o') : f_t(t) = t') \ \wedge $$
$$ \forall t \in Del(o) \ (\exists t' \in Del(o') : f_t(t) = t') \ \wedge $$
$$ \forall t \in Add(o) \ (\exists t' \in Add(o') : f_t(t) = t') \ \wedge $$
$$ \forall t \in Atoms(o) \ (\exists t' \in Atoms(o') : f_t(t) = t' \wedge Arity(t) = Arity(t') ) \ \wedge $$
$$ \forall t_1,t_2 \in Atoms(o) \ ( Pred(t_1) = Pred(t_2) \rightarrow \exists t'_1,t'_2 \in Atoms(o') \ ( Pred(t'_1) = Pred(t'_2)$$
$$ \wedge \ (f_t(t_1) = t'_1 \wedge f_t(t_2) = t'_2 ) \vee (f_t(t_2) = t'_1 \wedge f_t(t_1) = t'_2 ) ) ) \ \wedge $$
$$ \forall t_1,t_2 \in Atoms(o), \forall v_1 \in Var(t_1), \forall v_2 \in Var(t_2) \ ( v_1 = v_2 \rightarrow $$
$$\exists t'_1,t'_2 \in Atoms(o'), \exists v'_1 \in Var(t'_1), \exists v'_2 \in Var(t'_2) \ ( v'_1 = v'_2$$
$$ \wedge \ ( (f_t(t_1) = t'_1 \wedge f_t(t_2) = t'_2 \wedge Position(v_1,t_1) = Position(v'_1,t'_1) \wedge Position(v_2,t_2) = Position(v'_2,t'_2))$$
$$ \vee \ (f_t(t_2) = t'_1 \wedge f_t(t_1) = t'_2 \wedge Position(v_2,t_2) = Position(v'_1,t'_1) \wedge Position(v_1,t_1) = Position(v'_2,t'_2)) ) ) ) ) $$

\end{restatable}

This theorem is proven in \Cref{ProofOfTheoremOSRST}.

We have introduced our approach in \Cref{sec:roadmap}.
Furthermore, we have formalised and proved all required theorems and lemmas in this section.
In the following section, we will provide an overview of our method. 
\section{Validating the functional equivalence of simple domains.}\label{sec:ValidatingFESimpleDomains}
To prove the functional equivalence of two simple planning domain models, we have to show that the reach set of one domain is equal to the reach set of the other domain under a bijective predicate mapping as per \Cref{Def:functional-equivalence}. 
The logical steps required to prove the reach set of an image of a domain $D_1$ is equal to the reach set of a domain $D_2$ under a bijective predicate mapping are depicted in \Cref{fig:roadmap_fun_equi}.
These logical steps motivate developing this search task: 
\begin{task}\label{SearchTask}
\textit{Consider two simple planning domain models $D_1$ and $D_2$. If $D_1$ and $D_2$ have an equal number of primitive operators, find a bijective mapping from the atoms of each primitive operator $o$ in $D_1$ to the atoms of a primitive operator $o'$ from $D_2$, such that these atom mappings are consistent with regards to the mapping of their predicates and respect the conditions specified in \Cref{th:operatorsStructurReachSet}}.
\end{task}
The first step to proving the functional equivalence of two simple planning domain models \D1 and \D2 is to check if the two domains have an equal number of primitive operators according to the requirement of \Cref{th:SimpleDomainsReachabilityTheorem}.
Another reason to find primitive operators is that in the next step we need to find a mapping from the atoms of each primitive operator in \D1 to the atoms of a primitive operator in \D2.
Therefore, we must find and remove all other types of operators from \D1 and \D2 as we are only concerned with primitive operators.
Any operator from \D1 and \D2 cannot be a split operator because \D1 and \D2 are simple domains and simple domains do not have split operators.
The next step is to find and remove macro operators from \D1 and \D2.
This task is explained in \Cref{sec:checkMacro}.
The renaming non-macro operators can be either split-macros or primitive.
\Cref{sec:checkPrimitive} describes how to check if the renaming non-macro operators are primitive or not.
After removing all types of operators apart from primitive operators from \D1 and \D2, we can check the number of primitive operators in these two domains.
If the two domains do not have an equal number of primitive operators, then according to \Cref{th:SimpleDomainsReachabilityTheorem}, the reach sets of the two domains are not equal under any predicate mapping.
Hence, \D1 and \D2 are not functionally equivalent.
On the other hand, if the two domains have an equal number of primitive operators, then we can validate the functional equivalence of \D1 and \D2 according to \Cref{th:SimpleDomainsReachabilityTheorem} by finding a mapping $F_p$ from the predicates of \D1 to the predicates of \D2 such that this mapping can make the reach set of each primitive operator from \D1 equals to the reach set of a primitive operator in \D2.

To find a predicate mapping $F_p$ that satisfies the conditions of \Cref{th:SimpleDomainsReachabilityTheorem}, we have to find bijective mappings from the atoms of each primitive operator $o$ in $D_1$ to the atoms of a primitive operator $o'$ in $D_2$, such that these atom mappings are consistent with regards to their predicates and satisfy the conditions of \Cref{th:operatorsStructurReachSet}.
We start the search for the atom mappings that satisfy the conditions of \Cref{th:operatorsStructurReachSet} from a bigger set of atom mappings which are only constrained by the first condition and that satisfy part of the requirement of the fourth condition of this theorem.
These two criteria are chosen because checking if an atom mapping satisfies them is a simple numerical comparison.

To find mappings that satisfy all the conditions of this theorem, we narrow down the mappings found in the previous step by applying the remaining constraints of \Cref{th:operatorsStructurReachSet}. 

This search problem is presented in \Cref{subsec:FindingConsistentAtomMappings} and detailed in \Cref{sec:SMT}.
This approach necessitates that for every operator $o$ in $D_1$, we find every operator $o'$ from $D_2$ that supports the existence of mappings from the atoms of $o$ to the atoms of $o'$ that satisfy the first condition and part of the requirement of the fourth condition of \Cref{th:operatorsStructurReachSet}.
This requirement is essential, so we do not exclude any potential mappings which can possibly prove the functional equivalence between the two domains.

Finding every operator $o'$ in \D2 with atoms that can be mapped to the atoms of an operator $o$ from \D1 as per the first condition of \Cref{th:operatorsStructurReachSet} can be performed by directly comparing the number of atoms in the preconditions, add and delete effects of the operators $o$ and $o'$; basically, $o$ and $o'$ should have an equal number of preconditions, add and delete effects.
In addition, $o$ and $o'$ must have the same number of parameters; if these two operators have a different number of variables, then no mapping between the atoms of $o$ and $o'$ will satisfy the fourth condition of \Cref{th:operatorsStructurReachSet}.
This task is further explained in \Cref{subsec:FindingPFEPOs}.
If we find mappings that satisfy all conditions of \Cref{th:operatorsStructurReachSet}, then these mappings are guaranteed to produce predicate mappings that respect the second condition of \Cref{th:SimpleDomainsReachabilityTheorem}.

To ensure the mappings found by our method can produce predicate mappings which satisfy the third condition of \Cref{th:SimpleDomainsReachabilityTheorem}, we further constrain our search problem to find only mappings that are consistent with regard to the mapping of atoms with shared predicates. 
This consistency constraint is explained in \Cref{subsec:SMTThirdCondition} and formalised in \Cref{eq:SPC}.

Since the two domains are checked to have an equal number of primitive operators, the first condition of \Cref{th:SimpleDomainsReachabilityTheorem} is satisfied.
Therefore, if we find a mapping $F_P$ from the predicates of \D1 to the predicates of \D2 such that $F_p$ satisfies the second and third conditions of \Cref{th:SimpleDomainsReachabilityTheorem}, then we prove \D1 and \D2 are functionally equivalent.

\subsection{Finding Potentially Functionally Equivalent Operators (PFEOs) from $D_2$ for each operator $o$ from $D_1$} \label{subsec:FindingPFEPOs}
The first condition of \Cref{th:operatorsStructurReachSet} says \emph{atoms in the preconditions, delete effects and add effects of an operator $o$ from domain $D_1$ must be mapped to atoms in the preconditions, delete effects and add effects of the operator $o'$ from domain $D_2$respectively.}
Hence, for a mapping from the atoms of $o$ to the atoms of $o'$ to satisfy the first condition of \Cref{th:operatorsStructurReachSet}, the operator $o$ must have a number of atoms in its precondition, add effects and delete effects that is equal to the number of atoms in the precondition, add effects and delete effects of the operator $o'$. These conditions are captured in the following equations. 
\begin{alignat}{1}
|Pre(o)| &= |Pre(o')| \label{eq:Preo=Preo'} \\ 
|Add(o)| &= |Add(o')| \label{eq:Addo=Addo'} \\ 
|Del(o)| &= |Del(o')| \label{eq:Delo=Delo'}
\end{alignat}

Furthermore, since we are looking for bijective mappings from the atoms of the operator $o$ to the atoms of each operator $o'$ in \D2, $o$ and $o'$ must have an equal number of atoms.
Note that the add effects of an operator cannot appear as delete effects or preconditions in any valid operator.
Therefore, there is no intersection between the add and delete effects and between the add effects and the preconditions of any operator.
Moreover, the delete effects of an operator must be a subset of the preconditions of that operator. 
Therefore, we can say the set of atoms of an operator is the union of the atoms of its preconditions and the atoms of its add effects.
Hence, \Cref{eq:Preo=Preo'} and \Cref{eq:Addo=Addo'}  guarantee the existence of a bijective mapping from the atoms of an operator $o$ to the atoms of an operator $o'$.

A necessary but not sufficient condition for an atom mapping to satisfy the fourth constraint of \Cref{th:operatorsStructurReachSet} is for $o$ and $o'$ to have an equal number of variables. 
\begin{alignat}{1}
|Var(o)| = |Var(o')| \label{eq:VariablesO=O'}
\end{alignat}
 
We group the quantities referenced in \Cref{eq:Preo=Preo',eq:Addo=Addo',eq:Delo=Delo',eq:VariablesO=O'} in a tuple which we call operator signature.
\begin{defn} \label{Def:OperatorSignature}
The 
\textbf{\emph{signature}} of an operator $o$, $\operatorname{OpSig}(o)$, is defined based on the number of atoms in its different parts as follows.
\begin{alignat}{1}
\operatorname{OpSig}(o) = (|Pre(o)|,|Add(o)|,|Del(o)|,|Var(o)|)
\end{alignat}
\end{defn}
The second stage of our method is to find all operators from $D_2$ that have the same signature as each operator in $D_1$.
\begin{defn} \label{Def:PFEO}
We call an operator $o$ in a planning domain model $D_2$ a \textbf{\emph{Potentially Functionally Equivalent operator (PFEO)}} of an operator $o$ from a planning domain model $D_1$ if $o$ and $o'$ have the same signature.
\end{defn}
As such, the set of PFEOs from a domain $D_2$ of an operator $o$ is defined as follows.
\begin{alignat}{1}
\operatorname{PFEOs(o)} = \{ o': o' \in O_2, \operatorname{OpSig}(o') = \operatorname{OpSig}(o) \} \label{eq:PFEOs}
\end{alignat}
Finding the $\operatorname{PFEOs}$ from $D_2$ for each operator $o$ in $D_1$ provides us with sets of bijective mappings from the atoms $o$ to the atoms of each of its PFEOs from \D2.
These mappings are bijective and are guaranteed to satisfy the first condition and part of the requirement of the fourth condition of \Cref{th:operatorsStructurReachSet}.
Starting from these sets, we can check the other conditions in the second stage.

\subsection{Finding consistent atom mappings that satisfy the conditions of \Cref{th:operatorsStructurReachSet}} \label{subsec:FindingConsistentAtomMappings}
This section will discuss how to find a consistent atom mapping that satisfies the conditions of \Cref{th:operatorsStructurReachSet}.
In the previous section, we have seen how restricting the mappings of the atoms of the operators of $D_1$ to atoms of their PFEOs guarantees these mappings are bijective and satisfy the first condition of \Cref{th:operatorsStructurReachSet}.
So, we need to find a consistent atom mapping that respects the restriction imposed by the PFEOs and satisfies the remaining conditions of \Cref{th:operatorsStructurReachSet}.
We encode the problem of finding such an atom mapping as a Satisfiability Modulo Theories (SMT) problem.
An SMT problem consists of a set of decision variables and a set of constraints.
A solution to an SMT problem is an assignment of the decision variables that satisfies the constraints of this SMT problem.
The following sections explain the encoding of our SMT problem and its solution. 
\subsubsection{Description of the decision variables of our SMT problem}\label{sec:descriptionOfDecisionVariableMapping}
The first step in encoding the problem of finding a consistent atom mapping that satisfies the conditions of \Cref{th:operatorsStructurReachSet} as an SMT problem is to define the decision variables of this problem.
We must analyse our problem carefully to choose appropriate decision variables for our encoding.
Note that the fourth condition of \Cref{th:operatorsStructurReachSet} constraints the mapping of atoms based on their variables. 
Therefore, to capture this condition, we need decision variables that can express constraints at the level of the variables of the atoms of operators.
Decision variables at the level of the variables of the atoms mean the SMT problem will reason at the level of variables.
The solution to our SMT problem is a variable mapping.
This mapping maps the variables of the atoms of the operators in $D_1$ to the variables of the atoms of some of the PFEOs from $D_2$.
To guarantee the variable mappings produce consistent atom mappings, we have to add a set of constraints to ensure the variables of one atom in an operator in a domain are mapped to the variables of some atom in an operator in the other domain.
With these additional constraints, the variable mappings will be merely another representation of atom mappings.

To map variables of atoms of operators, we have to first agree on how to define unique variables.
We define the canonical signature of variables by concatenating the variable name with some features of its atom and its operator as follows.
\begin{defn} \label{Def:variable-signature}

The \emph{variable signature} of a variable consists of a concatenation of the following items:
\begin{enumerate}
 \item the name of the given variable,
 \item the arity of the atom of the given variable,
 \item the name of the predicate of the atom of the given variable,
 \item the name of the operator of the atom of the given variable,
 \item a concatenation of all variables of the atom of the given variable,
 \item and the position of the variable in the parameters of the atom of the given variable.
\end{enumerate}
\end{defn}
We refer to the signature of a variable $v$ by $\operatorname{VarSig}(v)$, and to reference a part of the signature of the variable $v$ we subscript $\operatorname{VarSig}(v)$ with the number of the part from the above list.
For example, the signature of the variable ?r in the atom (at ?r ?x) in the operator "move" is "?r 2 at move ?r?x 1"; the arity of the atom of the variable "?r" is $\operatorname{VarSig}_2(at\ ?r\ ?x)$ which equals 2. 
Two variables are different in a domain if their signatures differ by at least one part.
For instance, the four variables of the atoms (at ?r ?x) and (at ?r ?y) in the same operator "move" are different.
The variables with the signatures "?x 2 at move ?r?x 2" and "?y 2 at move ?r?y 2" are clearly different as they differ by the variable symbol.
Mo rover, the two variables with the signatures "?r 2 at move \textbf{?r?x} 1" and "?r 2 at move \textbf{?r?y} 1" are also different.
They differ by $\operatorname{VarSig}_5$, i.e. the item number five of the signature.
These two variables are considered different because they are parts of two different atoms.
The proposed variable signature provides us with enough information about the variables of the atoms of the operators in order to implement the conditions of \Cref{th:operatorsStructurReachSet} in addition to the condition of consistent mappings.
Since our problem is encoded at the level of variables, we call it a variable mapping SMT problem.
The next subsection describes the solutions to the variable mapping SMT problem.
\subsubsection{Description the solutions of variable mapping SMT problem}
The solution to our variables mapping SMT problem is a bijective function $d: V_1 \rightarrow N $.
This function relates the variables of the atoms of $D_1$ to a set of natural numbers ($N \subset \mathbb{N}$) such that the members of the set $N$ represent the variables of the atoms of $D_2$.
The natural numbers in this set are linked to the variables of $D_2$ by the inverse of the bijective function $b : V _2 \rightarrow N$.
The composition of functions $d$ and $b^{-1}$ produces a mapping function that maps the variables of the atoms of \D1 to the variables of the atoms of \D2.
\begin{alignat}{1}
\forall v \in V_1, \exists v' \in V_2 \ ( v' = b^{-1} \circ d(v)) \label{eq:FromVtoV'}
\end{alignat}
The function $b$ can be defined arbitrarily.
For example, sorting the variables of the atoms of $D_2$ according to some criteria on the signature of these variables defines a function $b$.
Actually, we are not sorting the variables of the atoms of $D_2$, but we are sorting the list of their signatures.
In this case, $b$ links every variable or every variable signature of the atoms of $D_2$ to the index of this variable signature in the sorted list of the variable signatures.
So, the function $b$ is the function that returns the index of a variable signature in a sorted list of variable signatures.
After we have defined the function $b$, we need to find the function $d$ in order to define the mapping from the variables of the atoms of $D_1$ to the variables of the atoms of $D_2$. 
The function $d$ can be explained as a relation between the variables of the atoms of $D_1$ and the natural numbers in the domain of the function $b$.
This relation can be described with the help of decision variables that are related to the variables of the atoms of $D_1$.
The value of a decision variable is a member of the set of the natural numbers that forms the range of $d$, which is also the domain of $b$.
As such, every assignment to a decision variable forms a map from a variable of an atom of $D_1$ to a variable of an atom of $D_2$. 

In this section, we have introduced the variable mapping SMT problem.
The following section explains this problem and lists the SMT constraints that capture the conditions of \Cref{th:operatorsStructurReachSet} in addition to the condition of consistent domains predicate mapping. 
\section{Variable Mapping SMT Problem}\label{sec:SMT}
The variable mapping SMT problem consists of a set of decision variables and a set of constraints on these decision variables.
Each decision variable is associated with a variable from the atoms of the operators of $D_1$.
A solution to this problem is an assignment of the decision variables that satisfies the problem\textquotesingle s constraints.
To solve this problem, we sort the variables of the atoms of the operators of $D_2$.
Then, we define the function $b$ to store the order of the sorted variables.
After that, we compile the set of constraints that captures the atom mapping consistency criterion and the conditions of \Cref{th:operatorsStructurReachSet}.
The compilation of these constraints is explained in the following sections.
Then we task the Z3 SMT solver~\cite{de2008z3} to solve our variable mapping SMT problem.
The solution found by the SMT solver assigns a natural number to each decision variable $d(v)$.
Thus, it defines the function $d$.
Then a mapping from the variables of $D_1$ to the variables of $D_2$ is extracted from the functions $b$ and $d$ with the help of \Cref{eq:FromVtoV'}.
For each variable $v$ in $D_1$, 
$d(v)$ points through $b^{-1}$ to a variable $v'$ in the atoms of $D_2$. 
In other words, the function $d$ maps $v$ to $v'$.
For example, if $b(v_5^{'}) = 5$, then $ d(v_1) = 5$ means the variable $v_1$in $D_1$ is mapped to the variable $v_5^{'}$ in $D_2$.

To formalise constraints for the third and fourth conditions of \Cref{th:operatorsStructurReachSet}, we need the variables of $D_2$ to be sorted in two specific orders, one order for each condition.
On the other hand, the way we order the variables of $D_2$ is irrelevant for compiling the first two conditions of \Cref{th:operatorsStructurReachSet} as long as the variables are sorted in a known way.
Therefore, we will discuss the SMT constraints of the first two conditions before we explain the idea of the ordered variables when we introduce the constraints of the third and fourth conditions.

The following three subsections explain the SMT constraints that capture the conditions of \Cref{th:operatorsStructurReachSet}. 
\subsection{The SMT constraints of the first and second conditions, and the second part of the fourth condition of \Cref{th:operatorsStructurReachSet} }\label{sec:12-2of4}
The first condition of \Cref{th:operatorsStructurReachSet} requests preconditions, add effects and delete effects of $o$ are mapped to their counterparts in $o'$.
This condition is captured by contrasting the decision variables of the variables of the preconditions of $o$ to the values of the variables of the preconditions of $o'$. The same constraints are defined for the add and delete effects.
The second condition of \Cref{th:operatorsStructurReachSet} mandates that only atoms of the same arity are mapped to each other.
This condition is satisfied by requesting the second parts of the signatures of the mapped variables to be equal to each other, i.e. any mapped variables must belong to atoms of equal arity. 
We know that the fourth condition of \Cref{th:operatorsStructurReachSet} necessitates the positions of mapped variables in the parameters of their respective atoms to be equal.
We can impose this part of the fourth condition by requesting the sixth parts of the signatures of the mapped variables to be equal to each other, i.e. any mapped variables must be in the same positions in the parameters of their respective atoms.
The second condition and the second part of the fourth condition are simple. We can use explicit equality constraints on the arity of the atoms of the variables and the position of the variables in the parameters of their atoms because these properties are natural numbers. 

Our constraints are defined using nested loops.
For each operator $o$ in \D1, for each operator $o' \in \operatorname{PFEO}(o)$, for each variable in the atoms of the preconditions, add effects and delete effect of $o$, for each variable in the atoms of the preconditions, add effects and delete effect of $o'$, we will add a constraint.
The constraints at the bottom level are defined as sets of constraints.
\subsubsection{Constraints grouping at the level of the variables of an operator $o$}
One set of constraints for mapping the variables of the atoms in each part of the operators: preconditions, add effects and delete effects.
These sets are called PreConst, AddConst, and DelConst, respectively.
These sets constrain the assignment of the decision variable of a given variable $v$ to the indices of variables in the preconditions, add effects, and delete effects of an operator $o'$.
As such, we represent them with $\operatorname{PreConst}(v,o')$, $\operatorname{AddConst}(v,o')$, and $\operatorname{DelConst}(v,o')$. 
These sets are defined as follows.
\begin{alignat}{1}
\operatorname{PreConst}(v,o') = &\{(d(v) == b(v')) \ | \ \forall p' \in Pre(o'), \label{eq:PreConst}\\
& \forall v' \in \operatorname{variables}(p')( \operatorname{VarSig}_2(v) = \operatorname{VarSig}_2(v') \wedge \operatorname{VarSig}_6(v) = \operatorname{VarSig}_6(v')) \} \nonumber
\end{alignat}
\begin{alignat}{1}
\operatorname{AddConst}(v,o') = &\{(d(v) == b(v')) \ | \ \forall p' \in Add(o'), \label{eq:AddConst}\\
& \forall v' \in \operatorname{variables}(p')( \operatorname{VarSig}_2(v) = \operatorname{VarSig}_2(v') \wedge \operatorname{VarSig}_6(v) = \operatorname{VarSig}_6(v')) \} \nonumber
\end{alignat}
\begin{alignat}{1}
\operatorname{DelConst}(v,o') = &\{(d(v) == b(v')) \ | \ \forall p' \in Del(o'),\label{eq:DelConst} \\
& \forall v' \in \operatorname{variables}(p')( \operatorname{VarSig}_2(v) = \operatorname{VarSig}_2(v') \wedge \operatorname{VarSig}_6(v) = \operatorname{VarSig}_6(v')) \} \nonumber
\end{alignat}

Note that the scope of $v$ and $o'$ in the definition of the above sets are not defined yet.
These scopes will be defined as we explain the top levels of our constraints.
The main idea is for $\operatorname{PreConst}(v,o')$, $v$ will be quantified over all variables of all preconditions of an operator $o$, and $o'$ is quantified over all PFEOs of the operator $o$.
In its turn, the operator $o$ will be quantified over all the operators of the domain $D_1$. 
Another important note is that the constraints that will be made from $\operatorname{PreConst}(v,o')$ do not just constrain the mapping of a variable $v$ to the variables of the preconditions of $o'$ but also impose the second condition and the second part of the fourth condition of \Cref{th:operatorsStructurReachSet}.
Of course, the above notes applies to $\operatorname{AddConst}(v,o')$ and $\operatorname{DelConst}(v,o')$ as well.

If the reach set of $o'$ is equal to the reach set of the operator $o$ that has the variable $v$ in one of its preconditions, then one member of the $\operatorname{PreConst}(v,o')$ must be true.
To achieve this requirement, we need a disjunction of the members of $\operatorname{PreConst}(v,o')$ as represented in the following equation, let us call this disjunction $\operatorname{PreConst_{or}}(v,o')$.
\begin{alignat}{1}
\operatorname{PreConst_{or}}(v,o') = \bigvee\limits_{\forall \operatorname{const} \in \operatorname{PreConst}(v,o') } \ \operatorname{const} \label{eq:PreConst-or}
\end{alignat}

Similarly, we call the disjunctions of the add effects $\operatorname{AddConst_{or}}(v,o')$ and delete effects $\operatorname{DelConst_{or}}(v,o')$. These disjunctions are expressed in \Cref{eq:AddConst-or} and \Cref{eq:DelConst-or}.
\begin{alignat}{1}
\operatorname{AddConst_{or}}(v,o') = \bigvee\limits_{\forall \operatorname{const} \in \operatorname{AddConst}(v,o') } \ \operatorname{const}\label{eq:AddConst-or} 
\end{alignat}
\begin{alignat}{1}
\operatorname{DelConst_{or}}(v,o') = \bigvee\limits_{\forall \operatorname{const} \in \operatorname{DelConst}(v,o') } \ \operatorname{const}\label{eq:DelConst-or}
\end{alignat}

We know the disjunction $\operatorname{PreConst_{or}}(v,o')$ must be used to constrain the mapping of variables from the preconditions of an operator $o$.
As such, the variable $v$ in $\operatorname{PreConst_{or}}(v,o')$ must be quantified over all variables of the preconditions of an operator $o$.
This will produce a disjunction $\operatorname{PreConst_{or}}(v,o')$ for every variable $v$ in the preconditions of an operator $o$.
These disjunctions must be combined in a conjunction to enforce the decision variables of every variable in the preconditions of $o$ to be assigned to a value.
We call this conjunction $\operatorname{PreConst_{and}}(o,o')$. 
\begin{alignat}{1}
\operatorname{PreConst_{and}}(o,o') = \bigwedge \limits_{\forall p \in pre(o), \forall v \in \operatorname{variables}(p)} \ \operatorname{PreConst_{or}}(v,o')\label{eq:PreConst-and}
\end{alignat}

Similar to the conjunction $\operatorname{PreConst_{and}}(o,o')$ , we call the conjunction of the disjunctions of the constraints of the add effects as $\operatorname{AddConst_{and}}(o,o')$ and the conjunction of the disjunctions of the delete effects as $\operatorname{DelConst_{and}}(o,o')$. These conjunctions are expressed in \Cref{eq:AddConst-and} and \Cref{eq:DelConst-and}.
\begin{alignat}{1}
\operatorname{AddConst_{and}}(o,o') = \bigwedge \limits_{\forall p \in pre(o), \forall v \in \operatorname{variables}(p)} \ \operatorname{AddConst_{or}}(v,o')\label{eq:AddConst-and}
\end{alignat}
\begin{alignat}{1}
\operatorname{DelConst_{and}}(o,o') = \bigwedge \limits_{\forall p \in pre(o), \forall v \in \operatorname{variables}(p)} \ \operatorname{DelConst_{or}}(v,o')\label{eq:DelConst-and}
\end{alignat}

Note that the scope of $v$ is now defined.
Also, note that the scopes of the operator $o$ and the macro $o'$ are still not defined yet.
These scopes will be defined shortly as we progress in explaining our constraints.

\subsubsection{Bijective mapping Constraints}
We are looking for a bijective mapping between the atoms of the two domains. Hence, each variable must be mapped to just one variable from the other domain.
However, our constraints, which are conjunctions of disjunctions of the assignments of decision variables to natural numbers, can be satisfied by assigning two design variables with the same value.
For example, assume for a variable $v$ and a macro $o'$ we have: 
\begin{alignat*}{1}
\operatorname{PreConst_{and}}(o,o') = \{((d(v_1) == 1) \vee (d(v_1) == 2) \vee (d(v_1) == 3)) \wedge \\ 
((d(v_2) == 1) \vee (d(v_2) == 2) \vee (d(v_2) == 3)) \wedge \\
 ((d(v_3) == 1) \vee (d(v_3) == 2) \vee (d(v_3) == 3)) \}
\end{alignat*}
Then, the assignments $d(v_1) == 1$, $d(v_2) == 1$, and $d(v_3) == 1$ satisfy the conjunction $\operatorname{PreConst_{and}}(v,o')$.
Such cases should be avoided to ensure the produced mapping is a bijective one. 
Thus, we need to add more constraints to guarantee that the values of the decision variables are unique.
This requirement is also applicable to the decision variables for the variables in the add and delete effects.
As such, the uniqueness constraints are defined for the decision variables of all variables as follows.
\begin{alignat}{1}
\operatorname{UniqeVariables}(O_1) = \{ \operatorname{Distinct}(d(v)) : \forall v \in Var(O_1) \} \label{eq:Uniqevariables}
\end{alignat}
Where the constraint \textit{Distinct} is defined as follows.
\begin{alignat}{1}
\forall v, v' \in V \ ( \operatorname{Distinct}(d(v)) \rightarrow d(v) \not = d(v'))
\end{alignat}

\subsubsection{Constraints grouping at the level of the PFEOs of an operator $o$}
Each of the previous conjunctions $\operatorname{PreConst_{and}}(o,o')$, $\operatorname{AddConst_{and}}(o,o')$, and $\operatorname{DelConst_{and}}(o,o')$ imposes partial mapping from the variables of an operator $o$ to the variables of a macro $o'$. 
The conjunction of these conjunctions demands all decision variables of the variables in the preconditions, add effects and delete effects of $o$ to be assigned to the indices of variables in the preconditions, add effects and delete effects of $o'$.
Let us call the conjunction of these conjunctions $\operatorname{OMConst}(o,o')$. 
\begin{alignat}{1}
\operatorname{OMConst}(o,o') = \operatorname{PreConst_{and}}(o,o') \wedge \operatorname{AddConst_{and}}(o,o') \wedge \operatorname{DelConst_{and}}(o,o') \label{eq:OMConst}
\end{alignat}

We know the variables of the operator $o$ must be mapped to the variables of one of its PFEOs. 
This requirement is expressed by the disjunction of $\operatorname{OMConst}(o,o')$ for every PFEO of $o$
.
We call this disjunction $\operatorname{OConst}(o)$.
\begin{alignat}{1}
\operatorname{OConst}(o) = \bigvee_{m \in \operatorname{PFEO}(o)} \operatorname{OConst}(o,o') \label{eq:OConst}
\end{alignat}

\subsubsection{Constraints grouping at the level of the operators of $D_1$}
Now that we have defined the scope of $o'$, we are left with defining the scope of $o$. 
For the reach set of $D_1$ to be a subset of the reach set of $D_2$ under a predicate bijective mapping, we need to find a bijective mapping from the atoms of each operator of $D_1$ to the atoms of a macro from $D_2$ as per the logical steps depicted in \Cref{fig:roadmap_fun_equi}.
Therefore, the variables of every operator $o$ from the operators of the domain $D_1$ must be mapped to the variables of one of its PFEOs.
This demand is expressed by the conjunction of $\operatorname{OConst}(o)$ for every $o$ of $O_1$.
We call this disjunction $\operatorname{DConst}(O_1)$.
\begin{alignat}{1}
\operatorname{DConst}(O_1) = \bigwedge_{o \in O_1} \operatorname{OConst}(o) \label{eq:DConst}
\end{alignat}

\subsubsection{The constraints of the range of the decision variables}
The domain of the function $d$ is $N \subset \mathbb{N}$ where $N = |\operatorname{variables}(O_1)|$.
Therefore, to ensure the values of the decision variables of $D_1$ are within the range of the function $d$, we need the following set of constraints. 
\begin{alignat}{1}
\operatorname{VRange}(O_1) = \{ (1 \leq d(v) \leq |\operatorname{variables}(O_1))| : \forall v \in \operatorname{variables}(O_1) \} \label{eq:VRange}
\end{alignat}
\subsection{The SMT constraints of the third condition of \Cref{th:operatorsStructurReachSet} }\label{subsec:SMTThirdCondition}
The third condition of \Cref{th:operatorsStructurReachSet} states that the atoms that share the same predicate in one domain must be mapped to atoms with one predicate in the other domain.
In our variable mapping SMT problem, we need constraints to guarantee that the variables that belong to atoms with a shared predicate are mapped to variables of atoms that share one predicate in the other domain. 
This kind of constraint relates the assignments of two variables to each other.
To compile such constraints, we have to sort the list of variable signatures of the variables of $D_2$ according to the predicate name part of the variable signature ($\operatorname{Sig}_3$).
Then we will have the variables of $D_2$ that belongs to the atoms that share the same predicate in adjacent places in the sorted list of variable signatures. 
We have to ensure the decision variables of the variables of the atoms of $D_1$ that have the same predicate are assigned to the indices of variables of atoms of $D_2$, which share one predicate.
Therefore, we have to constrain the assignments of these decision variables such that the difference between the values of these decision variables is less than a specific value. 
We will call this specific value the spacing value (Sv) for the reason that will become clearer shortly.
In other words, we have to ensure that the decision variables of the variables of $D_1$ that belong to atoms that share the same predicate are assigned to adjacent indices in the list of sorted variable signatures of the atoms of $D_2$.

So far, we have arranged the variable signatures in such a way that the signature of the variables of the atoms that share the same predicates are adjacent. 
However, we also have to ensure the signatures of the variables that belong to atoms with different predicates are not adjacent, i.e. separated.
Therefore, we have to create buffers between the groups of the signatures of the variables that belong to atoms with different predicates.
The size of these buffers is the spacing value Sv.
We propose the size of these buffers to be greater than the cardinality of the largest group of the variables of $D_2$ that belong to atoms with one predicate.

We refer to the spacing value of a list of signatures of variables which are sorted and spaced according to the predicate name of the atoms of these variables as $\operatorname{Sv}_p$.
To ensure the decision variables associated with the variables of atoms with the same predicate in $D_1$ are mapped to variables in $D_2$ that belong to atoms with one predicate from the predicates of $D_2$, the values of the decision variables of two variables in $D_1$ that belong to atoms with the same predicate must not differ by more than the spacing value $\operatorname{Sv}_p$.
We call the set of constraints that captures these requirements as Same Predicate Constraints, $\operatorname{SPC}(O_1)$, which is formally defined as follows.
\begin{alignat}{1}
\operatorname{SPC}(O_1) = \{ (|d(v_1) - d(v_2)| \leq \operatorname{Sv}_p) : \ \forall v_1, v_2 \in \operatorname{variables}(O_1)( \operatorname{VarSig}_3(v_1) = \operatorname{VarSig}_3(v_2)) \} \label{eq:SPC}
\end{alignat}

\subsection{The SMT constraints of the first part of the fourth condition of \Cref{th:operatorsStructurReachSet} }
The Fourth condition of \Cref{th:operatorsStructurReachSet} states that the atoms that share the same variable $v$ in one domain must be mapped to atoms with one variable $v'$ and these mapped atoms must have $v$ and $v'$ in the same position in the parameters of their atoms. 
The second part of this condition is compiled to constraints along with the first and second conditions in \Cref{sec:12-2of4}.
In this section, we will explain the compilation of the first part of the fourth condition into SMT constraints as per our SMT problem representation.
We need constraints to guarantee the solution to our SMT problem, which is a variable mapping that satisfies this condition.
Similar to the constraints of the third condition, the constraints of the fourth condition also relate the assignments of two variables to each other.
To compile such constraints, we also have to sort the list of the variable signatures of the variables of $D_2$.
However, this time we will sort this list differently. 
We know the fourth condition is concerned with the variables of the atoms of the two domains.
Therefore, the constraints that will capture this condition will have to deal with the variable names. 
Thus, we sort the list of the signature of the variables of $D_2$ according to the variable name part of the variable signature ($\operatorname{Sig}_1$).
Then we will have the variables of $D_2$ that belongs to the atoms that share the same variable in adjacent places in the sorted list of the variable signatures. 
We have to ensure the decision variables of the variables of the atoms of $D_1$ that have the same variables are assigned to the indices of variables of atoms of $D_2$, which share one variable.
Therefore, we have to constrain the assignment of these decision variables such that the difference between the values of these decision variables is less than a specific value. 
Similar to the third condition, we will call this specific value the spacing value Sv.
In other words, we have to ensure that the decision variables of the variables of $D_1$ that belong to atoms that share the same variable are assigned to adjacent indices in the list of sorted variable signatures of the atoms of $D_2$.

So far, we have arranged the variable signatures so that the signatures of the variables of the atoms that share the same variables are adjacent. 
Moreover, we also have to ensure the signatures of the variables that belong to atoms with different variables are not adjacent, i.e. separated.
Therefore, we have to create buffers between the groups of the signatures of the variables that belong to atoms with different variables.
The size of these buffers is the spacing value Sv. 
We propose the size of these buffers to be greater than the cardinality of the largest group of the variables of $D_2$ that belong to atoms with one variable.
We refer to the spacing value of a list of signatures of variables that are sorted and spaced according to the variable name of the atoms of these variables as $\operatorname{Sv}_v$.
To ensure the decision variables associated with the variables of atoms with the same variable in $D_1$ are mapped to variables in $D_2$ that belong to atoms with one variable in $D_2$, the values of the decision variables of two variables in $D_1$ that belong to atoms with the same variable must not differ by more than the spacing value $\operatorname{Sv}_V$.
We call the set of constraints that capture these requirements as Same Variable Constraints, $\operatorname{SVC}(O_1)$, which is formally defined as follows.
\begin{alignat}{1}
\operatorname{SVC}(O_1) = \{ (|d(v_1) - d(v_2)| \leq \operatorname{Sv}_V) : \ \forall v_1, v_2 \in \operatorname{variables}(O_1)( \operatorname{VarSig}_1(v_1) = \operatorname{VarSig}_1(v_2)) \} \label{eq:SVC}
\end{alignat}

In this section, we have explained the variable mapping SMT problem and listed the SMT constraints that capture the conditions of \Cref{th:operatorsStructurReachSet} in addition to the condition of consistent domains predicate mapping. 
The following section describes the meta-planning task of finding PFEOs from $D_2 
$ for the operators of $D_1$. This meta-planning task was introduced in \Cref{subsec:FindingPFEPOs}.

\section{Checking if an operator is a macro operator}\label{sec:checkMacro}
Checking if an operator is a macro is a straightforward task.
For an operator $o$ to be a macro in its domain, there must be a sequence of operators such that the consolidation of this sequence produces an operator that is identical to the operator $o$.
Testing if an operator $o$ is a macro in its domain depends on the concept of the core transitions of the operator $o$. This concept is defined as follows.
\begin{defn}
The core transitions of an operator $o$ are defined for a set of objects $obj$ which contains one object for each parameter in $o$ as the set of all transitions produced by applying each action $a_o$ from the set $A_{o,obj}$ to the states where the truth evaluations of the atoms of the preconditions of $a_o$ are the only true atoms.
\end{defn}
Note that this definition implies that in the core transition\textquotesingle s initial states, the truth evaluations of the atoms in the add effects of $a_o$ are false, and the truth evaluation of the atoms in the delete effects of $o$ are true.
Moreover, assume a sequence of operators $seq$ with the same parameters as $o$.
Let instantiate $o$ and $seq$ with the objects from the set of objects $obj$ which contains one object for each parameter in $o$.
If the union of the reach sets of the actions that are instantiated from $seq$ contains the union of the reach sets of the actions that are instantiated from $o$, then $seq$ and $o$ have the same effects, and the preconditions of $seq$ are in of the preconditions of $o$. 
Thus, we can infer that the reach set of $o$ is a subset of the reach set of $seq$ for any set of objects.
So, the reach set of $o$ is either equal to or a proper subset of the reach set of $seq$.
If the reach set of $o$ is equal to the reach set of $seq$, the operator $o$ is a macro and is made by consolidating the operators in the sequence $seq$.
On the other hand, if the reach set of $o$ is a proper subset of the reach set of $seq$, then $o$ is not a macro, but it is not a primitive operator because its reach set is a subset of the reach set of a sequence of operators.
In either case, we have to remove the operator $o$ as it is not primitive.

On the contrary, if no sequence of operators meets the description of the sequence $seq$ explained above, then we can confirm that the operator $o$ is not a macro. 
Note that in this case, we can only prove the reach set of $o$ is not a subset of the reach set of any sequence of operators from its domain; we cannot prove the reach set of $o$ is not a subset of the union of the reach sets of multiple sequences of operators. 
Therefore, we cannot judge whether the operator $o$ is primitive. 
We will discuss checking if non-macro operators are primitive in \Cref{sec:checkPrimitive}.

We formalise the problem of finding a sequence of operators $seq$ such that the reach set of $seq$ contains the core transitions of $o$ as a meta-planning problem.
Starting from the domain of the operator $o$, we create a modified domain that does not have the operator $o$ and any operator which cannot be part of any sequence of operators that has the same effects as $o$, and its preconditions are a subset of the preconditions of $o$. 
There are three rules an operator $o_o$ must satisfy to be part of the sequence of operators that makes a macro $m$ with effects similar to the effects of $o$ and with preconditions in the preconditions of $o$.
These rules are derived from the description of consolidated sequence of operators in \Cref{sec:sequenceOfOperatorsConsolidation}.
\paragraph*{First rule} \textit{The types of the parameters of $o_o$ must be in the types of the macro $m$.}
\paragraph*{Second rule} \textit{The number of parameters of any type in $o_o$ must be equal to or less than the number of parameters of the same type in $m$.} 
\paragraph*{Third rule} \textit{The predicates in the preconditions, add effects and delete effects of $o_o$ must be in the predicates of the preconditions, add effects and delete effects of $m$, respectively.} 

So the first step is to remove any operator that does not satisfy any of these rules.
Then, we define a dummy object for each variable in the parameters of the operator $o$.
After that, we use the FF planner \cite{hoffmann2001ff} to find a plan starting from a specific initial state $s_{init}$ to a goal state from a set of states $S_g$ that fulfils certain conditions.
Note that a plan is a sequence of actions which can be lifted to a sequence of operators.
The specific initial state $s_{init}$ is a state where only the truth evaluations of the atoms of the preconditions of $o$ are true, and the truth evaluation of any other atom is false.
For a state $s_g$ to be in the set $S_g$, the truth evaluations of the atoms of the add effects of $o$ in $s_g$ must be true and the truth evaluations of the atoms of the delete effects of $o$ in $s_g$ are false.
Moreover, the truth evaluation of any atom in the preconditions of $o$ that is not in its delete effects must be true in any goal state in $S_g$. 
This requirement excludes any sequence of operators with more delete effects than $o$.
Furthermore, the truth evaluation of any atom other than the atoms in the add effects of $o$ or the preconditions of $o$ that are not in its delete effects must be false in any goal state in $S_g$.
This condition prevents any sequence of operators with more add effects than $o$.

If such a plan is found, we can conclude that operator $o$ is a macro because the found plan represents a sequence of operators that can produce the same transitions as the operator $o$.
On the other hand, if no plan is found, then the operator $o$ is not a macro operator because there is no sequence of operators with a reach set equal to the reach set of $o$.
We call such an operator a non-macro operator.
We still have to do further checks to confirm whether the non-macro operator $o$ is primitive or not; this process is explained in \Cref{sec:checkPrimitive}. 
The task described in this section of checking if the reach set of a given operator is a subset of the reach set of a sequence of operators must be repeated for each operator in $D_1$ and $D_2$.
\section{Checking if a non-macro operator is a primitive operator in simple domains}\label{sec:checkPrimitive}
According to \Cref{def:primitieOperator}, to confirm if an operator from a simple domain $D$ is primitive, we have to prove that the reach set of this operator is not a subset of the union of the reach set of the members of any subset of the set of all sequences of operators in its domain without the given operator.
In \Cref{sec:checkMacro}, we have described how to check and remove any operator if its reach set is a subset of the reach set of any sequence of operators in its domain without this operator. 
Thus, the reach set of any reaming operator $o$ in \D1 and \D2 cannot be a subset of the reach set of a single sequence of operators because the operators with reach sets that are subsets of a single sequence of operators are moved in the previous section.
Nevertheless, the reach set of $o$ can be a subset of the union of reach sets of the members of any subset of the set of all sequences of operators in its domain without $o$.
This case is possible if the reach set of $o$ is a proper superset set of the reach sets of two or more sequences of operators.

Thus, for an operator $o$ to be not primitive, the reach sets of two or more sequences of operators must be proper subsets of the reach set of $o$, and the union of their reach sets must be a superset of the reach set of $o$.
Since the reach sets of two or more sequences of operators must be proper subsets of the reach set of $o$, the union of the reach sets of these sequences cannot be a proper superset of the reach set of $o$.
Hence, for an operator $o$ to be not primitive, the union of the reach sets of these sequences must be equal to the reach set of $o$.
For the reach set of a sequence $seq$ to be a proper subset of the reach set of $o$, the consolidation of $seq$ and the operator $o$ must fulfil the following conditions.
\begin{enumerate}
 \item The consolidation of $seq$ must have the same add and delete effects as $o$; and
 \item the set of the preconditions of the consolidation of $seq$ must be a proper superset of the set of the preconditions of $o$,~i.e. the preconditions of the consolidation of $seq$ must contain the preconditions of $o$ and at least one more additional precondition.
\end{enumerate}
\begin{defn}
We call a \textbf{S}equence of \textbf{O}perators with a reach \textbf{S}et that is a proper subset of the reach set of an \textbf{O}perator $o$ the \textbf{SOSO} for the operator $o$.
\end{defn}
Note that the consolidation of the SOSO of an operator $o$ has the same effects as $o$ and has a set of preconditions that is a proper superset of the preconditions of $o$. 

To check if an operator $o$ is primitive or not, we have to prove the existence or non-existence of two or more SOSOs for the operator $o$.
Unfortunately, we cannot use the method described in \Cref{sec:checkMacro} to find such a set of sequences.
Leveraging the meta-planning problem proposed in \Cref{sec:checkMacro} to find the SOSOs of a given operator $o$ would require the following modifications. 
The specific initial state $s_{init}$ will become a state where the truth evaluations of all atoms in the domain of $o$ are true apart from the atoms of the add effects of $o$, which must be false.
This requirement is needed to ensure that any found sequence of operators can have any set of preconditions that is a proper superset of the preconditions of $o$.
Note that since we have removed all sequence of operators which, when consolidated, will have a set of preconditions that is a subset of the preconditions of $o$ and have the same add and delete effects of $o$ in \Cref{sec:checkMacro}, the requirement of the modified specific initial state in the section guarantees the found sequence of operators starting from this specific initial state can have any set of preconditions that is a proper superset of the preconditions of $o$. 
The second condition is for a state $s_g$ to be in the set $S_g$, the truth evaluations of any atom in the domain of $o$ must be true in $s_g$ apart from the atoms the delete effects of $o$ which must be false in $s_g$.
This constraint ensures that any found sequence of operators will have the same delete effects as $o$.
However, it is impossible to have the conditions of the initial state and the goal states while ensuring the found sequence of operators has the same add effects as $o$.
Because it is possible for a sequence of operators to have one more add effect that is not in the add effects of $o$ and for this sequence of operators to satisfy the two conditions mentioned above as the atom in this add effect is already true in the initial state.
Therefore, we need another method to find such sequences. The method is explained in \Cref{sec:findSOSOsForO}.
For each non-macro operator in \D1 and \D2, this new method checks if there are multiple sequences of operators that satisfy the conditions mentioned above. 

If we prove the non-existence of two or more SOSOs for the operator $o$, then we can prove that the operator $o$ is a primitive operator.
On the other hand, if we find two or more SOSOs for the operator $o$, it is possible for the operator $o$ to be not primitive. 
To confirm whether the operator $o$ is primitive or not, we have to check if the union of the reach sets of the found sequence of operators is equal to the reach set of $o$.
If the reach set of $o$ is not equal to the union of the reach sets of the found sequences of operators, then $o$ is proved to be primitive.
On the contrary, if the reach set of $o$ is equal to the union of the reach sets of the found sequences of operators, then $o$ is proved to be not primitive. 

Checking whether the reach set of $o$ is equal to the union of the reach sets of the found sequences of operators is not in the scope of our automated method.
Instead, if we find two or more SOSOs for the operator $o$, then we ignore the possibility of the operator $o$ to be not a primitive operator, and we continue our search algorithm.
In this case, because we assumed all the operators of \D1 and \D2 are primitive, our method cannot conclusively disprove the functional equivalence between \D1 and \D1.
In lieu, our approach can provide a conditional verdict of no functional equivalence between \D1 and \D2 under the assumption that all operators are primitive. 
Conversely, our method can give a conclusive verdict of functional equivalence between \D1 and \D2 if we can find a predicate mapping that makes the reach set of \D1 equal to the reach set of \D2; even when some operators of \D1 and \D2 might not be primitive.

Nevertheless, the user can still get a conclusive verdict with some manual intervention when our method cannot prove that all \D1 and \D2 operators are primitive.
The user must manually verify whether any operator with a reach set equal to the union of the reach sets of multiple sequences of operators is primitive or not.
After that, the user can remove any operator found to be not primitive and re-invoke our method on the modified domains.

Then, our method will be guaranteed to provide a conclusive verdict. 
For each of these operators, our method provides the user with the set of consolidated sequences of operators that share the same effects as the given operator and with a set of preconditions that is a proper superset of the preconditions of the given operator.
\Cref{sec:ComparingReachsetsOperatorAndUion} explains how the user can use the information provided by our method to validate if such operators are primitive or not. 
\section{The meta-planning task of finding SOSOs of an operator $o$}\label{sec:findSOSOsForO}
The problem of checking the existence of multiple SOSOs for a given operator $o$ from a domain $D$ is treated as a meta-planning task.
Starting from the domain $D$, we create a modified domain $D'$ that does not have the operator $o$ and any operator that cannot be part of any sequence of operators with the same effects as $o$. 
We remove the operators that have effects not in the effects of the operator $o$ because for a sequence of operators to have the same effects as $o$, none of its operators can have an add effect or a delete effect which is not in the add and delete effects of $o$, respectively.
This meta-planning task aims to find a sequence of operators from the operators of $D'$ such that this sequence is a SOSO for the operator $o$.
The solution to this meta-planning problem is a plan that specifies which operators from $D'$ have to be added to an empty sequence of operators and in what order, so the produced sequence of operators is an SOSO for the operator $o$.
This planning task consists of a meta-planning domain model which is explained in \Cref{sec:Meta-planning-domain-model} and a meta-planning problem that is explained in \Cref{sec:Meta-planning-problem}.

The meta planning domain model is a macro-building planning domain model.
In a nutshell, it provides the planner with the tools to gradually add operators from the domain $D'$ to a macro template, which we call the macro under construction. 
The meta planning problem specifies the initial state of macro under construction, which should be empty without any operators.
Additionally, the meta planning problem sets out the conditions of the macro under construction in the meta goal states.
In any goal state, the macro under construction should have the required operators and be in the correct order to make an SOSO for the given operator $o$.
The search space of this meta-planning task is the operators of $D'$ that are instantiated with the minimal set of objects of $D'$ and $o$, which is described in \Cref{subsec:How-to-find-SOSOForO}.

Once the meta-planning domain and problem are produced for a given operator $o$, we use the FF planner to find a solution to the meta-planning problem.
If the planner confirms the meta-planning problem is unsolvable, then the operator $o$ is primitive.
On the contrary, if the planner found a plan, then the operator $o$ is potentially not a primitive operator. 
Thus, we need to do further checks.
First, we have to find more SOSOs for the operator $o$; this process is explained in \Cref{subsec:MultipleSOSOs}.
If we do not find any SOSOs other than the first one, then the operator $o$ is primitive.
However, suppose we find two or more SOSOs. In that case, if the domains \D1 and \D2 are not proven to be functionally equivalent, the user can manually check if the potentially not primitive operators are primitive or not as described in \Cref{sec:ComparingReachsetsOperatorAndUion}. 

\subsection{How to find an SOSO for an operator $o$?}\label{subsec:How-to-find-SOSOForO}
To find one SOSO from $D'$ for an operator $o$ form domain $D$, we have to find a macro with the same effects as the operator $o$.
That means we have to find a sequence of lifted operators from $D'$ and a unification and unionisation configuration of the parameters of consecutive operators, such that the macro produced from consolidating this sequence of operators has the same effects as the operator $o$.

Instead of searching for a sequence of operators in the space of lifted operators of $D'$ and then searching for a proper unification and unionisation configuration of the parameters of consecutive operators, we propose to do the search in the space of ground operators of $D'$.
The advantage of searching for SOSO for the operator $o$ in the space of ground operators of $D'$ is that we do not have to worry about finding the proper unification and unionisation configuration of the parameters of consecutive operators.
This advantage is realised because the search in the space of ground operators deals with objects rather than parameters.
Unification and unionisation configuration of the parameters of consecutive operators is required when we reason about lifted operators because the scope of parameters is limited to their operators. 
On the other hand, we can readily consolidate a sequence of ground operators because the scope of objects, which depends on the scope of types, is general to all operators of $D'$.

Note that a ground macro and its lifted macro have the same effects.
Therefore, finding a ground macro with the same effects as $o$ is the same as finding a SOSO for the operator $o$.
After finding a ground macro with the same effects as the operator $o$, we have to lift the objects of this macro to get a SOSO for the operator $o$.
Lifting the objects of a macro is a straightforward process because we know the types of the objects.

To work with ground operators, we need to instantiate the operators of $D'$ with enough objects to ensure no SOSO of the operator $o$ is excluded from the search space.
We need to have enough ground operators from $D'$ to produce all possible SOSOs of the operator $o$.
Note that the SOSOs of $o$ with the maximum number of parameters is the sequence of operators where the parameters of the preconditions of each two consecutive operators are united.
The set of the preconditions of the operators of any sequence of operators from $D'$ is equal to the union of the preconditions of the operators of $D'$.
Moreover, any parameter of the parameters of the add effects of the operator $o$ that is not in the parameters of the preconditions of the $o$ must also be united with the parameters of the SOSOs of $o$ with the maximum number of parameters.
Therefore, the number of objects of each type in $D'$ must be equal to the number of the parameters of the SOSOs of $o$ with the maximum number of parameters.
We call this set of objects the minimal set of objects for the domain $D'$ and the operator $o$.
With the minimal set of objects, the search space of ground operators from $D'$ will enable us to find any sequence of operators $seq$ with any unionisation or unification configuration, such that $seq$ is an SOSO for the operator $o$.
\subsection{Meta planning domain model}\label{sec:Meta-planning-domain-model}
Like other regular planning domain models, our meta-planning domain model has types, atoms, and operators.
We suffix these components with the word ``meta'' to indicate that these parts belong to a meta planning domain model. 
The meta parts of our meta planning domain model \D M are derived from the respective parts of the domain $D'$.
\paragraph*{Meta atoms}
The atoms in the meta domain \D M are derived from the atoms of the domain $D'$.
For every atom that appears as a precondition in the operators of $D'$, a meta atom is added to the meta domain \D M.
The name of this added atom is the name of the original atom suffixed with ``pre-om''.
\sloppy For example, for the atom $\operatorname{(at-soil-sample\ ?p)}$ which is a precondition in the operator ``$\operatorname{sample-soil}$'' in $D'$,
we add the meta atom $\operatorname{(at-soil-sample-pre-om\ ?p)}$ to the atoms of the meta domain \D M.
Similarly, new atoms are added to the meta domain with suffixes ``add-om'' and ``del-om'' for the atoms in add and delete effects in the operators of the domain $D'$.
The meta-atoms will track the preconditions, add effects and delete effects of the macro under construction.
For instance, when a meta operator adds ``$\operatorname{(at-soil-sample\ ?p)}$'' as a precondition to the macro under construction $m$,
this meta operator has to set to true the meta atom ``(at-soil-sample-pre-om ?p)'' to indicate ``(at-soil-sample ?p)'' is now a precondition of $m$.

Note that the difference between the atoms of $D'$ and the meta-atoms of the meta domain of \D M is just the name of the atoms.
Therefore, the types of the meta-atom variables are the same as those of the domain $D'$ atoms.
\paragraph*{Meta operators}
The meta operators of the meta planning domain \D M add the operators from domain $D'$ to the macro $m$.
For each operator $o$ of $D'$, there is a meta operator that adds the operator $o$ to the end of the macro $m$.
\textbf{The name of the meta operator} that adds an operator $o$ is the name of the operator $o$ prefixed with the word``add''.
For example, the meta operator that adds the operator ``$\operatorname{sample-soil}$'' from $D'$ to the end of a macro $m$ is ``$\operatorname{add-sample-soil}$''.
Besides the name, meta operators also have parameters, preconditions, add effects, conditional add effects, and conditional delete effects, but no delete effects.

The preconditions and effects of the meta operators have meta-atoms.
The meta-atoms are derived from the atoms of the normal operators for which the meta operators are defined.
Since the meta-atoms of the meta operators are derived from the atoms of the normal operators by only suffixing the name of the atoms, both normal and meta-atoms have the same parameters.
Furthermore, since the parameters of the operators are made from the union of the parameters of their atoms, \textbf{the parameters of the meta operators} are the same as the parameters of the normal operators.

\textbf{The preconditions of the meta operators} ensure only valid macros are built by guaranteeing any operator $o_o$ from $D'$ is added by its meta operator to the macro under construction only if none of its preconditions appears as a delete effect in the macro under construction.
This is achieved by using negative preconditions in the meta operator. 
The atoms in the preconditions of the operator $o_o$ are suffixed with ``del-om'' and added as negative preconditions to the meta operator. 
For example, let assume the operator ``$\operatorname{soil-sample}$'' have a precondition ``$\operatorname{(at-soil-sample\ ?p)}$'', then its meta operator ``$\operatorname{add-soil-sample}$'' will have as a meta preconditions ``$\operatorname{(not\ (at-soil-sample-del-om\ ?p))}$''.
This negative precondition means the meta operator ``add-soil-sample'' cannot be scheduled,~i.e. the operator ``soil-sample'' cannot be added to the macro under construction if the meta atom ``(at-soil-sample-del-om ?p)'' is true.
This meta atom is true if only there is a proceeding operator in the macro under construction that has ``(at-soil-sample ?p)'' as a delete effect, and none of the subsequent operators in the macro under construction has it as an add effect.

\textbf{The meta add effects} add the delete effects of the operator represented by the meta operator to the macro under construction.
The atoms in the delete effects of the normal operator are suffixed with ``del-om'' and added as meta add effects to the meta operator. 
For example, let assume the operator ``$\operatorname{soil-sample}$'' have a delete effect ``$\operatorname{(at-soil-sample\ ?p)}$'', then its meta operator ``$\operatorname{add-soil-sample}$'' will have as a meta add effect ``$\operatorname{(at-soil-sample-del-om\ ?p)}$''. 
This meta add effect means when the meta operator ``add-soil-sample'' is scheduled,~i.e. the operator ``soil-sample'' is added to the macro under construction, then the meta atom ``(at-soil-sample-del-om ?p)'' will become true to indicate the macro under construction has ``(at-soil-sample ?p)'' as a delete effect.

The meta-add effects have another important use.
Meta add effects play an essential role in excluding previously found macros when searching for new macros is the process of finding all PFEOs of a given operator.
This process is further explained in \Cref{subsec:MultipleSOSOs}.

\textbf{The meta conditional effects} are used by the meta operators to update the preconditions, add effects, and delete effects of the macro under construction according to the preconditions, add effects, and delete effects of their represented normal operators and according to the status of the macro under construction. 
The meta conditional effects are generated using four rules. 
These rules are explained in the following subsection.
\subsubsection{Meta conditional effects rules}
In the description of the meta conditional effects we will use the functions $pre(o_o)$, $add(o_o)$ and $del(o_o)$.
These three functions return the atoms in the preconditions, add effects and delete effects of the operator $o_o$, respectively.

Let us consider an operator $o_o$ and a macro under construction $m$. When appending the macro $m$ with the operator $o_o$, the first rule dedicates each atom $t$ in the preconditions of $o_o$ must be added as a precondition to $m$ unless $t$ has already been added to the preconditions of $m$ in a previous step, or the precondition $t$ has a supporter in $m$,~i.e. $t$ is already supported by an add effect in $m$. Hence $t$ does not have to be added as a precondition to $m$. 
\begin{enumerate} \label{list:build-macro}
 \item For all $t$ in $pre(o_o)$, $t$ is added to $pre(m)$ if $t$ is neither in $pre(m)$ nor in $add(m)$
 \end{enumerate}
The second rule states that when adding an operator $o_o$ with an atom $t$ as an add effect to the macro $m$ which has the same atom $t$ as a delete effect, the delete effect of $t$ in the macro $m$ has to be removed.
It is essential to remove any violators for $t$ in $m$ when appending $o_o$ with the add effect $t$ to $m$.
\begin{enumerate}[resume]
 \item For all $t$ in $add(o_o)$, $t$ is removed from $del(m)$ if $t$ is in $del(m)$\label{remove_delete_p_delm-prem}
\end{enumerate}
The third rule dedicates any add effect $t$ of $o_o$ have to be added as an add effect to $m$ unless $t$ has already been added to the add effects of $m$ in a previous step, or the atom $t$ does not have to be added as an add effect to $m$ as it is already a precondition for $m$ and the macro $m$ does not have a delete effect of this atom.
If a delete effect of this atom existed in $m$, it would have been removed from $m$ by the second rule.
\begin{enumerate}[resume]
 \item For all $t$ in $add(o_o)$, $t$ is added to $add(m)$ if $t$ is neither in $add(m)$ nor in $pre(m)$ \label{add_addm}
\end{enumerate}
The fourth rule implies that the add effects of $m$ that match the delete effects of $o_o$ have to be removed as add effects of $m$, so we do not have delete and add effects of the same atoms in the macro $m$.
\begin{enumerate}[resume]
 \item For all $t$ in $del(o_o)$, $t$ is removed from $add(m)$ if $t$ is in $add(m)$
 \end{enumerate}

This section explained our meta-planning domain model, including the meta-atoms and meta-operators.
The following section describes the meta-planning problem, which includes our meta-planning task's initial state and goal condition.
Then, \Cref{subsec:MultipleSOSOs} explains how the meta-planning domain model and the meta-planning problem are modified to find multiple SOSOs for each operator from $D$.
\subsection{Meta planning problem}\label{sec:Meta-planning-problem}
Our method produces a new planning problem for every operator $o$ in $D$.
The solution to this planning problem is a sequence of meta actions that gradually build a macro from the operators of $D'$ with the same effects as the operator $o$. 
The components of our meta-planning problem are discussed in the following paragraphs.

\paragraph*{Meta objects}
In the previous section, we have explained that the meta-planning domain model has the same types as the domain $D'$. 
 Moreover, in \Cref{subsec:How-to-find-SOSOForO}, we have explained the number of required objects for each type of $D'$.
\paragraph*{Meta initial state}
The meta initial state does not have any atom. This configuration means the macro under construction has no precondition, add effect or delete effect. This means the macro under construction does not have any operator yet.
\paragraph*{Meta goal}
The goal condition requests only the meta-atoms that capture the add and delete effects of $o$ to be true in any meta goal state.
If the truth evaluations of the meta-atoms that capture the add and delete effects of $o$ are true in a meta state, then the transitions from the empty meta initial state to this meta state define a sequence of ground operators such that its consolidation has the add and delete effects of $o$.
Note that the goal condition implies that all meta-atoms other than the meta-atoms of the add and delete effects of $o$ must be false in any meta goal state.
This requirement enforces that any found sequence of operators does not have add or delete effects other than those of $o$.

Though the goal condition does not specify any constraint on the meta-atoms that represent the preconditions of $o$ and the operators of $D'$, the preconditions of the consolidations of the sequences of operators which are found by this meta-planning problem are guaranteed to be proper supersets of the preconditions of $o$.
Note that any sequence of operators with the same effects as $o$ and with a set of preconditions which is a subset of the preconditions of $o$ was identified and removed in \Cref{sec:checkMacro}.
Therefore any renaming sequence of operators with the same effects as $o$ must have a set of preconditions that is a proper superset of the preconditions of $o$.

Therefore, the steps to build macro with the same effects as $o$ and with a set of preconditions that is a superset of the preconditions of $o$ is a solution to this problem task. 
If we do not find any such sequence or if we find just one such sequence, then the operator $o$ is a primitive operator. 
On the other hand, if we find two or more such sequences, then the operator $o$ is potentially not a primitive operator. 
In this section, we have explained how to find a sequence of operators with the same effects as $o$ and with a set of preconditions that is a superset of the preconditions of $o$.
The following section explains how to search for multiple SOSOs for a given operator $o$.

\subsection{Finding multiple SOSOs for an operator $o$} \label{subsec:MultipleSOSOs}
To check if an operator $o$ is primitive, we have to show that its reach set is not equal to the union of reach sets of multiple SOSOs of the operator $o$.
Every SOSO of $o$ is described by a solution to the macro-building planning task of the operator $o$.
Hence, after every time a solution to this meta-planning task is found using a planner, we have to modify the meta-planning domain model and the meta-problem of this task to exclude the previous solution and then rerun the planner with the modified planning task to find a new SOSO to the operator $o$.
This means we evolve the meta-planning task gradually to exclude found SOSOs one by one until no more SOSOs can be found for the given operator $o$.

Excluding previous solutions (SOSOs) is achieved by updating the meta domain and problem with an automaton to reject the last returned solutions by the planner.
A solution to the macro-building planning task is a sequence of meta operators from the meta planning domain model \D M, which orders a configured sequence of operators from $D'$.
Thus, to exclude a sequence of operators from $D'$, we have to exclude the sequence of meta operators from \D M that specifies the sequence of operators from $D'$.
To exclude a sequence of meta operators from \D M, we use auxiliary atoms and conditional effects to capture the automaton that will reject a sequence of meta operators.

To exclude a ground macro, we amend the meta operator $mo_0$ that adds the first ground operator $go_0$ to the macro with an add effect to set an auxiliary atom $s_0$ to true.
Then, we add a conditional effect to the meta operator $mo_1$ that adds the second ground operator $go_1$ to the macro.
The condition of this conditional effect is the atoms $s_0$.
The effect of the conditional effect is an atom $s_1$.
We continue adding conditional effects to the meta operators that add the subsequent ground operators to the macro. 
The condition of the conditional effect of a meta operator $mo_i$ is the auxiliary atom $s_{i-1}$ from the effect of the conditional effect of the meta operator $mo_{i-1}$.
The effect of the conditional effect of the meta-operator $mo_i$ is an atom $s_i$.
After all meta operators are augmented with the required additional conditional effects, we add the negation of the auxiliary atom $s_n$ of the last meta operator in the sequence that we want to exclude to the goal condition of the meta planning problem.

Since any goal state cannot have $s_{n}$ set to true, the sequence of operators that sets $s_{n}$ to true cannot be returned by the planner as a solution to the meta-planning problem. 
Thus, this sequence of operators is excluded from the reachable state space of the macro-building planning task of the operator $o$.

Every time we want to exclude a ground macro, we have to amend its meta operators with the required auxiliary atoms and their conditional effects.
In addition, we have to amend the meta-planning problem with the negation of the auxiliary atom $s_n$.
We repeat this process until the planner proves the augmented macro-building task for the operator $o$ is unsolvable.
The set of SOSOs of the operator $o$ is provided to the user to check whether the operator $o$ is primitive or not.
The following section describes the process of checking if an operator $o$ with a set of multiple SOSOs is primitive or not.

\section{Comparing the reach sets of an operator and the union of the reach sets of multiple sequences of operators}\label{sec:ComparingReachsetsOperatorAndUion}
In the case a conclusive decision cannot be made by our method because some operators in \D1 or \D2 are not guaranteed to be primitive, it is possible for the user to further investigate the functional equivalence of \D1 and \D2 by removing non-primitive operators from the two domains and then re-invoke our method on the modified domains.

For the reach set of an operator $o$ to be equal to the union of the reach sets of some of its SOSOs, the disjunction of the preconditions of the consolidated SOSOs must be equal to the preconditions of the operator $o$.
To simplify the disjunction of the preconditions of the consolidated SOSOs, we need to transform the atoms from the predicate logic to the propositional logic. To do so, we simply replace each atom with its predicate. 
Then we replace each predicate with a proposition.
Note that the user needs to know the pairs of mutually exclusive atoms in each domain, so their predicates can be assigned with commitment propositions.
Under the closed-world assumption, a pair of atoms are mutually exclusive if, in any state, when both atoms are instantiated with the same objects, only one is true, and the other is false.

Since the scope of our method is limited to PDDL2.1 without negative and disjunctive preconditions, the preconditions of the operators can be expressed as a Disjunctive Normal Form (DNF) formula.

If the disjunction of the preconditions of the SOSOs of the operator $o$ is equivalent to the precondition of the operator $o$, then the reach set of $o$ is proven to be equal to the union of the reach sets of its SOSOs.
Therefore, the operator $o$ is not a primitive operator.
On the other hand, if the simplification of the disjunction of the preconditions of the SOSOs of the operator $o$ is not equivalent to the precondition of the operator $o$, then the reach set of $o$ is proven to be not a subset of the union of the reach sets of its SOSOs.
Thus, the operator $o$ is a primitive operator.
After identifying and removing any non-primitive operators, the user can re-invoke our method to get a conclusive verdict on the functional equivalence of the given planning domain models. 

This section concludes the description of our method for validating the functional equivalence of simple planning domain models. 
The following section explains our approach to validating the functional equivalence of complex planning domain models. 
\section{Worked example} %
\label{sec:example}
There is no better example to explain our method than validating the functional equivalence between the mystery domain \cite{McDermott_2000} and the transportation domain from which the mystery domain was devised.
To conceal the identity of the transportation domain, the names of its predicates and operators were disguised by random names like eats, craves, fears, orbits, overcome, feast and succumb. 
This disguise makes the transportation and mystery domains very nice candidate test domains.
Before we tested the functional equivalence of these two domains, we had to apply some changes to make them fit the requirements of our tool. 
In the mystery domain, we had to transform the static atoms that implicitly specify the types of the parameters of the operators into explicit types. 
For instance the operator ``overcome'' has the untyped parameters (?c ?v ?n ?s1 ?s2 ).
The types of these variables are defined by the static atoms (pain ?c) (pleasure ?v) (food ?n), and (planet ?s1). 
Thus, we have removed these static atoms and constrained the parameters of this operator with types that are derived from their static predicates. 
Hence, we have changed the untyped parameters of the operator ``overcome'' to (?c - pain ?v - pleasure ?n - food ?s1 - planet ?s2 - planet).
Furthermore, in the mystery domain, we had to split the predicate ``(at ?v ?n)'' to two predicates ``(craves\_v ?v - pleasure ?n - food)'' and ``(craves\_c ?c - pain ?n - food)''.
Similarly, in the transportation domain, we had to split the predicate ``(at ?v ?n)'' to two predicates ``(at\_v ?v - vehicle ?n - node)'' and ``(at\_c ?c - cargo ?n - node)''.

The verdict of our tool is that the two domains are functionally equivalent.
Moreover, our tool provided the variable mappings listed in \Cref{tab:MysteryVariableMapping}.
\MysteryVariableMapping
These variable mappings are the solution to the SMT problem explained in \Cref{sec:SMT}. 
As described in \Cref{sec:descriptionOfDecisionVariableMapping}, the variable mappings provided in \Cref{tab:MysteryVariableMapping} represent atom mappings.
These atom mappings respect the conditions of \Cref{th:operatorsStructurReachSet} because they are generated from the solution of our SMT problem, which is designed to enforce the constraints of \Cref{th:operatorsStructurReachSet}.
Note that these atom mappings are not shown explicitly; they are impeded in the variable mappings in \Cref{tab:MysteryVariableMapping}.
Thus, according to \Cref{th:operatorsStructurReachSet}, for each atom mapping that is produced from the variable mappings in \Cref{tab:MysteryVariableMapping}, there is a predicate mapping that makes the reach set of each operator from the Transportation domain equals to the reach set of an operator from the Mystery domain. 
The predicate mappings that result from the atom mappings, which in their turn are produced from the variable mappings shown in \Cref{fig:MysteryPredicateMapping}.
\MysteryPredicateMapping

Moreover, these mappings are consistent with regard to the mapping of their predicates because they are produced from the solution of our SMT problem, which has the constraint that enforces this predicate consistency.
Therefore according to \Cref{th:SimpleDomainsReachabilityTheorem}, under the predicate mapping, which is made from the union of the predicate mappings in \Cref{fig:MysteryPredicateMapping}, the reach set of the Transportation domain is equal to the reach set of the Mystery domain. 
Hence, as per the \Cref{Def:functional-equivalence}, the Transportation domain is functionally equivalent to the Mystery domain.
Furthermore, our tool extracted the operator mapping depicted in \Cref{fig:MysteryOperatorMapping} from the found variable mappings.
\MysteryOperatorMapping

\section{Validating the functional equivalence of complex planning domain models}\label{sec:ValidatingFEComplexDomains}
This section presents a method for proving the functional equivalence of complex planning domain models.
According to \Cref{th:SimpleDomainsReachabilityTheorem}, if the reach set of a primitive operator $o$ in \D1 is not equal to the reach set of any primitive operator in \D2, then \D1 and \D2 do not have an equal reach set.
This conclusion is only valid for simple planning domain models. 
For complex domains, \Cref{th:SimpleDomainsReachabilityTheorem} does not stand.
For instance, assume a complex planning domain model \D1 that has an \SE set with three operators $o_1$, $o_2$, and $sp$. Let $sp$ be a split operator, and $o_1$ and $o_2$ are the core operators of $sp$.
Let \D2 be a planning domain model with an operators $o'$ such that $\Gamma(F_p(sp)) = \Gamma(o')$. Additionally, suppose $\forall o' \in O_2(\Gamma(F_p(o_1)) \neq \Gamma(o') \wedge \Gamma(F_p(o_2)) \neq \Gamma(o'))$
Since $\Gamma(F_p(sp)) = \Gamma(o')$, any transition produced by $F_p(sp)$ can be produced by $o'$ and vice versa.
Moreover, even the reach set of $o_1$ and $o_2$ are not equal to the reach set of any operator in \D2, any transition produced by either $o_1$ and $o_2$ can be also produced by $o'$.
Thus, the reach set of \D1 is equal to the reach set of \D2 under $F_p$ even with the reach sets of the primitive operators $o_1$ and $o_2$ are not equal to the reach set of any operator in \D2.
Note that according to \Cref{th:SimpleDomainsReachabilityTheorem}, the reach set of \D1 would have been considered as not equal to the reach set of \D2 because the reach set of the primitive operators $o_1$ and $o_2$ are not equal to any primitive operator from \D2.
However, \D1 and \D2 have an equal reach set.

Therefore, we need to relax \Cref{th:SimpleDomainsReachabilityTheorem} to design a new theorem that is valid for complex planning domain models.
The new theorem will not rule that the reach set of \D1 is not equal to the reach set of \D2 under $F_p$.
First, we remove the implication that states if the reach set of a primitive operator from \D1 is not equal to the reach set of any primitive operator in \D2, then the reach set of \D1 is not equal to the reach set of \D2.
Thus, the implication that remains from \Cref{th:SimpleDomainsReachabilityTheorem} states if \D1 and \D2 have an equal number of primitive operators and the reach set of each primitive operator from \D1 is equal to the reach set of a primitive operator from \D2, then the reach set of \D1 is equal to the reach set of \D2.
Note that this implication is restricted to the set of primitive operators.
However, since \D1 and \D2 are not simple domains, they might have other types of operators.
Therefore, we further modify this implication to include all types of operators. 
As such, we omit the restriction of the operators to be primitive only, and we remove the condition that the two domains must have an equal number of primitive operators.
The premise of the produced implication is that the reach set of each operator from \D1 must be equal to the reach set of an operator from \D2.
This new premise does not require the reach sets of all operators of \D2 to be equal to the reach set of some operators in \D1. 
The consequence of this premise cannot imply that \D1 and \D2 have an equal reach set.
Therefore, we propose the conclusion of this new implication to be that the reach set of \D1 is a subset of the reach set of \D2.
So, the new implication is that if the reach set of each operator from \D1 is equal to the reach set of an operator from \D2, then the reach set of \D1 is a subset of the reach set of \D2.
This new implication is captured in \Cref{th:ComplexDomainsReachabilityTheorem} and proved in \Cref{ProofOfTheoremCDRT}.

Note that the consequence of the implication in \Cref{th:ComplexDomainsReachabilityTheorem} can only imply a containment relation rather than an equality relation between the reach sets of planning domain models.
Therefore to prove $\Gamma(F_p(D_1)) = \Gamma(D_2)$, we need to prove $\Gamma(F_p(D_1)) \subseteq \Gamma(D_2)$ and $\Gamma(D_2) \subseteq \Gamma(F_p(D_1))$.
Rather than proving $\Gamma(D_2) \subseteq \Gamma(F_p(D_1))$, it is easier to prove the existence of a bijective mapping $G_p: P_2 \rightarrow P_1$ such that $\Gamma(G_p(D_2)) \subseteq \Gamma(D_1)$.
We found that starting from $\Gamma(F_p(D_1)) \subseteq \Gamma(D_2)$ and $\Gamma(G_p(D_2)) \subseteq \Gamma(D_1)$ we can deduce $\Gamma(F_p(D_1)) = \Gamma(D_2)$.
This claim is captured in \Cref{th:ContainmentTheorem} and proved in \Cref{ProofOfTheoremDCT}.

The first step to validating the functional equivalence of complex planning domain models is to remove macro operators because we know macro operators do not alter the reach set of their domains.
Removing macro operators is explained in \Cref{sec:ValidatingFESimpleDomains}.
After that, we follow the steps in \Cref{sec:ValidatingFESimpleDomains} to find a mapping $f_p$ from the predicates of each operator $o$ from \D1 to the predicates of an operator $o'$ from \D2 such that $\Gamma(f_p(o)) = \Gamma(o')$ and the union of all individual mappings from the operators of \D1 to operators in \D2 is a well-defined function $F_p$.
Then, according to \Cref{th:ComplexDomainsReachabilityTheorem}, the reach set of \D1 is a subset of the reach set of \D2 under $F_p$.
We repeat the previous steps to find a mapping $g_p$ from the predicates of each operator $o'$ from \D2 to the predicates of an operator $o$ from \D1 such that $\Gamma(g_p(o')) = \Gamma(o)$ and the union of all individual mappings from the operators of \D2 to operators in \D1 is a well-defined function $G_p$.
Then, according to \Cref{th:ComplexDomainsReachabilityTheorem}, the reach set of \D2 is a subset of the reach set of \D1 under $G_p$.

So, we have $\Gamma(F_p(D_1)) \subseteq \Gamma(D_2)$ and $\Gamma(G_p(D_2)) \subseteq \Gamma(D_1)$. 
Therefore, according to \Cref{th:ContainmentTheorem}, $\Gamma(F_p(D_1)) =\Gamma(D_2)$. 
Thus, as per \Cref{Def:functional-equivalence}, \D1 and \D2 are functionally equivalent.

Instead of finding two mappings $F_p$ and $G_p$ to prove the equality of the reach sets of \D1 and \D2, \Cref{cor:DomainReachSetsEqualityCorollary} states that if \D1 and \D2 have an equal number of operators and the reach set of either of the domains is a subset of the reach set of the other domain under a predicate mapping. The reach sets of both domains are equal under that mapping and its inverse mapping as well.
\Cref{cor:DomainReachSetsEqualityCorollary} dramatically reduces the time required to prove the functional equivalence of two domains by enabling us to search for one mapping rather than two and still reach the same conclusion.

This description completes the explanation of our approach to validate the functional equivalence of complex planning domain models.
The following section lists the theorems used in our approach and points to their proofs.

\subsection{Complex Domains Reachability Theorem}\label{sec:ComplexDomainsReachabilityTheorem}
Let $\mathbb{F}$ be the set of all bijective mappings from predicates of equal arity from every operator in $D_1$ to every operator in $D_2$.
\begin{alignat}{1}
\mathbb{F} = \{ f_p | f_p: Predicates(o) \twoheadrightarrowtail Predicates(o') \ &\text{ where } o \in O_1, o' \in O_2 \nonumber \\
\text { and if } f_p(p) = p' \text{ then } Arity(p) = Arity(p') \}\nonumber 
\end{alignat}

Let $R_{OM}$ be a relation between operators from $D_1$ and predicate mappings from $\mathbb{F}$. An operator $o$ from $D_1$ is related to a mapping $f_p$ from $\mathbb{F}$ by $R_{OM}$ means there exists an operator $o'$ from $D_2$ such that the reach set of $f_p(o)$ is equal to the reach set of $o'$.
$$ R_{OM} = \{ (o,f_p) \in O_1 \times \mathbb{F} \ | \ \exists o' \in O_2, \ \Gamma(f_p(o),Obj) = \Gamma(o',Obj) \}$$
\begin{restatable}[Complex Domains Reachability Theorem]{theorem}{CDRT}
\label{th:ComplexDomainsReachabilityTheorem}
Consider a set of objects \textit{Obj}, two planning domain models, $D_1$ and $D_2$, a bijective function $F_p$ from the predicates of $D_1$ to the predicates of $D_2$ with equal arities, and the relation $R_{OM}$ that relates each operator $o$ in $O_1$ to a bijective predicate mapping $f_p$ that makes the reach set of $f_p(o)$ equals to the reach set of an operator from $O_2$. We have:
\begin{multline}
\exists R'_{om} \subseteq R_{OM}( Domain(R'_{om}) = O_1 \ \wedge \\ (F = \bigcup\limits_{f_p \in Range(R'_{om}) } f_p )\text{ is a well-defined function}) \implies \Gamma(F_p(D_1),Obj) \subseteq \Gamma(D_2,Obj) \nonumber
\end{multline}
\end{restatable}
\begin{restatable}[Complex Domain Reachability Lemma]{lem}{CDRL}
\label{lem:ComplexDomainReachabilityLemma}
Consider a set of objects \textit{Obj}, two planning domain models $D_1$ and $D_2$, and a bijective function $F_p: P_1 \twoheadrightarrowtail P_2$.
We have
\begin{multline}
\forall o \in O_1, \exists o' \in O_2 ( \Gamma(F_p(o),Obj) = \Gamma(o',Obj)) \implies \Gamma(F_p(D_1),Obj) \subseteq \Gamma(D_2,Obj)
\end{multline}
\end{restatable}
\Cref{th:ComplexDomainsReachabilityTheorem} is proven in \Cref{ProofOfTheoremCDRT} with the help of \Cref{lem:ComplexDomainReachabilityLemma}.
The proof of this lemma is provided in \Cref{sec:ComplexDomiansProofOfLemma}.
\subsection{Reach Sets Containment Theorem}

In order to prove $D_1$ and $D_2$ are functionally equivalent according to \Cref{Def:functional-equivalence}, we have to prove $\Gamma(F_p(D_1),Obj) = \Gamma(D_2,Obj)$. The Reach Sets Containment Theorem shows that for two planning domain models $D_1$ and $D_2$, we have $\Gamma(F_p(D_1),Obj) = \Gamma(D_2,Obj)$ if and only if $\Gamma(F_p(D_1),Obj) \subseteq \Gamma(D_2,Obj)$ and $ \Gamma(G_p(D_2),Obj) \subseteq \Gamma(D_1,Obj)$. This theorem is formalised as follows.

\begin{restatable}[Reach Sets Containment Theorem]{theorem}{DCT}
\label{th:ContainmentTheorem}
For two planning domain models $D_1$ and $D_2$ and two bijective functions $F_p: P_1 \twoheadrightarrowtail P_2$ and $G_p: P_2 \twoheadrightarrowtail P_1$,
let $F_p(D_1)$ be the image of $D_1$ using $F_p$ to substitute its predicates with those of $D_2$, and $G_p(D_2)$ be the image of $D_2$ under $G_p$. Then we have for a set of objects {\it Obj}:

\begin{enumerate}
 \item $\Gamma(F_p(D_1),Obj) \subseteq \Gamma(D_2,Obj)$ and $ \Gamma(G_p(D_2),Obj) \subseteq \Gamma(D_1,Obj) \iff \Gamma(F_p(D_1),Obj) = \Gamma(D_2,Obj)$ and $F_p = G_p^{-1}$.
 \item $\Gamma(F_p(D_1),Obj) \subseteq \Gamma(D_2,Obj)$ and $ \Gamma(G_p(D_2),Obj) \subseteq \Gamma(D_1,Obj) \iff \Gamma(G_p(D_2),Obj) = \Gamma(D_1,Obj)$ and $G_p = F_p^{-1} $.
\end{enumerate}
\end{restatable}

This theorem is proven in \Cref{ProofOfTheoremDCT}

\subsection{Domain Reach Sets Equality Theorem}\label{sec:DomainReachSetsEqualityTheorem}
The Domain Reach Sets Equality Theorem states that for any set of objects, the reach set of a planning domain model $D_1$ is equal to the reach set of a planning domain model $D_2$ under a bijective predicate mapping $F_p$ if and only if the reach set of the image of $D_1$ under $F_p$ is a subset of the reach set of $D_2$, and the two domains $D_1$ and $D_2$ have an equal number of operators.
This theorem is formalised as follows:

\begin{restatable}[Domain Reach Sets Equality Corollary]{cor}{DRSEC}
\label{cor:DomainReachSetsEqualityCorollary}

Consider a set of objects \textit{Obj}, two planning domain models, $D_1$ and $D_2$, a bijective function $F_p$ from the predicates of $D_1$ to the predicates of $D_2$ with equal arities. We have:
\begin{equation}
\forall o \in O_1, \exists o' \in O_2 (\Gamma(F_p(o)) = \Gamma(o')) \wedge |O_1| = |O_2| \implies \Gamma(F_p(D_1) = \Gamma(D_2) \nonumber
\end{equation}
\end{restatable}
This corollary is proven in \Cref{ProofOfCorollaryDRSEC}.

\section{Experiment} %
\label{sec:experiment}
To demonstrate the feasibility of our method, we implemented it and tested it with 75 functional equivalence validation tasks.
We used 13 planning domain models from the International Planning Competition (IPC).
From 10 domains, we created six modified versions, and from three domains, we created five modified versions so that we could validate the functional equivalence between the original domains and their modified versions.
The six proposed modifications are intended to test the response of our method to the change in the number of operators, change of atoms in the preconditions and add effects, and change in the parameters of the atoms.

Three modifications increase the number of operators by augmenting the modified versions with a hand-crafted valid macro operator, a randomly created valid macro, and a randomly created invalid macro.
One modification decreases the number of operators by deleting a randomly selected operator from the original domain to create a modified version. 
The modification that changes the atoms of operators swaps one atom from the add effect of an operator with a precondition from the same operator to create a modified version.
To make a change in the parameters of an atom, the last modification swaps two variables of the same type in the parameters of an atom in any part of an operator.

We have developed a random test generation tool to apply these modifications to the original domains. 
This tool randomly selects the targeted operator, atom and variable for each modification.
To create randomly valid macros, we have leveraged the method of finding SOSOs for operators, explained in \Cref{sec:findSOSOsForO}. 
We have modified the macro-building task to find a valid sequence of operators of a certain length.
We have added a numerical fluent ``operators-count'' to track the number of operators added to the macro under construction.
Each macro operator is augmented with an add effect to increase the fluent ``operators-count''.
Thus, whenever the planner schedules a meta-operator to add its normal operator to the macro under construction, the value of ``operators-count'' gets increased by one.
The goal condition of this modified macro building problem is the number of the required operators in the targeted macro.
The initial state of this problem has the value of ``operators-count'' as zero to indicate that the initial macro under construction is empty.
In this experiment, we choose to build the randomly created valid macros such that each macro has three unique operators.
To ensure the macro-building task returns macros with distinct operators, we supplemented the meta operators with axillary atoms to work as interlocks to prevent an operator from being scheduled twice.

Randomly created invalid macros are produced from the randomly created macros by randomly selecting a precondition, and an add effect from the valid macro and swapping them. 

\BlocksworldModifications
\Cref{tab:BlocksworldModifications} lists the description of the modifications applied to the Gripper domain to produce its modified versions. 
The table also states if the modified version is expected to be functionally equivalent to the original domain. 
In theory, adding hand-crafted and randomly created valid macros should not change the reach sets of the modified version from the reach set of its original domain.
Hence the modified version in these cases and their original domains must be functionally equivalent.
On the contrary, the other modifications are expected to produce versions with reach sets different from the reach sets of their original domains.

The modifications to the other domains are detailed in \Cref{sec:ExperimentsModifications}.
Note that the test generation tool did not generate versions of the Gripper, Child-snack and Logistics domains with swapped variables because these domains do not have any predicate with two variables of the same type.
The variables of an atom must have the same types to be swapped by our tool because an atom with swapped variables of different types will not match its predicate definition.
A domain with an atom that does not match a defined predicate would be considered invalid.
Thus, to swap variables of different types in an atom, we would have to add a new predicate to match the atom with the swapped variables.
If we add a new predicate to the modified version of the original domain, then the two domains, the original and its modified version, will have a different number of predicates.
Hence, our tool will judge the two domains as not functionally equivalent without any proper investigation as the scope of our method is limited to domains with an equal number of predicates.
Therefore, the condition that an atom's variables must be of the same type to be swapped is essential to ensure the produced validation test is meaningful.

The results in \Cref{sub:results} report the verdict of our tool on the functional equivalence of each domain and its modified versions.
In addition, the results provide the CPU time spent by the planner and the SMT solver for each validation task. 
The experiments were run on a computer node from BlueCrystal Phase 4 in the Advanced Computing Research Centre at the University of Bristol\footnote{\url{https://www.bristol.ac.uk/acrc/}}. The used computer node has two 14 core 2.4 GHz Intel E5-2680 v4 (Broadwell) CPUs and 128 GiB of RAM. Each validation test was limited to one CPU with 20 GiB of RAM.
The domains used in these experiments and their modified versions, along with the output of our tool for validating the functional equivalence of these domains, are available online\footnote{\url{https://github.com/Anas-Shrinah}}.

\subsection{Results and discussion} %
\label{sub:results}
\Cref{tab:GripperResults,tab:BlocksworldResults,tab:ElevatorResults,tab:ParkingResults,tab:HikingResults,tab:FloortileResults,tab:ChildsnackResults,tab:LogisticsResults,tab:CavedivingResults,tab:RoverResults,tab:PipesworldResults,tab:ScanalyzerResults,tab:FreecellResults} provide the results of our experiments.
These experiments show that our method decided on the functional equivalence of all simple domains and their versions and on functionality equivalent complex domains and their versions in less than 43 seconds.
Furthermore, our tool took just a few seconds to terminate with the output no conclusive verdict can be reached for validating the functional equivalence between the complex domains Elevator, Floor-tile, Child-snack and Pipesworld and their versions that are supposed to be not functionally equivalent to their original domains. 

Our tool produced correct verdicts on the functional equivalence of all 75 validation tasks.
These verdicts matched our expectations apart from three tests: Hiking with swapped variables, Hiking with deleted operators and Scanalyzer with swapped variables.

The version of the Hiking domain with swapped variables is produced from the original Hiking domain by swapping the positions of the variable ?x3 with the variable ?x5 in the parameters of the precondition ``(partners ?x6 ?x3 ?x5)'' in the operator ``walk\_together\_M''.
This operator is depicted in \Cref{fig:walkTogetherM}.
\begin{figure}
\begin{center}
\begin{Verbatim}[commandchars=\\\{\}]
(:action walk_together_M
 :parameters (?x1 - tent ?x2 ?x4 - place ?x3 ?x5 - person ?x6 - couple)
 :precondition (and (at_tent ?x1 ?x2) (up ?x1) (at_person ?x3 ?x4)
   (next ?x4 ?x2) (at_person ?x5 ?x4) 
   (walked ?x6 ?x4) \userinput{(partners ?x6 ?x5 ?x3)})
 :effect (and (at_person ?x3 ?x2) (at_person ?x5 ?x2) (walked ?x6 ?x2)
  (not (at_person ?x3 ?x4)) (not (at_person ?x5 ?x4)) 
  (not (walked ?x6 ?x4))))
\end{Verbatim}
\end{center}
\caption{The operator ``walk\_together'' from the version of the Hiking domain with swapped variables.}
\label{fig:walkTogetherM}
\end{figure}
To clarify why our method found the Hiking domain to be functionally equivalent to its version with swapped variables, we have to compare this validation task to a similar one where our tool produced a non-functional equivalence verdict.
For this purpose, we choose the validation task of testing the functional equivalence of the Blocksworld domain and its version with swapped variables.

In the version of the Hiking domain with swapped variables, the variables ?x3 and ?x5 are symmetric in the atoms of the operator ``walk\_together\_M''.
Two variables are symmetric if they appear in the same positions of the same atoms with the same other variables.
In ``walk\_together\_M'', both variables ?x3 and ?x5 are the first variables in the preconditions and delete effects (at\_person ?x3 ?x4) and (at\_person ?x5 ?x4), and the second variable in both preconditions and delete effects is ?x4. 
Moreover, both variables ?x3 and ?x5 are the first variables in the add effects (at\_person ?x3 ?x2) and (at\_person ?x5 ?x2), and the second variable in both add effects is ?x2. 

The impact of swapping ?x3 and ?x5 in the operator ``walk\_together\_M'' can be observed when both domains, the original one and the modified version, are used in a planning task where the operators ``walk\_together'' from the original domain and ``walk\_together\_M'' from the modified domain are instantiated with the same objects and scheduled to produce the same transition. 
For example, let Anas and Lujain be the persons who make the couple1.
Assume this couple has the tent1 and are at Snuff Mills, and they want to walk together to Oldbury Court.
From this initial state, a planner using the original domain will produce this plan:
\begin{verbatim}
walk_together (tent1, Snuff Mills, Oldbury Court, Anas, Lujain, couple1)
\end{verbatim}
On the other hand, form the same initial state, the same planner using the modified domain will produce this plan:
\begin{verbatim}
walk_together (tent1, Snuff Mills, Oldbury Court, Lujain, Anas, couple1)
\end{verbatim}
The only difference will be in the order of the objects in the parameters of the two operators.
Nevertheless, both domains can be used to reach the same end state for this initial state.
Thus, the two domains are functionally equivalent.

The aim of this example is not to prove that swapping the position of symmetric variables in the parameters of an operator does not change the reach set of the operator's domain. 
This property is valid according to \Cref{th:operatorsStructurReachSet}.
This example is meant to provide a concrete example of this case and show the impact of this modification. 

This is not the case for the original Blocksword domain and its modified version.
The version of the Blocksword domain with swapped variables is produced from the original Blocksword domain by swapping the positions of the variable ?x with the variable ?y in the parameters of the add effect ``(on ?y ?x)'' in the operator ``stack\_M''.
This operator is depicted in \Cref{fig:StackM}.
\begin{figure}
\begin{center}
\begin{Verbatim}[commandchars=\\\{\}]
(:action stack_M
 :parameters (?x ?y - block )
 :precondition (and (holding ?x) (clear ?y))
 :effect (and (clear ?x) (handempty) \userinput{(on ?y ?x)} 
  (not (holding ?x)) (not (clear?y))))
\end{Verbatim}
\end{center}
\caption{The operator ``stack'' from the version of the Blocksworld domain with swapped variables.}
\label{fig:StackM}
\end{figure}
Note that the variables ?x and ?y are not symmetric in the atoms of the operator ``stack\_M''.
For instance, ?x is the first and only variable in the precondition (holding ?x), and ?y is the first and only variable in the precondition (clear ?y). 

The impact of swapping ?x and ?y in the operator ``stack\_M'' can be observed when both the original domain and the modified version are used in a planning task where the operators ``stack'' from the original domain and ``stack\_M'' from the modified domain are instantiated with the same objects and scheduled to produce the same transition. 
Assume two blocks: $B_1$ and $B_2$ and we want to stack $B_1$ on top of $B_2$.
According to the original domain, a robot should hold $B_1$, and $B_2$ must be clear. Then the action ``stack $B_1$ $B_2$ '' will cause $B_1$ to be on $B_2$.
However, from the same initial state and according to the modified domain, applying the action ``stack\_M $B_1$ $B_2$ '' will cause $B_2$ to be on $B_1$.
Note that ``stack\_M $B_2$ $B_1$ '' is not applicable from the assumed initial state because this action requires the robot to hold $B_2$ and $B_1$ must be clear, which is the opposite of the initial state in this example.
Thus, both domains cannot be used to reach the same end state for this initial state.
Hence, the original Blocksworld domain and its version with swapped variables are not functionally equivalent. 

The same discussion is applicable for the validation task of the functional equivalence between the original Scanalyzer domain and its version with swapped variables.

The last case where our tool gave an unexpected outcome is when validating the functional equivalence between the original Hiking domain and its version with a deleted operator. 
The version of the Hiking domain with deleted operator is produced from the original Hiking domain by deleting the operator ``drive\_tent\_passenger''.
Intuitively, deleting an operator from a domain is expected to produce a domain that is not functionally equivalent to its original domain. 
However, this was a different case.
It turned out the operator ``drive\_tent\_passenger'' is a macro in the original domain.
Any transition produced by this operator can be matched by the sequence $\langle$``drive\_tent'', ``drive'', ``drive\_passenger''$\rangle$.
In other words, any state reachable by the operator ``drive\_tent\_passenger'' is also reachable by this sequence.
Thus, removing this operator would not render any reachable state unreachable.
Hence, this operator is redundant from the perspective of domain reach sets. 
Therefore, the original Hiking domain and its version with the deleted operator are functionally equivalent.

Observe that the planning time took just a fraction of seconds to complete the tasks of testing whether the operators of each domain and its modified versions are not macros and proving the non-macros are primitive operators when disproving the functional equivalence of simple domains.
This observation is interesting because these planning tasks aim to prove their planning problems unconsolable.
It is known that proving unsolvability is much harder than finding a solution to planning problems in general.
However, our planning problems are different.
In the first problem, we search for macros with the same reach set of each operator in both domains.
As discussed in \Cref{sec:checkMacro}, the first step in this search process is to remove any operator that does not satisfy three rules.
Thanks to these rules, some search tasks will have domains without operators or just a few operators. 
This reduction in the number of operators in the meta-planning problems makes proving their unsolvability easy.
Moreover, even in the case of the domains with valid macros, solving the meta-planning problem and detecting macros also took a fraction of a second. 
This impressive performance is also attributed to the reduction of unrelated operators before commencing the meta-planning task.

The same observation and justification apply to the problem of finding SOSOs for the operators of the simple domains when disproving their functional equivalence.
The first step in these search tasks is also to remove unrelated operators that cannot be part of any SOSO of the given operators as described in \Cref{sec:findSOSOsForO}.

The other general observation is that the time required for the Z3 solver to solve our variable mapping SMT problem increases as the complexity of the planning domain models involved in the validation task increases.
This behaviour is anticipated as the number of SMT constraints increases as the number of operators, their atoms, and the number of the variables of the atoms increases.

\GripperResults
\ReasonsDecoder
\BlocksworldResults
\ElevatorResults
\ParkingResults
\HikingResults
\FloortileResults
\ChildsnackResults
\LogisticsResults
\CavedivingResults
\RoverResults
\PipesworldResults
\ScanalyzerResults
\FreecellResults

\section{Conclusions and future work} %
\label{sec:conclusion_and_future_work}
Validating the functional equivalence of planning domain models has many applications in KEPS, yet this topic has not received much attention from researchers.
In this paper, we formally defined the concept of functional equivalence between planning domain models.
Additionally, we introduced D-VAL, an automatic functional equivalence validation tool that uses a planner and an SMT solver to validate the functional equivalence of planning domain models.
The soundness and completeness of our approach are theoretically proven.

For functionally equivalent domains, D-VAL finds a consistent predicate mapping; under this mapping, the domains are functionally redundant.
On the other hand, for not functionally equivalent domains, the output of D-VAL depends on the type of the given domains.
For simple domains that are not functionally equivalent, D-VAL can produce a conclusive verdict of non-functional equivalence.
On the contrary, D-VAL cannot return a decisive judgment of non-functional equivalence for complex domains that are not functionally equivalent. 
Extending D-VAL to provide conclusive verdicts for complex domains is a key future work. 

Our experimental evaluation confirms the feasibility of our method and shows that D-VAL can validate the functional equivalence of all simple domains in less than 45 seconds. 
These domains include complicated domains such as Scanlayser and Free-cell domains.
The impressive performance of D-VAL in validating the functional equivalence of 13 planning domain models from the IPC and their modified versions motivates the next step of extending the experiments to evaluate the scalability of D-VAL for validating the functional equivalence of industrial scale domains.

The scope of D-VAL is limited to planning domain models with an equal number of predicates of equal arities.
This restriction limits the applications of D-VAL.
Nevertheless, this paper opens new research avenues by formalising and solving a slightly restricted version of the problem of validating the functional equivalence of planning domain models. 
As future work, we intend to explore the possibilities of extending the scope of D-VAL to cover domains with an unequal number of predicates. 

In addition, we aim to develop D-VAL to return a distance between the given domains if they are found to be not functionally equivalent. 
When the variable mapping SMT problem of a functional equivalence validation task is found to be unsatisfiable, we could find the maximum satisfiability core for this problem and then use it to calculate a distance between the two domains.
Another ambitious research direction related to the idea of domains' functional distance is to investigate how we can improve D-VAL to recommend corrections to the given domains to make them functionally equivalent. 

To deploy D-VAL, its implementation must be enhanced to perform preprocessing checks to test the validity and suitability of the given domains to its scope before testing their functional equivalence. 
Additionally, given the proven theoretical base of D-VAL and its critical role as a validation tool, its implementation must be verified to increase confidence in its correctness.
\renewcommand{\theHsection}{A\arabic{section}}
\appendix
\section[Simple Domains Reachability Theorem]{Proof of Theorem~\ref{th:SimpleDomainsReachabilityTheorem} (See page~\pageref{th:SimpleDomainsReachabilityTheorem})}
\label{ProofOfTheoremRT}

To prove this theorem, we first prove \Cref{PrimitiveNoOperatorNFELemma} and \Cref{reachabilityLemma} in \Cref{ProofOfLemmas} and then the main theorem is proven in \Cref{ProofOfTheoremRTusingRL}.

\subsection[Simple Domains Reachability Lemmas]{Proof of the Lemmas of \Cref{th:SimpleDomainsReachabilityTheorem}}
\label{ProofOfLemmas}

\ReachSetOneOperatorSubsetUnionSEOperator*
\begin{proof}
This lemma implies that for the reach set of $o$ to be a subset of the union of the reach sets of multi operators $\{o'_1,o'_2,\dots,o'_n\}$ from \D2, these operators must share the same effects,~i.e. they must be in the same \SE set.

For any operator $o'$ from \D2 to produce some of the transitions that start from states where all the atoms in the add effects of $o$ are false and all the delete effects of $o$ are true, the add and delete effects of $o'$ must contain images of all effects of $o$ under $F_p$.
We call these transitions the core transitions of $o$.
Since these operators have the same delete effects as $o$, they also must have the same preconditions that appear as delete effects in $o$.
We call these preconditions the necessary preconditions and other preconditions the optional preconditions of $o$. 

Suppose an operator $o'$ from \D2 has the same add effects as the operator $o$, and the preconditions of $o'$ contain the preconditions of $o$.
Moreover, assume $o'$ has the same delete effects as $o$ in addition to one more delete effect $d'$.
The intersection of the reach set of $o'$ and $o$ under $F_p$ is empty because applying $o$ to the states where $d'$ is true will produce states where $d'$ is true, whereas $o'$ will produce states where $d'$ is false. 
On the other hand, applying $o$ to the states where $d'$ is false will produce states where $d'$ is false, whereas $o'$ is not applicable in such states because the delete effect $d'$ is also a precondition of $o'$.
Therefore, if an operator $o'$ from \D2 has more delete effects than the images of the delete effects of $o$ under $F_p$, the reach set of $o'$ will not contain any of the transitions of $o$.

Now suppose an operator $o'$ from \D2 has the same delete effects as the operator $o$, and the preconditions of $o'$ contain the preconditions of $o$.
Moreover, assume $o'$ has the same add effects as $o$ in addition to one more add effect $d'$.
The intersection of the reach set of $o'$ and $o$ under $F_p$ is half of the reach set of $o$ because applying $o$ to the states where $d'$ is false will produce states where $d'$ is false, whereas $o'$ will produce states where $d'$ is true. 
Thus, any transitions of $o$ from states where $d'$ is false are not in the reach set of $o'$.
Note that the other half of the reach set of $o$, which contains the transitions from the states where $d'$ is true, is also contained in the reach set of $o'$.
Therefore, if an operator $o'$ from \D2 has more add effects than the images of the add effects of $o$ under $F_p$, the reach set of $o'$ will only contain part of the transitions of $o$.

Thus, the reach set of $o$ is guaranteed not to be a subset of the union of the reach sets of any set of operators from \D2 that do not have only the images of the effects of $o$ as their effects.
These arguments prove that for the image of the reach set of an operator from one domain to be a subset of the union of the reach sets of multi operators from another domain under a predicate mapping, these operators must share the same effects,~i.e. they must be in the same \SE set.
\end{proof}
\PrimitiveNoOperatorNFELemma*
\begin{proof}
From the antecedent of this lemma we have
\begin{multline} 
\exists o \in Primitive(O_1), \forall Seq' \subseteq SEQ_2(\Gamma(F_p(o)) \not = \bigcup_{ seq' \in Seq'}(\Gamma(seq')) \implies \\ \ \Gamma(F_p(o)) \not \subseteq \bigcup_{ seq' \in Seq'}(\Gamma(seq')) \vee \bigcup_{ seq' \in Seq'}(\Gamma(seq')) \not \subseteq \Gamma(F_p(o))) \label{eq:L1_1}
\end{multline}
Since $a \vee b \iff a \vee ( \bar{a } \wedge b)$, we transform \Cref{eq:L1_1} to
\begin{multline}
\exists o \in Primitive(O_1), \forall Seq' \subseteq SEQ_2( \ \Gamma(F_p(o)) \not \subseteq \bigcup_{ seq' \in Seq'}(\Gamma(seq')) \ \vee \bigcup_{ seq' \in Seq'}(\Gamma(seq')) \not \subseteq \Gamma(F_p(o)) \implies \\ \Gamma(F_p(o)) \not \subseteq \bigcup_{ seq' \in Seq'}(\Gamma(seq'))\ \vee \\ ( \Gamma(F_p(o)) \subseteq \bigcup_{ seq' \in Seq'}(\Gamma(seq')) \wedge \bigcup_{ seq' \in Seq'}(\Gamma(seq')) \not \subseteq \Gamma(F_p(o)) ) )
\end{multline}
Thus, to prove this lemma, we have to prove
\begin{multline}
\exists o \in Primitive(O_1), \forall Seq' \subseteq SEQ_2( \Gamma(F_p(o)) \not \subseteq \bigcup_{ seq' \in Seq'}(\Gamma(seq'))\ \vee \\ ( \Gamma(F_p(o)) \subseteq \bigcup_{ seq' \in Seq'}(\Gamma(seq')) \wedge \bigcup_{ seq' \in Seq'}(\Gamma(seq'))\not \subseteq \Gamma(F_p(o)) ) \implies \Gamma(F_p(D_1)) \neq \Gamma(D_2) ) \label{eq:L1_disjunction}
\end{multline}
We break the disjunction in \Cref{eq:L1_disjunction} into two implications in \Cref{eq:L1_firstImplication} and \Cref{eq:L1_secondtImplication}.
Then we prove each one separately.
First we prove 
\begin{equation}
\exists o \in Primitive(O_1), \forall Seq' \subseteq SEQ_2( \Gamma(F_p(o)) \not \subseteq \bigcup_{ seq' \in Seq'}(\Gamma(seq'))\ \implies \Gamma(F_p(D_1)) \neq \Gamma(D_2) ) \label{eq:L1_firstImplication}
\end{equation} 
If $Seq' = SEQ_2$ then $\bigcup_{ seq' \in Seq'}(\Gamma(seq')) = \Gamma(D_2)$. From \Cref{eq:L1_firstImplication}, $\Gamma(F_p(o)) \not \subseteq \bigcup_{ seq' \in Seq'}(\Gamma(seq'))$. Thus, $\Gamma(F_p(o)) \not \subseteq \Gamma(D_2)$. Hence, $\Gamma(o) \not \subseteq \Gamma(F_p^{-1}(D_2))$. However, we know $\Gamma(o) \subseteq \Gamma(D_1)$. Thus, $\Gamma(D_1) \not \subseteq \Gamma(F_p^{-1}(D_2))$. Therefore, $\Gamma(F_p(D_1)) \not \subseteq \Gamma(D_2)$.
This conclusion completes the proof of \Cref{eq:L1_firstImplication}.
\end{proof}
Before we prove the second implication of this lemma, we prove the following implication.
\textit{If the image of reach set of any primitive operator from \D1 under $F_p$ is not equal to the reach set of any operator from \D2, then the image of the reach set of \D1 under $F_p$ is not equal to the reach set of \D2.}
\begin{proof}
\begin{equation}
\forall o \in Primitive(O_1), \forall Seq' \subseteq SEQ_2( \Gamma(F_p(o)) \neq \bigcup_{ seq' \in Seq'}(\Gamma(seq'))) \implies \Gamma(F_p(D_1)) \neq \Gamma(D_2) \label{eq:NotEqual_NotEqual}
\end{equation}
\begin{multline}
\forall o \in Primitive(O_1), \forall Seq' \subseteq SEQ_2( \Gamma(F_p(o)) \not \subseteq \bigcup_{ seq' \in Seq'}(\Gamma(seq')) \ \wedge \\ \bigcup_{ seq' \in Seq'}(\Gamma(seq')) \not \subseteq \Gamma(F_p(o)) ) \implies \Gamma(F_p(D_1)) \neq \Gamma(D_2) \label{eq:conjunction_implication}
\end{multline}
\begin{multline}
\forall o \in Primitive(O_1), \forall Seq' \subseteq SEQ_2( \Gamma(F_p(o)) \not \subseteq \bigcup_{ seq' \in Seq'}(\Gamma(seq')) \implies \Gamma(F_p(D_1)) \neq \Gamma(D_2) \ \vee \\ \bigcup_{ seq' \in Seq'}(\Gamma(seq')) \not \subseteq \Gamma(F_p(o)) \implies \Gamma(F_p(D_1)) \neq \Gamma(D_2)) \label{eq:NotEqDisjunction}
\end{multline}
Split disjunction in \Cref{eq:NotEqDisjunction} into two implications: 
\begin{multline}
\forall o \in Primitive(O_1), \forall Seq' \subseteq SEQ_2( \Gamma(F_p(o)) \not \subseteq \bigcup_{ seq' \in Seq'}(\Gamma(seq')) \implies \Gamma(F_p(D_1)) \neq \Gamma(D_2)) \label{eq:NotEqDisjunction-1}
\end{multline}
\begin{multline}
\forall o \in Primitive(O_1), \forall Seq' \subseteq SEQ_2( \bigcup_{ seq' \in Seq'}(\Gamma(seq')) \not \subseteq \Gamma(F_p(o)) \implies \Gamma(F_p(D_1)) \neq \Gamma(D_2)) \label{eq:NotEqDisjunction-2}
\end{multline}
Proving either of \Cref{eq:NotEqDisjunction-1} or \Cref{eq:NotEqDisjunction-2} will prove \Cref{eq:NotEqDisjunction} which proves \Cref{eq:conjunction_implication} then we can prove \Cref{eq:NotEqual_NotEqual}.
Note that we have already proved \Cref{eq:NotEqDisjunction-1} which is the first implication of this lemma. 
\Cref{eq:NotEqDisjunction-1} implies \Cref{eq:NotEqDisjunction-1}. 
These arguments form the proof of \Cref{eq:NotEqual_NotEqual}. 
\end{proof}
Next, we prove the second implication this lemma. 
\begin{proof}
\begin{multline}
\exists o \in Primitive(O_1),\forall Seq' \subseteq SEQ_2( \Gamma(F_p(o)) \subseteq \bigcup_{ seq' \in Seq'}(\Gamma(seq'))\ \wedge \bigcup_{ seq' \in Seq'}(\Gamma(seq')) \not \subseteq \Gamma(F_p(o)) \\ \implies \Gamma(F_p(D_1)) \neq \Gamma(D_2) ) \label{eq:L1_secondtImplication}
\end{multline}
We start the proof from the contrapositive form of \Cref{eq:NotEqual_NotEqual} which is proved above. 
\begin{equation}
\Gamma(F_p(D_1)) = \Gamma(D_2) \implies \exists o_o \in Primitive(O_1), \exists Seq' \subseteq SEQ_2( \Gamma(F_p(o)) = \bigcup_{ seq' \in Seq'}(\Gamma(seq'))) \label{eq:NotEqual_NotEqual_contra}
\end{equation}
For the sake of contradiction, assume $\Gamma(F_p(D_1)) = \Gamma(D_2)$. Hence from \Cref{eq:NotEqual_NotEqual_contra}, we have 
\begin{equation}
\exists o_o \in Primitive(O_1), \exists Seq' \subseteq SEQ_2( \Gamma(F_p(o_o)) = \bigcup_{ seq' \in Seq'}(\Gamma(seq'))) \label{eq:o_o=o'}
\end{equation}
Thus,
\begin{equation}
\exists o_o \in Primitive(O_1), \exists Seq' \subseteq SEQ_2( \Gamma(F_p(o_o)) \subseteq \bigcup_{ seq' \in Seq'}(\Gamma(seq')) \ \wedge \bigcup_{ seq' \in Seq'}(\Gamma(seq')) \subseteq \Gamma(F_p(o_o)) ) \label{eq:o_o}
\end{equation}
The antecedent of \Cref{eq:L1_secondtImplication} is
\begin{equation}
\exists o \in Primitive(O_1), \forall Seq' \subseteq SEQ_2( \Gamma(F_p(o)) \subseteq \bigcup_{ seq' \in Seq'}(\Gamma(seq')) \ \wedge \bigcup_{ seq' \in Seq'}(\Gamma(seq')) \not \subseteq \Gamma(F_p(o))) \label{eq:o}
\end{equation}
Note that $o_o$ in \Cref{eq:o_o} and $o$ in \Cref{eq:o} are different operators.
Therefore,
\begin{equation}
\exists o, \exists o_o \in Primitive(O_1), \exists Seq' \subseteq SEQ_2( \Gamma(F_p(o)) \subseteq \bigcup_{ seq' \in Seq'}(\Gamma(seq')) \ \wedge \bigcup_{ seq' \in Seq'}(\Gamma(seq')) \subseteq \Gamma(F_p(o_o)))
\end{equation}
Hence,
\begin{equation}
\exists o, \exists o_o \in Primitive(O_1)( \Gamma(F_p(o)) \subseteq \Gamma(F_p(o_o))) \label{eq:o=o_o}
\end{equation}
The operators $o$ and $o_o$ are primitive operators and \Cref{eq:o=o_o} states the reach set of $o$ is a subset of the reach set of $o_o$.
Therefore, \Cref{eq:o=o_o} contradicts the definition of primitive operators which requests the reach set of a primitive operator not to be a subset of the reach set of another primitive operator.
This contradiction proves $\Gamma(F_p(D_1)) \not = \Gamma(D_2)$ 

Proving the implications \Cref{eq:L1_firstImplication} and \Cref{eq:L1_secondtImplication} proves \Cref{eq:L1_disjunction} which in turns proves this lemma.
\end{proof}
\RL*
To prove this lemma, the biconditional statement is broken into forward and backward implications which will be proven separately.
For brevity, the set {\it Obj} is dropped from the reach set symbols in the proofs as it is common to all reach sets.

\begin{proof}
First, we prove the backward implication: 
\begin{multline}
\Gamma(F_p(D_1)) = \Gamma(D_2) \rightarrow \forall o \in Primitive(O_1), \exists o' \in Primitive(O_2) (\\
\Gamma(F_p(o)) = \Gamma(o') \wedge |Primitive(O_1)| = |Primitive(O_2)|)
\end{multline}
The contrapositive of the forward implication is:
\begin{multline}
\exists o \in Primitive(O_1), \forall o' \in Primitive(O_2) (\Gamma(F_p(o)) \not = \Gamma(o')) \vee \\ |Primitive(O_1)| \not = |Primitive(O_2)| \rightarrow \Gamma(F_p(D_1)) \not = \Gamma(D_2) \label{eq:ForwardImplicationContraositive}
\end{multline}
We prove this implication starting from $\exists o \in Primitive(O_1), \forall o' \in Primitive(O_2) (\Gamma(F_p(o)) \not = \Gamma(o'))$. 
Then from $|Primitive(O_1)| \not = |Primitive(O_2)|$.
So, we first prove
\begin{equation}
\exists o \in Primitive(O_1), \forall o' \in Primitive(O_2) (\Gamma(F_p(o)) \not = \Gamma(o')) \rightarrow \Gamma(F_p(D_1)) \not = \Gamma(D_2) \label{eq:ForwardImplicationContraositive_part1}
\end{equation}
If the image of the reach set of $o$ is not equal to the reach set of a primitive operator from \D2, then it must be either equal or not equal to the union of the reach sets of the sequences of operators from a subset $Seq'$ from $SEQ_2$.
In the former case, if the reach sets of the primitive operators that make the sequences of operators in $Seq'$ are equal to the union of the images of the reach sets of the sequences of operators from a subset $Seq$ from $SEQ_1$, then the reach set of $o$ will also be equal to the union of the reach sets of the sequences of operators from the set $Seq$.
Thus, the operator $o$ will be a redundant operator because any transition in the reach set of $o$ can be made by a sequence of operators from $O_1$. 
However, the operator $o$ is a primitive operator as per the antecedent of this implication; therefore, $o$ is not redundant in \D1.
This contradiction proves that if the image of the reach set of the primitive operator $o$ is equal to the union of the reach sets of the sequences of operators from a subset $Seq'$ from $SEQ_2$, then the reach sets of the primitive operators from \D2 that make the sequences of operators in $Seq'$ are not equal to the union of the images of the reach sets of the sequences of operators from a subset $Seq$ from $SEQ_1$.
Hence, as per \Cref{PrimitiveNoOperatorNFELemma}, the reach set of \D2 is not equal to the image of the reach set if \D1.

In the latter case, if the reach set of any operator from the primitive operators that make the sequences of operators in $Seq'$ is not equal to the union of the images of the reach sets of the sequences of operators from a subset $Seq$ from $SEQ_1$, then as per \Cref{PrimitiveNoOperatorNFELemma} the reach set of \D2 is not equal to the image of the reach set of \D1.

These arguments conclude the proof of the first part of the backward implication of this theorem.

Then we prove the second part of the implication starting from $|Primitive(O_1)| \not = |Primitive(O_2)|$. 
\begin{equation}
|Primitive(O_1)| \not = |Primitive(O_2)|\rightarrow \Gamma(F_p(D_1)) \not = \Gamma(D_2) \label{eq:ForwardImplicationContraositive_part2}
\end{equation}
If two domains have a different number of primitive operators, then one domain will have at least one primitive operator more than the other domain. 
From the definition of primitive operators, we know the reach set of a primitive operator is not equal to the reach set of any other primitive operators in the same domain.
Moreover, from the proof of the first part of the contrapositive form of the backward implication of this lemma, we know If the reach set of a primitive operator in one domain is not equal to the reach set of a primitive operator in the other domain, then the two domains do not have equal reach sets.
Thus, if the reach set of each primitive operator from the domain with the fewer primitive operators is equal to the reach set of a primitive operator from the other domain, then the reach set of one primitive operator from the domain with the greater number of primitive operators cannot be equal to the reach set of any primitive operator from the other domain. 
Therefore, two domains with an unequal number of primitive operators are guaranteed to have unequal reach sets.
This concludes the proof of the second part of the backward implication. Thus, we prove the backward implication of this lemma.
Second, we prove the forward implication:
\begin{multline}
\forall o \in Primitive(O_1), \exists o' \in Primitive(O_2) (\Gamma(F_p(o)) = \Gamma(o')) \wedge \\ |Primitive(O_1)| = |Primitive(O_2)| \rightarrow \Gamma(F_p(D_1)) = \Gamma(D_2) \label{eq:backwordImplicationLemma}
\end{multline}

Let $D_3 = (P_2, O_3)$ be a planning domain model with the same set of predicates as $D_2$, and the set of its primitive operators $Primitive(O_3)$ satisfies the following conditions:
\begin{gather}
Primitive(O_3) \subseteq Primitive(O_2) \\ \nonumber
\wedge \\
\forall o \in Primitive(O_1), \exists o' \in Primitive(O_3) (\Gamma(F_p(o)) = \Gamma(o') \nonumber
\end{gather}

From the definition of $O_3$, we have $Primitive(O_3) \subseteq Primitive(O_2)$; thus, any sequence of operators that can be made by the primitive operators of $O_3$ can also be made by the primitive operators of $D_2$.
Since the reach set of a domain is equal to the union of the reach sets of its sequence of operators, we have $\Gamma(D_3) \subseteq \Gamma(D_2)$.

Furthermore, the definition of $O_3$ implies the image under $F_p$ of the reach set of each primitive operator in $D_1$ is equal to the reach set of a primitive operator from $D_3$.
This means the reach set of any sequence of operators made by the primitive operators of $D_1$ is equal to the reach set of a sequence of operators which can be made from the primitive operators of $D_3$.
Thus we have $\Gamma(F_p(D_1)) \subseteq \Gamma(D_3)$. Therefore, $\Gamma(F_p(D_1)) \subseteq \Gamma(D_2)$.

From the antecedent of \Cref{eq:backwordImplicationLemma}, since the reach set of each primitive operator in $O_1$ is equal to the reach set of a primitive operator in $O_2$; because $O_1$ and $O_2$ have an equal number of primitive operators; and as the reach set of a primitive operator is not equal to the reach set of any other primitive operators in the same domain, we have the reach set of each primitive operator in $O_2$ is equal to the reach set of a primitive operator in $O_1$.
\begin{equation}
\forall o' \in Primitive(O_2), \exists o \in Primitive(O_1)(\Gamma(o') = F_p(o))
\end{equation}

Let $D_4 = (P_1, O_4)$ be a planning domain model with the same set of predicates as $D_1$, and the set of its primitive operators $Primitive(O_4)$ satisfies the following conditions:
\begin{gather}
Primitive(O_4) \subseteq Primitive(O_1) \nonumber \\ 
\wedge \\
\forall o' \in Primitive(O_2), \exists o \in Primitive(O_4) (\Gamma(o') = \Gamma(F_P(o)) \nonumber
\end{gather}
With the help of $D_4 $ we can prove $\Gamma(D_2) \subseteq \Gamma(F_p(D_1))$ in a similar way as we proved $\Gamma(F_p(D_1)) \subseteq \Gamma(D_2)$.
Therefore, we prove $\Gamma(F_p(D_1)) = \Gamma(D_2)$.
Proving the forward implication concludes the proof of this lemma.
\end{proof}
\subsection[Simple Domains Reachability Theorem]{Proof of Theorem~\ref{th:SimpleDomainsReachabilityTheorem} using Lemma~\ref{reachabilityLemma} (See page~\pageref{th:SimpleDomainsReachabilityTheorem})}
\label{ProofOfTheoremRTusingRL}
\RT*

Let $\mathbb{F}$ be the set of all bijective mappings between predicates of equal arities from the predicates of every primitive operator in $D_1$ to the predicates of every primitive operator in $D_2$.
\begin{alignat}{1}
\mathbb{F} = \{ f_p | f_p: Predicates(o) \twoheadrightarrowtail Predicates(o') \ &\text{ where } o \in Primitive(O_1), o' \in Primitive(O_2) \nonumber \\
\text { and if } f_p(p) = p' \text{ then } Arity(p) = Arity(p') \}\nonumber 
\end{alignat}

Let $R_{OM}$ be a relation between primitive operators from $D_1$ and predicate mappings from $\mathbb{F}$. A primitive operator $o$ from $D_1$ is related to a mapping $f_p$ from $\mathbb{F}$ by $R_{OM}$ means there exists a primitive operator $o'$ from $D_2$ such that the reach set of $o$ is equal to the reach set of $o'$ under the mapping $f_p$.
\begin{equation}
R_{OM} = \{ (o,f_p) \in Primitive(O_1) \times \mathbb{F} \ | \ \exists o' \in Primitive(O_2), \ \Gamma(f_p(o),Obj) = \Gamma(o',Obj) \} \label{eq:Rom}
\end{equation}
\begin{proof}
To prove this theorem, the biconditional statement is broken into forward and backward implications which will be proven separately.
First, we prove the forward implication: 
\begin{multline}
\Gamma(F_p(D_1),Obj) = \Gamma(D_2,Obj) \implies
 \exists R'_{om} \subseteq R_{OM}(Domain(R'_{om}) = Primitive(O_1) \\ \wedge (F = \bigcup\limits_{f_p \in Range(R'_{om}) } f_p )\text{ is a well-defined function}) \wedge |Primitive(O_1)| =|Primitive(O_2)| \nonumber
\end{multline}
For brevity, the set {\it Obj} is dropped from the reach set symbols in the proofs as it is common to all reach sets.
From the antecedent of this implication and from \Cref{reachabilityLemma}, we have:
\begin{multline}
\Gamma(F_p(D_1)) = \Gamma(D_2) \implies \forall o \in Primitive(O_1), \exists o' \in Primitive(O_2) (\Gamma(F_p(o)) = \Gamma(o')) \\ \wedge |Primitive(O_1)| =|Primitive(O_2)| \label{eq:th_rt_forward_1}
\end{multline}
This proves if the reach set of $F_p(D_1)$ is equal to the reach set of $D_2$, then both $D_1$ and $D_2$ have an equal number of primitive operators. 

From the definition of the relation $R_{OM}$ (\Cref{eq:Rom}) and \Cref{eq:th_rt_forward_1}, we can infer 
\begin{alignat}{1}
\Gamma(F_p(D_1)) = \Gamma(D_2) \implies \forall o \in Primitive(O_1), \ (o,F_p) \in R_{OM} \label{eq:th_rt_forward_2}
\end{alignat}
Let the relation $R'_{om}$ be defined as follows.
\begin{alignat}{1}
R'_{om} = \{ (o,F_p) | \forall o \in Primitive(O_1) \}\label{eq:th_rt_forward_3}
\end{alignat}
From \Cref{eq:th_rt_forward_2}, we deduce that $R'_{om}$ is a subset of $R_{OM}$.
Furthermore, from the definition of $R'_{om}$ in \Cref{eq:th_rt_forward_3}, we have $Domain(R'_{om}) = Primitive(O_1)$.
Thus, we conclude 
\begin{alignat}{1}
\Gamma(F_p(D_1)) = \Gamma(D_2) \implies \exists R'_{om} \subseteq R_{OM} \wedge Domain(R'_{om}) = Primitive(O_1)\label{eq:th_rt_forward_4}
\end{alignat}
From \Cref{eq:th_rt_forward_3}, we have
\begin{alignat}{1}
Range(R'_{om}) = F_p
\end{alignat}
So, 
\begin{alignat}{1}
F = \bigcup\limits_{f_p \in Range(R'_{om}) } f_p = F_p\label{eq:th_rt_forward_5}
\end{alignat}
From the antecedent of the forward implication, we know $F_p$ is a well-defined function, therefore, based on \Cref{eq:th_rt_forward_5}, we conclude $F = \bigcup\limits_{f_p \in Range(R'_{om}) } f_p$ is also a well-defined function. 

This conclusion and \Cref{eq:th_rt_forward_4} proves
\begin{multline}
\Gamma(F_p(D_1)) = \Gamma(D_2) \implies \exists R'_{om} \subseteq R_{OM}(Domain(R'_{om}) = Primitive(O_1) \wedge \\
(F = \bigcup\limits_{f_p \in Range(R'_{om}) } f_p ) \text{ is a well-defined function} )\label{eq:th_rt_forward_6}
\end{multline}
The proofs of \Cref{eq:th_rt_forward_6} and \Cref{eq:th_rt_forward_1} complete the proof of the forward implication of this theorem.

Second, we prove the backward implication: 
\begin{multline}
 \exists R'_{om} \subseteq R_{OM}(Domain(R'_{om}) = Primitive(O_1) \wedge (F = \bigcup\limits_{f_p \in Range(R'_{om}) } f_p )\text{ is a well-defined function} ) \\ \wedge |Primitive(O_1)| =|Primitive(O_1)| 
 \implies \Gamma(F_p(D_1),Obj) = \Gamma(D_2,Obj)
\end{multline}
The antecedent $\exists R'_{om} \subseteq R_{OM}( Domain(R'_{om}) = Primitive(O_1))$ implies 
\begin{alignat}{1}
\forall o \in Primitive(O_1), \exists f \in \mathbb{F}, (o,f) \in R'_{om} \label{eq:th_rt_backword_1}
\end{alignat}
From the definition of $R_{OM}$ in \Cref{eq:Rom} and \Cref{eq:th_rt_backword_1}, we have 
\begin{alignat}{1}
\forall o \in Primitive(O_1), \exists f \in Range(R'_{om}), \exists o' \in Primitive(O_2)( \Gamma(f(o)) = \Gamma(o')) \label{eq:th_rt_backword_2}
\end{alignat}
The antecedent defines $F$ as the union of all mappings in $Range(R'_{om})$ ($F = \bigcup\limits_{f_p \in Range(R'_{om}) } f_p$). Hence, we have
\begin{alignat}{1}
\forall f \in Range(R'_{om})(f \subseteq F)
\end{alignat}
Moreover, because $F$ is a well-defined function according to the antecedent of this backward implication, so we can replace $f$ with $F$ in \Cref{eq:th_rt_backword_2}. Then this equation can be written as
\begin{alignat}{1}
\forall o \in Primitive(O_1), \exists o' \in Primitive(O_2)( \Gamma(F(o)) = \Gamma(o')) \label{eq:th_rt_backword_2.5}
\end{alignat}
The domain of every mapping in $Range(R'_{om})$ is the set of predicates of one operator from $Primitive(O_1)$. 
\begin{alignat}{1}
\forall f \in Range(R'_{om}) (Domain(f) = \{ p | p \in Predicates(o) \text{ where } (o,f) \in R'_{om} \}) \label{eq:th_rt_backword_3}
\end{alignat}
Since $F$ is the union of all mappings in $Range(R'_{om})$, the domain of $F$ is equal to the union of the domains of all mappings in $Range(R'_{om})$. 
\begin{alignat}{1}
Domain(F) = \bigcup\limits_{f_p \in Range(R'_{om}) } Domain(f_p) \label{eq:th_rt_backword_4}
\end{alignat}
From \Cref{eq:th_rt_backword_1}, \Cref{eq:th_rt_backword_3}, and \Cref{eq:th_rt_backword_4}, we conclude 
\begin{alignat}{1}
Domain(F) = Predicates(Primitive(O_1))
\end{alignat}
However, since the predicates of the macro and split operators of any domain are subset of the predicates of its primitive operators, we have $Predicates(Primitive(O_1)) = P_1$.
Hence, 
\begin{alignat}{1}
Domain(F) = P_1 \label{eq:th_rt_backword_5_1}
\end{alignat}
The range of every mapping in $Range(R'_{om})$ is the set of predicates of an operator $o'$ from $Primitive(O_2)$ such that the reach set of an operator from $Primitive(O_1)$ under the given mapping is equal to the reach set of the primitive operator $o'$.
The following equation formally defines the range of a mapping from $Range(R'_{om})$. 
\begin{alignat}{1}
\forall f \in Range(R'_{om}) (Range(f) = \{ p | p \in Predicates(o') \label{eq:th_rt_backword_6_1} \\ 
\text{ where } o' \in Primitive(O_2) \text{ if }\ \exists o \in Primitive(O_1) (\Gamma(f(o)) \subseteq \Gamma(o') \}). \nonumber
\end{alignat}
Since $F$ is the union of all mappings in $Range(R'_{om})$ and $F$ is a well-defined function, the range of $F$ is equal to the union of the ranges of all mappings in $Range(R'_{om})$. 
\begin{alignat}{1}
Range(F) = \bigcup\limits_{f_p \in Range(R'_{om}) } Range(f_p)\label{eq:th_rt_backword_7_1}
\end{alignat}
From \Cref{eq:th_rt_backword_6_1} and \Cref{eq:th_rt_backword_7_1}, we conclude
\begin{alignat}{1}
Range(F) \subseteq Predicates(Primitive(O_2))
\end{alignat}
However, since the predicates of the macro and split operators of any domain are subset of the predicates of its primitive operators, we have $Predicates(Primitive(O_2)) = P_2$.
Hence, 
\begin{alignat}{1}
Range(F) \subseteq P_2 \label{eq:th_rt_backword_8_1}
\end{alignat}
The function $F$ is a bijective function because it is the union of a set of bijective functions. 
Hence, $|Domain(F)| = |Range(F)|$. From \Cref{eq:th_rt_backword_5_1}, we have $|Domain(F)| = |P_1|$. So, $|Range(F)| = |P_1|$. 
We know $|P_1| = |P_2|$ from the assumption of this theorem, therefore $|Range(F)| = |P_2|$. Thus, according to \Cref{eq:th_rt_backword_8_1} we infer

\begin{alignat}{1}
 Range(F) = P_2 \label{eq:th_rt_backword_10_1}
\end{alignat}

We have $F$ is a bijective function from $P_2$ to $P_1$ as per \Cref{eq:th_rt_backword_5_1} and \Cref{eq:th_rt_backword_10_1}.
Furthermore, we know from the antecedent of the backward implication $|Primitive(O_1)| =|Primitive(O_2)|$.
In addition to that, \Cref{eq:th_rt_backword_2.5} proves the image of the reach set of each primitive operator in $D_1$ under $F$ is equal to the reach set of a primitive operator in $D_2$.
Therefore, with the help of \Cref{reachabilityLemma}, we can deduce $\Gamma(F_p(D_1)) = \Gamma(D_2)$.
This concludes the proof of the backward implication of this theorem. Thus it completes the proof of the theorem.
\end{proof}
\section[Complex Domains Reachability Theorem]{Proof of Theorem~\ref{th:ComplexDomainsReachabilityTheorem} (See page~\pageref{th:ComplexDomainsReachabilityTheorem})}
\label{ProofOfTheoremCDRT}

To prove this theorem, we first prove \Cref{lem:ComplexDomainReachabilityLemma} in \Cref{sec:ComplexDomiansProofOfLemma} and then the main theorem is proven in \Cref{sec:ProofOfTheoremCDRTusingCDRL}.

\subsection[Complex Domains Reachability Lemma]{Proof of the Lemma of \Cref{th:ComplexDomainsReachabilityTheorem}}
\label{sec:ComplexDomiansProofOfLemma}

\CDRL*
\begin{proof}
Let $D_3 = (P_2, O_3)$ be a planning domain model with the same set of predicates as $D_2$, and the set of its operators $O_3$ satisfies the following conditions:

\begin{gather}
O_3 \subseteq O_2 \\ \nonumber
\wedge \\
\forall o \in O_1, \exists o' \in O_3 (\Gamma(F_p(o)) = \Gamma(o') \nonumber
\end{gather}

From the definition of $O_3$, we have $O_3 \subseteq O_2$; thus, any sequence of operators that can be made by the operators of $O_3$ can also be made by the operators of $D_2$.
Since the reach set of a domain is equal to the union of the reach sets of its sequence of operators, we have $\Gamma(D_3) \subseteq \Gamma(D_2)$.

Furthermore, the definition of $O_3$ implies the image under $F_p$ of the reach set of each operator in $D_1$ is equal to the reach set of an operator from $D_3$.
This means the reach set of any sequence of operators made by the operators of $D_1$ is equal to the reach set of a sequence of operators which can be made from the operators of $D_3$.
Thus we have $\Gamma(F_p(D_1)) \subseteq \Gamma(D_3)$. Therefore, $\Gamma(F_p(D_1)) \subseteq \Gamma(D_2)$.
This argument concludes the proof of this lemma.
\end{proof}

\subsection[Complex Domains Reachability Theorem]{Proof of Theorem~\ref{th:ComplexDomainsReachabilityTheorem} using Lemma~\ref{lem:ComplexDomainReachabilityLemma} (See page~\pageref{th:ComplexDomainsReachabilityTheorem})}
\label{sec:ProofOfTheoremCDRTusingCDRL}
\CDRT*
Let $\mathbb{F}$ be the set of all bijective mappings from predicates of equal arity from every operator in $D_1$ to every operator in $D_2$.
\begin{alignat}{1}
\mathbb{F} = \{ f_p | f_p: Predicates(o) \twoheadrightarrowtail Predicates(o') \ &\text{ where } o \in O_1, o' \in O_2 \nonumber \\
\text { and if } f_p(p) = p' \text{ then } Arity(p) = Arity(p') \}\nonumber 
\end{alignat}

Let $R_{OM}$ be a relation between operators from $D_1$ and predicate mappings from $\mathbb{F}$. An operator $o$ from $D_1$ is related to a mapping $f_p$ from $\mathbb{F}$ by $R_{OM}$ means there exists an operator $o'$ from $D_2$ such that the reach set of $f_p(o)$ is equal to the reach set of $o'$.
\begin{equation}
R_{OM} = \{ (o,f_p) \in O_1 \times \mathbb{F} \ | \ \exists o' \in O_2, \ \Gamma(f_p(o),Obj) = \Gamma(o',Obj) \} \label{eq:Rom_c}
\end{equation}

\begin{proof}
The antecedent $\exists R'_{om} \subseteq R_{OM}( Domain(R'_{om}) = O_1)$ implies 
\begin{alignat}{1}
\forall o \in O_1, \exists f \in \mathbb{F}, (o,f) \in R'_{om} \label{eq:th_rt_backword_1_c}
\end{alignat}
From the definition of $R_{OM}$ in \Cref{eq:Rom_c} and \Cref{eq:th_rt_backword_1_c}, we have 
\begin{alignat}{1}
\forall o \in O_1, \exists f \in Range(R'_{om}), \exists o' \in O_2( \Gamma(f(o)) = \Gamma(o')) \label{eq:th_rt_backword_2_c}
\end{alignat}
The antecedent defines $F$ as the union of all mappings in $Range(R'_{om})$ ($F = \bigcup\limits_{f_p \in Range(R'_{om}) } f_p$). Hence, we have
\begin{alignat}{1}
\forall f \in Range(R'_{om})(f \subseteq F)
\end{alignat}
Moreover, because $F$ is a well-defined function according to the antecedent of this backward implication, so we can replace $f$ with $F$ in \Cref{eq:th_rt_backword_2_c}. Then this equation can be written as
\begin{alignat}{1}
\forall o \in O_1, \exists o' \in O_2( \Gamma(F(o)) = \Gamma(o')) \label{eq:th_rt_backword_2.5_c}
\end{alignat}
The domain of every mapping in $Range(R'_{om})$ is the set of predicates of one operator from $O_1$. 
\begin{alignat}{1}
\forall f \in Range(R'_{om}) (Domain(f) = \{ p | p \in Predicates(o) \text{ where } (o,f) \in R'_{om} \}) \label{eq:th_rt_backword_3_c}
\end{alignat}
Since $F$ is the union of all mappings in $Range(R'_{om})$, the domain of $F$ is equal to the union of the domains of all mappings in $Range(R'_{om})$. 
\begin{alignat}{1}
Domain(F) = \bigcup\limits_{f_p \in Range(R'_{om}) } Domain(f_p) \label{eq:th_rt_backword_4_c}
\end{alignat}
From \Cref{eq:th_rt_backword_1_c}, \Cref{eq:th_rt_backword_3_c}, and \Cref{eq:th_rt_backword_4_c}, we conclude 
\begin{alignat}{1}
Domain(F) = Predicates(O_1)
\end{alignat}
However, $Predicates(O_1) = P_1$.
Hence, 
\begin{alignat}{1}
Domain(F) = P_1 \label{eq:th_rt_backword_5}
\end{alignat}

The range of every mapping in $Range(R'_{om})$ is the set of predicates of an operator $o'$ from $O_2$ such that the reach set of an operator from $O_1$ under the given mapping is equal to the reach set of the operator $o'$.
The following equation formally defines the range of a mapping from $Range(R'_{om})$. 
\begin{multline}
\forall f \in Range(R'_{om}) (Range(f) = \{ p | p \in Predicates(o') 
\text{ where } o' \in O_2 \\ \text{ if }\ \exists o \in O_1 (\Gamma(f(o)) \subseteq \Gamma(o') \}) \label{eq:th_rt_backword_6}
\end{multline}
Since $F$ is the union of all mappings in $Range(R'_{om})$ and $F$ is a well-defined function, the range of $F$ is equal to the union of the ranges of all mappings in $Range(R'_{om})$. 
\begin{alignat}{1}
Range(F) = \bigcup\limits_{f_p \in Range(R'_{om}) } Range(f_p)\label{eq:th_rt_backword_7}
\end{alignat}
From \Cref{eq:th_rt_backword_6} and \Cref{eq:th_rt_backword_7}, we conclude
\begin{alignat}{1}
Range(F) \subseteq Predicates(O_2)
\end{alignat}
However, $Predicates(O_2) = P_2$.
Hence, 
\begin{alignat}{1}
Range(F) \subseteq P_2 \label{eq:th_rt_backword_8}
\end{alignat}
The function $F$ is a bijective function because it is the union of a set of bijective functions. 
Hence, $|Domain(F)| = |Range(F)|$. From \Cref{eq:th_rt_backword_5}, we have $|Domain(F)| = |P_1|$. So, $|Range(F)| = |P_1|$. 
We know $|P_1| = |P_2|$ from the assumption of this theorem, therefore $|Range(F)| = |P_2|$. Thus, according to \Cref{eq:th_rt_backword_8} we infer

\begin{alignat}{1}
 Range(F) = P_2 \label{eq:th_rt_backword_10}
\end{alignat}

We have $F$ is a bijective function from $P_2$ to $P_1$ as per \Cref{eq:th_rt_backword_5} and \Cref{eq:th_rt_backword_10}.
In addition to that \Cref{eq:th_rt_backword_2.5} proves the image of the reach set of each operator in $D_1$ under $F$ is equal to the reach set of a operator in $D_2$.
Therefore, with the help of \Cref{lem:ComplexDomainReachabilityLemma}, we can deduce $\Gamma(F_p(D_1)) \subseteq \Gamma(D_2)$.
These arguments conclude the proof of this theorem.
\end{proof}

\section{Proof of Theorem~\ref{th:ContainmentTheorem} (See page~\pageref{th:ContainmentTheorem})}
\label{ProofOfTheoremDCT}
\DCT*

\begin{proof}
For brevity, the set {\it Obj} is dropped from the reach set symbols in the proofs as it is common to all reach sets.
Items one and two of this theorem will be proven together as follows.
From the antecedent of the forward implication, we have
\begin{alignat}{1}
& \Gamma(F_p(D_1)) \subseteq \Gamma(D_2) \label{eq:d1ingd2}\\
& \Gamma(G_p(D_2)) \subseteq \Gamma(D_1) \label{eq:d2infd1}
\end{alignat}

$F_p(D_1)$ and $(D_2)$ have the same predicates and the reach set of $F_p(D_1)$ is a subset of the reach set of $D_2$.
So, if we rename the predicates of $F_p(D_1)$ and $D_2$ using $F_p^{-1}$, then both $F_p^{-1}(F_p(D_1))$ and $F_p^{-1}(D_2)$ will have the same predicates and the reach set of $F_p^{-1}(F_p(D_1))$ will be a subset of the reach set of $F_p^{-1}(D_2)$. Let us also rename the predicates of $G_p(D_2)$ and $D_1$ using $G_p^{-1}$. Then we have
\begin{alignat}{1}
& \Gamma(D_1) \subseteq \Gamma(F_P^{-1}(D_2)) \label{eq:d1subFpd2}\\
& \Gamma(D_2) \subseteq \Gamma(G_P^{-1}(D_1)) 
\end{alignat}

Since $\Gamma(G_p(D_2)) \subseteq \Gamma(D_1)$ and from (\ref{eq:d1subFpd2}) we have 

\begin{alignat}{1}
& |\Gamma(D_1)| \geq |\Gamma(G_p(D_2))| \label{eq:|d1|leq|gd2|} \\
& |\Gamma(D_1)| \leq |\Gamma(F_p^{-1}(D_2))| \label{eq:|d1|geq|f-1d2|} 
\end{alignat}
Since $f$ and $g$ are bijective functions, they do not change the size of the reach sets on which they are applied. Thus we have 
\begin{alignat}{1}
& |\Gamma(D_2)| = |\Gamma(G_p(D_2))| \\
& |\Gamma(D_2)|= |\Gamma(F_p^{-1}(D_2))|
\end{alignat}
Therefore,
\begin{alignat}{1}
|\Gamma(F_p(D_2))| = |\Gamma(g^{-1}(D_2))| \label{eq:|gd2|=|f-1d2|}
\end{alignat}

From (\ref{eq:|d1|leq|gd2|}), (\ref{eq:|d1|geq|f-1d2|}) and (\ref{eq:|gd2|=|f-1d2|}) we have $|\Gamma(D_1)| = \Gamma(G_p(D_2))| = |\Gamma(F_p^{-1}(D_2))|$.

$\Gamma(G_p(D_2))$ is a subset of $\Gamma(D_1)$ as per (\ref{eq:d2infd1}) and both have equal cardinality, hence:

\begin{alignat}{1}
\Gamma(G_p(D_2)) = \Gamma(D_1) \label{d1eqgd2}
\end{alignat}

$\Gamma(D_1)$ is a subset of $\Gamma(F_p^{-1}(D_2))$ as per (\ref{eq:d1subFpd2}) and both have equal cardinality, hence:

\begin{alignat}{1}
\Gamma(D_1) = \Gamma(F_p^{-1}(D_2)) \label{d1eqf-1d2}
\end{alignat}

Rename the predicates of $D_1$ and $F_p^{-1}(D_2)$ using $F_p$, then we have

\begin{alignat}{1}
\Gamma(F_p(D_1)) = \Gamma(D_2) \label{f_pd1eqd2}
\end{alignat}

From (\ref{d1eqgd2}) and (\ref{d1eqf-1d2}), we infer that $\Gamma(G_p(D_2)) = \Gamma(F_p^{-1}(D_2)) = $.
Hence, $G_p = F_p^{-1}$.
Furthermore, $G_p^{-1} = F_p^{-1^{-1}}$.
Thus, $F_p = G_p^{-1}$
These arguments proves the forward implication.
The proof of the backward implication is trivial because if two sets are equal then the subset relation follows directly.
\end{proof}

\section[Domain Reach Sets Equality Corollary]{Proof of Corollary~\ref{cor:DomainReachSetsEqualityCorollary} (See page~\pageref{cor:DomainReachSetsEqualityCorollary})}
\label{ProofOfCorollaryDRSEC}
\DRSEC*
\begin{proof}
From the antecedent of the forward implication and according to \Cref{lem:ComplexDomainReachabilityLemma}, we have
\begin{equation}
\Gamma(F_p(D_1)) \subseteq \Gamma(D_2) \label{eq:D1SubsetD2}
\end{equation}
As we assume the two domains \D1 and \D2 do not have duplication operators and according to \Cref{th:operatorsStructurReachSet}, we deduce 
\begin{equation}
\forall o_1, o_2 \in O_1\ (\Gamma(o_1) \not = \Gamma(o_2))\label{eq:NoDuplicate}
\end{equation}
From \Cref{eq:D1SubsetD2}, \Cref{eq:NoDuplicate} and because $F_p$ is a bijective function, we infer
\begin{equation}
\forall o' \in O_2, \exists o \in O_1 (\Gamma(F_p^{-1}(o')) = \Gamma(o)) \implies \Gamma(F_p^{-1}(D_2)) \subseteq \Gamma(D_1) \label{eq:D2SubsetD1}
\end{equation}
From \Cref{eq:D1SubsetD2} and \Cref{eq:D2SubsetD1}, and according to \Cref{th:ContainmentTheorem}, we conclude
\begin{equation}
\Gamma(F_p(D_1) = \Gamma(D_2)
\end{equation} 
Hence, we complete the prove of the this corollary. 
\end{proof}

\section[Operators Structure Reach Set Theorem]{Proof of Theorem~\ref{th:operatorsStructurReachSet} (See page~\pageref{th:operatorsStructurReachSet})}
\label{ProofOfTheoremOSRST}
\OSRST*

\subsection{ Definitions of some important relations between atom mappings}
In the proof of this theorem, we will construct a mapping between the atoms of two operators from the mappings between their preconditions, delete effects and add effects.
The mappings relations defined in this section facilitate describing the consistency conditions for unifying the mappings of the preconditions, delete effects and add effects of the two operators. 

The first mapping relation states that two atom mappings have predicate-consistent relation, or they are predicate-consistent mappings if they map the set of atoms that share a common predicate $p$ in one operator to the set of atoms that share some predicate $p'$ in the second operator. 

For example, Let $t_1$ and $t_2$ be two atoms in the preconditions and delete effects of an operator $m$ respectively, and let $t_1$ and $t_2$ to have the same predicate $p$. 
Let $f_{t-pre}$ be a mapping that maps the preconditions of an operator $m$ to the preconditions of an operator $o$ and $f_{t-del}$ a mapping which maps the delete effect of $m$ to the delete effects of $o$.
We say $f_{t-pre}$ and $f_{t-del}$ are predicate-consistent mappings if $f_{t-pre}$ maps $t_1$ to the atom $t'_1$ in the preconditions of $o$ and $f_{t-del}$ maps $t_2$ to the atom $t'_2$ in the delete effects of $o$ such that $t'_1$ and $t'_2$ share a common predicate $p'$.
\begin{defn}\label{Def:predicate-consistent-mappings}
Two bijective atom mappings $f$ and $g$ are \textbf{\emph{predicate-consistent mappings}}, $f \approx_{pred} g$, \textbf{iff} they map atoms with a shared predicate $p$ in their domains to atoms with some predicate $p'$ in their ranges.
$$ f \approx_{pred} g \iff \forall t_1 \in Domain(f), \forall t_2 \in Domain(g) \ \ (Pred(t_1) = Pred(t_2) \rightarrow Pred(f(t_1)) = Pred(g(t_2)))$$
\end{defn}

Another useful relation is the variable-order-consistent relation. This relation requests two different mappings to map atoms with a shared variable $v$ in one operator to atoms that also share some variable $v'$ in the other operator.
Furthermore, this relation has an additional condition that necessitates the positions of the variable $v$ in the parameters of the atoms of the first operator to be equal to the positions of the variable $v'$ in the parameters of the atoms of the second operator.
The function of this relation is to guarantee that when two different mappings are unified, the resulting mapping is consistent with regard to the shared variables and their orders.

For example, Let $t_1$ and $t_2$ be two atoms in the preconditions and delete effects of an operator $m$ respectively, and let $v$ be the second variable in the parameters of $t_1$ and the first variable in the parameters of $t_2$. 
We say $f_{t-pre}$ and $f_{t-del}$ are variable-order-consistent mappings if $f_{t-pre}$ maps $t_1$ to the atom $t'_1$ in the preconditions of $o$ and $f_{t-del}$ maps $t_2$ to the atom $t'_2$ in the delete effects of $o$ such that $v'$ is the second variable in the parameters of $t'_1$ and $v'$ is also the first variable in the parameters of $t'_2$. 

\begin{defn}\label{Def:variable-order-consistent-mappings}

Two bijective atom mappings $f$ and $g$ are \textbf{\emph{variable-order-consistent mappings}}, $f \approx_{var} g$, \textbf{iff} they map atoms with a shared variable in their domains to atoms with some shared variable in their ranges such that the positions of the variable in the parameters of the atoms in the domains of the mappings are equal to the positions of the variable in the parameters of the atoms in the ranges of the mappings.
$$ f \approx_{var} g \iff \forall t_1 \in Domain(f), \forall t_2 \in Domain(g) \ ( $$
$$\forall v_1 \in Var(t_1), \forall v_2 \in Var(t_2) \ ( v_1 = v_2 \rightarrow $$
$$\exists v'_1 \in Var(f(t_1)), \exists v'_2 \in Var(g(t_2)) \ ( v'_1 = v'_2$$
$$ \wedge Position(v_1,t_1) = Position(v'_1,f(t_1)) \wedge Position(v_2,t_2) = Position(v'_2,g(t_2)) \ ))) \ $$
\end{defn}

To simplify the process of referring to the two previous atom mapping relations, we define the atom-consistent relation which encompasses them both.

\begin{defn}\label{Def:atom-consistent-mappings}
Two bijective atom mappings $f$ and $g$ are \textbf{\emph{atom-consistent mappings}} \textbf{iff} they are predicate and variable-order consistent mappings. 
\end{defn}

\subsubsection{Properties of atom mapping relations}
Predicate-consistent, variable-order-consistent and atom-consistent mappings symmetric and transitive relations. Therefore proving the following formula:

$$\exists f_{t-pre}, \exists f_{t-add}, \exists f_{t-del} ( (f_{t-pre} \approx_{pred} f_{t-del}) \wedge (f_{t-del} \approx_{pred} f_{t-add}) \ )$$

Implies: 

$$\exists f_{t-pre}, \exists f_{t-add}, \exists f_{t-del} ( (f_{t-pre} \approx_{pred} f_{t-add}) \wedge (f_{t-add} \approx_{pred} f_{t-pre})$$
$$\wedge$$
$$(f_{t-del} \approx_{pred} f_{t-pre}) \wedge (f_{t-add} \approx_{pred} f_{t-del}) \ ) $$
\subsection{Theorem}
\OSRST*
\begin{proof}
To prove the bidirectional implication in this theorem, we prove its forward and backward implications.
We will provide the proofs of the forward and backward implications with regard to $f_t$. In the same way, we can prove the forward and backward implications with respect to $f_t^{-1}$, but this proof is not provided to avoid repetition.
In this proof, we refer to the second condition of this theorem as the ``arity condition'', the third condition as the ``predicate condition'', and the last condition as the ``variable-order condition''.
\subsection{The proof of the forward implication}

\subsubsection{Proof sketch}
To prove the forward implication, we will show that if $m$ and $o$ have equal reach sets under a bijective mapping $f_p$ between the predicates of $o$ with those of $m$ with equal arity, then there will exist an atom mapping $f_t$ that respects the conditions of this theorem.

Starting from the antecedent of the forward implication, we will prove the existence of three bijective mappings, $f_{t-pre}$,$f_{t-del}$ and $f_{t-add}$, from the atoms of the preconditions, delete effects and add effects of $m$ to the atoms of the preconditions, delete effects and add effects of $o$ respectively, such that these mappings respect the arity, predicate, and variable-order conditions of this theorem.
Then, we will prove that these three mappings are atom-consistent mappings with respect to each other.
Finally, we will show because $f_{t-pre}$,$f_{t-del}$ and $f_{t-add}$ respect the arity, predicate, and variable-order conditions and they are atom-consistent mappings, they can be used to define an atom mapping $f'_t$ that satisfies the conditions of $f_t$. Hence proving the existence of $f_t$ starting from the antecedent of the forward implication.
\subsubsection{The proof of the existence of preconditions, delete effects and add effects mappings that satisfy the arity, predicate, and variable-order conditions}
The antecedent of the forward implication, $ \Gamma(f_p(m)) = \Gamma(o)$, can be rewritten as follows. 
\begin{alignat}{1}
\forall (s_i,s_j) \in \Gamma(o) \rightarrow (s_i,s_j) \in \Gamma(f_p(m)) \ \wedge \forall (s_i,s_j) \in \Gamma(f_p(m)) \rightarrow (s_i,s_j) \in \Gamma(o) \label{eq:equal-reach-set}
\end{alignat}
$(s_i,s_j) \in \Gamma(o)$ means there is an action instantiated from $o$ which is applicable in $s_i$ and can reach $s_j$.
Similarly, $(s_i,s_j) \in \Gamma(f_p(m))$ means there exists an action instantiated from $f_p(m)$ which is also applicable in $s_i$ and can reach $s_j$.
So, both $f_p(m)$ and $o$ can produce actions applicable in the same states. 
For two operators to be able to instantiate actions applicable in the same states, the two operators must have the same preconditions up, but not necessarily, to the variables' names.

This implies that for every atom in the preconditions of $f_p(m)$, there is an atom in the preconditions of $o$ such that the two atoms have the same predicate and the same order of variables with respect to the other atoms in their perspective preconditions.

This means the number of atoms of each arity and the number of atoms of each predicate are equal in the preconditions of the two operators. Furthermore, for every two atoms in the preconditions of $f_p(m)$ that have a shared variable $v$, there are two atoms in the preconditions of $o$ that share some variable $v'$ and the positions of $v$ in the parameters of its atoms are equal to the positions of $v'$ in the parameters of its atoms.

The operator $f_p(m)$ is produced by renaming the predicates of $m$ using the predicate mapping $f_p$ which does not change the number of predicates, atoms or the variables of $m$. So, $f_p(m)$ and $m$ differ by only the name of the predicates.
Therefore, the number of atoms of each arity is equal in the preconditions of $f_p(m)$ and $m$.
Furthermore, the number of atoms that share some predicate is also equal in the preconditions of $f_p(m)$ and $m$. 
Moreover, $f_p(m)$ and $m$ have the same variables.

The conclusion is that the number of atoms of each arity and the number of atoms of each predicate are equal in the preconditions of $m$ and $o$. Furthermore, for every two atoms in $o$ that have a shared variable $v$, there are two atoms in the preconditions of $m$ that share some variable $v'$ and the positions of $v$ in the parameters of its atoms are equal to the positions of $v'$ in the parameters of its atoms.
This proves the existence of a bijective mapping from the atoms of the preconditions of $m$ to the atoms of the preconditions of $o$ that satisfies the following conditions:
\begin{enumerate}
 \item Atoms from the preconditions of $m$ are mapped to atoms in the preconditions of $o$ of equal arity;
 \item Atoms in the preconditions of $m$ that share a predicate $p$ must be mapped to atoms in the preconditions of $o$ that share some predicate $p'$; and
 \item Atoms in the preconditions of $m$ that share a variable $v$ must be mapped to atoms in the preconditions of $o$ that share some variable $v'$ and the position of $v$ in the parameters of its atoms is equal to the position of $v'$ in the parameters of its atoms.
\end{enumerate}
Formally,
$$\exists f_{t-pre}: Pre(m) \twoheadrightarrowtail Pre(o)) \ ( $$

$$ \forall t \in Pre(m) \ (\exists t' \in Pre(o) : f_{t-pre}(t) = t' \wedge Arity(t) = Arity(t') ) $$
$$\wedge$$
$$ \forall t_1,t_2 \in Pre(m) \ ( Pred(t_1) = Pred(t_2) \rightarrow \exists t'_1,t'_2 \in Pre(o) \ ( Pred(t'_1) = Pred(t'_2)$$
$$ \wedge (f_{t-pre}(t_1) = t'_1 \wedge f_{t-pre}(t_2) = t'_2 ) \vee (f_{t-pre}(t_2) = t'_1 \wedge f_{t-pre}(t_1) = t'_2 ) ) ) $$
$$\wedge$$
$$ \forall t_1,t_2 \in Atoms(m), \forall v_1 \in Var(t_1), \forall v_2 \in Var(t_2) \ ( v_1 = v_2 \rightarrow $$
$$\exists t'_1,t'_2 \in Atoms(o), \exists v'_1 \in Var(t'_1), \exists v'_2 \in Var(t'_2) \ ( v'_1 = v'_2$$
$$ \wedge ( (f_{t-pre}(t_1) = t'_1 \wedge f_{t-pre}(t_2) = t'_2 \wedge Position(v_1,t_1) = Position(v'_1,t'_1) \wedge Position(v_2,t_2) = Position(v'_2,t'_2))$$
$$ \vee (f_{t-pre}(t_2) = t'_1 \wedge f_{t-pre}(t_1) = t'_2 \wedge Position(v_2,t_2) = Position(v'_1,t'_1) \wedge Position(v_1,t_1) = Position(v'_2,t'_2)) ) ) ) ) $$

We have concluded from the proposition \eqref{eq:equal-reach-set} that both $f_p(m)$ and $o$ can produce actions that are applicable in the same states. Furthermore, from this proposition, we can also conclude that $f_p(m)$ and $o$ can produce actions that can reach the same end states from equal states.
For two operators to be able to instantiate actions that can reach the same states from equal states, the two operators must have the same preconditions, add effects and delete effects up, but not necessarily, to the variables name.
With a similar discussion as the one used to prove the existence of $f_{t-per}$, we can prove the existence of: 
\begin{enumerate}
 \item A bijective mapping between the atoms of the add effects of $m$ and $o$, $f_{t-add}: Add(m) \twoheadrightarrowtail Add(o)) $, and
 \item A bijective mapping between the atoms of the delete effects of $m$ and $o$ ,$f_{t-del}: Del(m) \twoheadrightarrowtail Del(o)) $.

\end{enumerate}
Such that these mappings satisfy the following conditions:
\begin{enumerate}
 \item Atoms in $add(m)$ ($del(m)$) are mapped to atoms in $add(o)$ ($del(o)$) of equal arity;
 \item Atoms in $add(m)$ ($del(m)$) that share a predicate $p$ must be mapped to atoms in $add(o)$ ($del(o)$) that share some predicate $p'$; and
 \item Atoms in $add(m)$ ($del(m)$) that share a variable $v$ must be mapped to atoms in $add(o)$ ($del(o)$) that share some variable $v'$ and the positions of $v$ in the parameters of its atoms is equal to the positions of $v'$ in the parameters of its atoms.
\end{enumerate}

\subsubsection{Proving the preconditions, delete effects and add effects mappings are atom-consistent mappings}

Let's call the set of mappings that satisfy the conditions of $f_{t-pre}$ as $Sf_{t-pre}$. This set is not empty as we have proved.
Similarly, we define the non-empty sets of mappings $Sf_{t-add}$ and $Sf_{t-del}$ which satisfy the conditions of $f_{t-add}$ and $f_{t-del}$ respectively.
Now, we have to prove that $f_{t-per}$, $f_{t-add}$ and $f_{t-del}$ are atom-consistent mappings with respect to each other where $f_{t-pre} \in Sf_{t-pre}$, $f_{t-add} \in Sf_{t-add}$, and $f_{t-del} \in Sf_{t-del}$.
Formally, we have to prove this formula:
$$\exists f_{t-pre} \in Sf_{t-pre}, \exists f_{t-add} \in Sf_{t-add}, \exists f_{t-del} \in Sf_{t-del} \ ($$

$$ (f_{t-pre} \approx_{pred} f_{t-del}) \wedge (f_{t-del} \approx_{pred} f_{t-add}) $$
$$\wedge$$
$$ (f_{t-pre} \approx_{var} f_{t-del}) \wedge (f_{t-del} \approx_{var} f_{t-add}) $$

Note, since $\approx_{pred}$ and $\approx_{var}$ are symmetric and transitive relations, we do not need to have all possible combinations of $f_{t-pre}$, $f_{t-del}$, and $f_{t-add}$ with $\approx_{pred}$ and $\approx_{var}$ to express $f_{t-pre}$, $f_{t-del}$, and $f_{t-add}$ are atom-consistent mappings with respect to each other.

We will prove that if $m$ and $o$ have equal reach sets under a bijective mapping between the predicates of $o$ and those of $m$ with equal arity, then $f_{t-per}$, $f_{t-add}$ and $f_{t-del}$ are atom-consistent mappings. For this purpose, we will prove the contrapositive form of this implication.
If $f_{t-per}$, $f_{t-add}$ and $f_{t-del}$ are not atom-consistent mappings then $m$ and $o$ do not have equal reach sets under any bijective between the predicates of $o$ and those of $m$ with equal arity.
The contrapositive form is formally defined as follows:
$$ \forall f_p: Predicates(m) \twoheadrightarrowtail Predicates(o) \ \text{ where if } \ f_p(p) = p' \text{ then } Arity(p) = Arity(p') ($$
$$\forall f_{t-pre} \in Sf_{t-pre}, \forall f_{t-add} \in Sf_{t-add}, \forall f_{t-del} \in Sf_{t-del} \ ($$
$$ (f_{t-pre} \not\approx_{pred} f_{t-del}) \vee (f_{t-del}\not\approx_{pred} f_{t-add}) $$
$$\vee$$
$$ (f_{t-pre}\not\approx_{var} f_{t-del}) \vee (f_{t-del}\not\approx_{var} f_{t-add}) \ )$$
$$ \rightarrow $$
$$\ \Gamma(f_p(m)) \not = \Gamma(o) \ ) $$

To prove this implication, we have to prove that every element in the antecedent implies the consequent.
First, we prove $\forall f_{t-pre} \in Sf_{t-pre}, \forall f_{t-del} \in Sf_{t-del} \
(f_{t-pre} \not\approx_{pred} f_{t-del} \rightarrow \Gamma(f_p(m)) \not = \Gamma(o))$.
If there are no precondition mappings and delete effect mappings from the atoms of $m$ to the atoms of $o$ such that these mappings are predicate-consistent, then there will be two atoms that share the same predicate $p$, one in the preconditions and the other in the delete effects of $m$, that cannot be mapped to atoms that share some predicate $p'$ in the preconditions and delete effects of $o$ respectively.
This means that under any bijective predicate mapping $f_p$, $f_p(m)$ will be different from $o$ by either a precondition or a delete effect.
If $f_p(m)$ differ from $o$ by a precondition, then the actions produced from $f_p(m)$ and $o$ will be applicable in different set of states.
On the other hand, if $f_p(m)$ differs from $o$ by a delete effect, then the actions produced from $f_p(m)$ and $o$ will not reach the same states. 
Therefore, $f_p(m)$ and $o$ will have different reach sets under any bijective mapping between the predicates of $o$ and those of $m$ with equal arity.
This concludes the proof of this sub-formula.
Similarly, we prove the correctness of the following formulas:

$$\forall f_{t-del} \in Sf_{t-del}, \forall f_{t-add} \in Sf_{t-add} \
(f_{t-del} \not\approx_{pred} f_{t-add} \rightarrow \Gamma(f_p(m)) \not = \Gamma(o))$$

Now we have to prove $\forall f_{t-pre} \in Sf_{t-pre}, \forall f_{t-del} \in Sf_{t-del} \
(f_{t-pre} \not\approx_{var} f_{t-del} \rightarrow \Gamma(f_p(m)) \not = \Gamma(o).$
There are two possible reasons for preconditions mapping and delete effects mapping to be not variable-order-consistent mappings.
Either the cardinality of the sets of atoms that share some variable $v$ in the preconditions and delete effects of $m$ is greater than the cardinality of the sets of atoms that share some variable $v'$ in the preconditions and delete effects of $o$.
In this case, there will be two atoms $t_1$ and $t_2$ that share the same variable $v$, one in the preconditions and the other in the delete effects of $m$, but $o$ does not have two atoms that share some variable $v'$ in the preconditions and delete effects of $o$ and can be mapped to the atoms $t_1$ and $t_2$.
Consequently, under any bijective predicate mapping $f_p$, $f_p(m)$ and $o$ will produce different actions for any set of objects.
Thus, the actions produced from $f_p(m)$ will have a precondition and a delete effect that have the same object as a parameter, whereas the same precondition and delete effect in the actions produced from $o$ will not have the same object as well.
Thus $f_p(m)$ differs from $o$ by either a precondition or a delete effect.

The other reason for not having precondition and delete effect mappings that are variable-order-consistent is as follows.
If the number of atoms with a shared variable $v$ in each of the preconditions and delete effects of $m$ is equal to the number of atoms with a shared variable $v'$ in each of the preconditions and delete effects of $o$ respectively, then the order of the variables $v$ and $v'$ in the atoms of $m$ and $o$ must be not the same.
Consequently, under any bijective predicate mapping $f_p$, $f_p(m)$ and $o$ will produce different actions for any set of objects.
In the actions produced from $f_p(m)$, there will be a precondition and a delete effect that have the same object as a parameter in specific positions, whereas the same precondition and delete effect in the actions produced from $o$ will have the same object as well but not in the same positions as the actions produced from $f_p(m)$.
Thus $f_p(m)$ differs from $o$ by either a precondition or a delete effect.

In either case, if $f_p(m)$ differ from $o$ by a precondition, then the actions produced from $f_p(m)$ and $o$ will be applicable in different set of states.
On the other hand, if $f_p(m)$ differs from $o$ by a delete effect, then the actions produced from $f_p(m)$ and $o$ will not reach the same states. 
Therefore, $f_p(m)$ and $o$ will have different reach sets under any bijective predicate mapping.
This concludes the proof of this sub-formula.
Similarly, we prove the correctness of the following formulas:

$$\forall f_{t-del} \in Sf_{t-del}, \forall f_{t-add} \in Sf_{t-add} \
(f_{t-del} \not\approx_{var} f_{t-add} \rightarrow \Gamma(f_p(m)) \not = \Gamma(o)). $$
\subsubsection{Proving the unification of the preconditions, delete effects and add effects mappings is a mapping that satisfy the conditions of $f_t$}
So far, we have proven that three atom-consistent bijective mappings exist from the preconditions, add effects and delete effects of the operator $m$ to the preconditions, add effects and delete effects of the operator the $o$ respectively. We also proved that these mappings satisfy the arity, predicate, and variable-order conditions.
Now, we will show that the unification of these mappings produces an atom mapping $f'_t$ that satisfies the conditions of $f_t$.
$$
 \text{Let } f'_t(t) =
 \begin{cases}
 f_{t-pre}(t) \ \text{ if } t \in Pre(m) \\
 f_{t-del}(t) \ \text{ if } t \in Del(m) \setminus Pre(m) \\
 f_{t-add}(t) \ \text{ if } t \in Add(m) \\
 
 \end{cases}
 $$
We know $f_t$ is defined over $Atoms(m)$ which is equal to $Pre(m) \cup (Del(m) \setminus Pre(m)) \cup Add(m)$.
The range of $f_t$ is $Atoms(o)$ which is equal to $Pre(o) \cup Del(o) \cup Add(o)$.
Therefore, $f'_t$ and $f_t$ have equal domains and ranges.
Moreover, since $f'_t$ consists of the bijective atom-consistent mappings $f_{t-pre}$, $f_{t-add}$, and $f_{t-del}$ which satisfy the arity, predicate, and variable-order conditions, the relation $f'_t$ is a bijective mapping and has the following properties. 
\begin{enumerate}
 \item Atoms in $pre(m)$, $del(m)$, and $add(m)$ are mapped to atoms in $pre(o)$, $del(o)$, and $add(o)$ respectively;

 \item Atoms in $m$ are mapped to atoms in $o$ with equal arity;
 
 \item Atoms in $m$ with the same predicate $p$ are mapped to atoms with some predicate $p'$ in $o$; and

 \item Atoms in $m$ with a shared variable $v$ are mapped to atoms in $o$ with some shared variable $v'$ such that the positions of $v$ and $v'$ in the mapped atoms are equal.
\end{enumerate}

Thus $f'_t$ satisfies the conditions of $f_t$ as stated in this theorem.
The existence of $f'_t$ is a prove of the existence of $f_t$.
Therefore the proof of the existence of $f'_t$ concludes the proof of the forward implication. 
\subsection{The proof of the backward implication using a constructive approach}

The backward implication states that the existence of a bijective mapping $f_t$ from the atoms of $m$ to the atoms $o$, which respects the conditions in this theorem, implies there exists a bijective mapping $f_p$ between the predicates of $m$ and those of $o$ of equal arity, and that the reach set of $m$ under $f_p$ is equal to the reach set of $o$.

\subsubsection{Proof sketch}
The backward implication is proven constructively starting from the existence of a bijective mapping $f_t$ that satisfies the conditions stated in this theorem. From this antecedent, we will prove the existence of a bijective mapping $f_p$ between the predicates of $m$ and those of $o$ of equal arity. Then, we will demonstrate that $\Gamma(f_p(m)) = \Gamma(o)$.

\subsubsection{Proving the existence of a bijective mapping $f_p$ between the predicates of $m$ and those of $o$ of equal arity} 
The function $f_t$ maps atoms with the same predicate $p$ in $m$ to atoms with some predicate $p'$ in $o$. This implies that both operators $m$ and $o$ have an equal number of predicates. Furthermore, $f_t$ maps atoms in $m$ to atoms in $o$ with equal arity. Therefore, any atom in $m$ with a predicate $p$ must be mapped by $f_t$ to an atom in $o$ with a predicate $p'$ such that $p$ and $p'$ have equal arity. Thus, there excites a bijective mapping $f_p$ between the predicates of $m$ and those of $o$ of equal arity. 

\subsubsection{Proving $\Gamma(f_p(m)) = \Gamma(o)$}
We will prove that $f_p(m)$ and $o$ are identical apart from the variable names. This means we will prove that one operator can be produced from the other by renaming its variables. This guarantees that $f_p(m)$ and $o$ have the same reach set.

First, the existence of the mapping $f_t$ between the atoms of $m$ and $o$ implies the number of atoms of each arity and the number of atoms of each predicate are equal in the preconditions of $m$ and $o$. Furthermore, for every two atoms in $o$ that have a shared variable $v$, there are two atoms in the preconditions of $m$ that share some variable $v'$ and the positions of $v$ in the parameters of its atoms are equal to the positions of $v'$ in the parameters of its atoms. This conclusion is supported by the properties of the mapping $f_t$.

Second, the operator $f_p(m)$ is produced by renaming the predicates of $m$ using the predicate mapping $f_p$ which does not change the number of predicates, atoms or the variables of $m$. So, $f_p(m)$ and $m$ differ by only the name of the predicates.
Therefore, the number of atoms of each arity is equal in the preconditions of $f_p(m)$ and $m$.
Furthermore, the number of atoms that share some predicate is also equal in the preconditions of $f_p(m)$ and $m$. 
Moreover, $f_p(m)$ and $m$ have the same variables.

From the previous two points, we can conclude that the number of atoms of each arity and the number of atoms of each predicate are equal in the preconditions of $f_p(m)$ and $o$. Furthermore, for every two atoms in the preconditions of $f_p(m)$ that have a shared variable $v$, there are two atoms in the preconditions of $o$ that share some variable $v'$ and the positions of $v$ in the parameters of its atoms are equal to the positions of $v'$ in the parameters of its atoms.
Therefore, $f_p(m)$ and $o$ must have the same preconditions apart from the variable names.

With a similar discussion to the one used to prove both $f_p(m)$ and $o$ have the same preconditions, we can prove the two operators $f_p(m)$ and $o$ have the same delete and add effects apart from the variable names as well.

This concludes that both $f_p(m)$ and $o$ are identical apart from the variable names. In other words, we can say that one operator is produced from the other by renaming its variables. As such, the two operators have equal reach sets.
This proves the backward implication of this theorem.
Hence, we conclude the proof of this theorem. 
\end{proof}
\section{Description of the modifications applied to the planning domain models in the experiments} \label{sec:ExperimentsModifications} %
\GripperModifications
\ElevatorModifications
\ParkingModifications
\HikingModifications
\FloortileModifications
\ChildsnackModifications
\LogisticsModifications
\CavedivingModifications
\RoverModifications
\PipesworldModifications
\ScanalyzerModifications
\FreecellModifications

\vskip 0.2in
\bibliography{Functional-equivalence-jair21}

\begin{thebibliography}{}

\bibitem[\protect\BCAY{Aineto, Jim{\'e}nez,\ \BBA\ Onaindia}{Aineto
  et~al.}{2018}]{aineto2018learning}
Aineto, D., Jim{\'e}nez, S., \BBA\ Onaindia, E. \BBOP2018\BBCP.
\newblock \BBOQ Learning strips action models with classical planning\BBCQ\
\newblock In {\Bem Twenty-Eighth International Conference on Automated Planning
  and Scheduling}.

\bibitem[\protect\BCAY{Aineto, Jim{\'e}nez, Onaindia,\ \BBA\
  Ram{\'\i}rez}{Aineto et~al.}{2019}]{aineto2019model}
Aineto, D., Jim{\'e}nez, S., Onaindia, E., \BBA\ Ram{\'\i}rez, M.
  \BBOP2019\BBCP.
\newblock \BBOQ Model recognition as planning\BBCQ\
\newblock In {\Bem Proceedings of the International Conference on Automated
  Planning and Scheduling}, \lowercase{\BVOL}~29, \BPGS\ 13--21.

\bibitem[\protect\BCAY{Areces, Bustos, Dom{\'\i}nguez,\ \BBA\ Hoffmann}{Areces
  et~al.}{2014}]{areces2014optimizing}
Areces, C., Bustos, F., Dom{\'\i}nguez, M.~A., \BBA\ Hoffmann, J.
  \BBOP2014\BBCP.
\newblock \BBOQ Optimizing planning domains by automatic action schema
  splitting.\BBCQ\
\newblock In {\Bem ICAPS}. Citeseer.

\bibitem[\protect\BCAY{Botea, Enzenberger, M{\"u}ller,\ \BBA\ Schaeffer}{Botea
  et~al.}{2005}]{botea2005macro}
Botea, A., Enzenberger, M., M{\"u}ller, M., \BBA\ Schaeffer, J. \BBOP2005\BBCP.
\newblock \BBOQ Macro-ff: Improving ai planning with automatically learned
  macro-operators\BBCQ\
\newblock {\Bem Journal of Artificial Intelligence Research}, {\Bem 24},
  581--621.

\bibitem[\protect\BCAY{Bryce, Benton,\ \BBA\ Boldt}{Bryce
  et~al.}{2016}]{bryce2016maintaining}
Bryce, D., Benton, J., \BBA\ Boldt, M.~W. \BBOP2016\BBCP.
\newblock \BBOQ Maintaining evolving domain models\BBCQ\
\newblock In {\Bem Proceedings of the twenty-fifth international joint
  conference on artificial intelligence}, \BPGS\ 3053--3059.

\bibitem[\protect\BCAY{Chakraborti, Sreedharan, Zhang,\ \BBA\
  Kambhampati}{Chakraborti et~al.}{2017}]{chakraborti2017plan}
Chakraborti, T., Sreedharan, S., Zhang, Y., \BBA\ Kambhampati, S.
  \BBOP2017\BBCP.
\newblock \BBOQ Plan explanations as model reconciliation: Moving beyond
  explanation as soliloquy\BBCQ\
\newblock {\Bem arXiv preprint arXiv:1701.08317}.

\bibitem[\protect\BCAY{Chrpa\ \BBA\ McCluskey}{Chrpa\ \BBA\
  McCluskey}{2012}]{chrpa2012exploiting}
Chrpa, L.\BBACOMMA\  \BBA\ McCluskey, T.~L. \BBOP2012\BBCP.
\newblock \BBOQ On exploiting structures of classical planning problems:
  Generalizing entanglements.\BBCQ\
\newblock In {\Bem ECAI}, \BPGS\ 240--245.

\bibitem[\protect\BCAY{Chrpa, Vallati,\ \BBA\ McCluskey}{Chrpa
  et~al.}{2015}]{chrpa2015online}
Chrpa, L., Vallati, M., \BBA\ McCluskey, T.~L. \BBOP2015\BBCP.
\newblock \BBOQ On the online generation of effective macro-operators\BBCQ\
\newblock In {\Bem Twenty-Fourth International Joint Conference on Artificial
  Intelligence}.

\bibitem[\protect\BCAY{Coles\ \BBA\ Smith}{Coles\ \BBA\
  Smith}{2007}]{coles2007marvin}
Coles, A.~I.\BBACOMMA\  \BBA\ Smith, A.~J. \BBOP2007\BBCP.
\newblock \BBOQ Marvin: A heuristic search planner with online macro-action
  learning\BBCQ\
\newblock {\Bem Journal of Artificial Intelligence Research}, {\Bem 28},
  119--156.

\bibitem[\protect\BCAY{Cresswell, McCluskey,\ \BBA\ West}{Cresswell
  et~al.}{2013}]{cresswell2013acquiring}
Cresswell, S.~N., McCluskey, T.~L., \BBA\ West, M.~M. \BBOP2013\BBCP.
\newblock \BBOQ Acquiring planning domain models using locm\BBCQ\
\newblock {\Bem Knowledge Engineering Review}, {\Bem 28\/}(2), 195--213.

\bibitem[\protect\BCAY{De~Moura\ \BBA\ Bj{\o}rner}{De~Moura\ \BBA\
  Bj{\o}rner}{2008}]{de2008z3}
De~Moura, L.\BBACOMMA\  \BBA\ Bj{\o}rner, N. \BBOP2008\BBCP.
\newblock \BBOQ Z3: An efficient smt solver\BBCQ\
\newblock In {\Bem International conference on Tools and Algorithms for the
  Construction and Analysis of Systems}, \BPGS\ 337--340. Springer.

\bibitem[\protect\BCAY{Felsing, Grebing, Klebanov, R{\"u}mmer,\ \BBA\
  Ulbrich}{Felsing et~al.}{2014}]{felsing2014automating}
Felsing, D., Grebing, S., Klebanov, V., R{\"u}mmer, P., \BBA\ Ulbrich, M.
  \BBOP2014\BBCP.
\newblock \BBOQ Automating regression verification\BBCQ\
\newblock In {\Bem Proceedings of the 29th ACM/IEEE international conference on
  Automated software engineering}, \BPGS\ 349--360.

\bibitem[\protect\BCAY{Godlin\ \BBA\ Strichman}{Godlin\ \BBA\
  Strichman}{2009}]{godlin2009regression}
Godlin, B.\BBACOMMA\  \BBA\ Strichman, O. \BBOP2009\BBCP.
\newblock \BBOQ Regression verification\BBCQ\
\newblock In {\Bem Proceedings of the 46th Annual Design Automation
  Conference}, \BPGS\ 466--471.

\bibitem[\protect\BCAY{Godlin\ \BBA\ Strichman}{Godlin\ \BBA\
  Strichman}{2013}]{godlin2013regression}
Godlin, B.\BBACOMMA\  \BBA\ Strichman, O. \BBOP2013\BBCP.
\newblock \BBOQ Regression verification: proving the equivalence of similar
  programs\BBCQ\
\newblock {\Bem Software Testing, Verification and Reliability}, {\Bem
  23\/}(3), 241--258.

\bibitem[\protect\BCAY{Hoffmann\ \BBA\ Nebel}{Hoffmann\ \BBA\
  Nebel}{2001}]{hoffmann2001ff}
Hoffmann, J.\BBACOMMA\  \BBA\ Nebel, B. \BBOP2001\BBCP.
\newblock \BBOQ The ff planning system: Fast plan generation through heuristic
  search\BBCQ\
\newblock {\Bem Journal of Artificial Intelligence Research}, {\Bem 14},
  253--302.

\bibitem[\protect\BCAY{Howey, Long,\ \BBA\ Fox}{Howey
  et~al.}{2004}]{howey2004val}
Howey, R., Long, D., \BBA\ Fox, M. \BBOP2004\BBCP.
\newblock \BBOQ Val: Automatic plan validation, continuous effects and mixed
  initiative planning using pddl\BBCQ\
\newblock In {\Bem 16th IEEE International Conference on Tools with Artificial
  Intelligence}, \BPGS\ 294--301. IEEE.

\bibitem[\protect\BCAY{Hutter, Hoos, Leyton-Brown,\ \BBA\ St{\"u}tzle}{Hutter
  et~al.}{2009}]{hutter2009paramils}
Hutter, F., Hoos, H.~H., Leyton-Brown, K., \BBA\ St{\"u}tzle, T.
  \BBOP2009\BBCP.
\newblock \BBOQ Paramils: an automatic algorithm configuration framework\BBCQ\
\newblock {\Bem Journal of Artificial Intelligence Research}, {\Bem 36},
  267--306.

\bibitem[\protect\BCAY{IPC}{IPC}{2014}]{IPC2014}
IPC \BBOP2014\BBCP.
\newblock \BBOQ International planning competition 2014 web site\BBCQ\
\newblock \url{https://helios.hud.ac.uk/scommv/IPC-14/domains.html}.
\newblock [Online; accessed 15-July-2019].

\bibitem[\protect\BCAY{Kuehlmann\ \BBA\ van Eijk}{Kuehlmann\ \BBA\ van
  Eijk}{2002}]{kuehlmann2002combinational}
Kuehlmann, A.\BBACOMMA\  \BBA\ van Eijk, C.~A. \BBOP2002\BBCP.
\newblock \BBOQ Combinational and sequential equivalence checking\BBCQ\
\newblock {\Bem Logic synthesis and Verification}, 343--372.

\bibitem[\protect\BCAY{McCluskey, Vaquero,\ \BBA\ Vallati}{McCluskey
  et~al.}{2017}]{mccluskey2017engineering}
McCluskey, T.~L., Vaquero, T.~S., \BBA\ Vallati, M. \BBOP2017\BBCP.
\newblock \BBOQ Engineering knowledge for automated planning: Towards a notion
  of quality\BBCQ\
\newblock In {\Bem Proceedings of the Knowledge Capture Conference}, \BPGS\
  1--8.

\bibitem[\protect\BCAY{McDermott}{McDermott}{2000}]{McDermott_2000}
McDermott, D.~M. \BBOP2000\BBCP.
\newblock \BBOQ The 1998 ai planning systems competition\BBCQ\
\newblock {\Bem AI Magazine}, {\Bem 21\/}(2), 35.

\bibitem[\protect\BCAY{Mishchenko, Chatterjee, Brayton,\ \BBA\ Een}{Mishchenko
  et~al.}{2006}]{mishchenko2006improvements}
Mishchenko, A., Chatterjee, S., Brayton, R., \BBA\ Een, N. \BBOP2006\BBCP.
\newblock \BBOQ Improvements to combinational equivalence checking\BBCQ\
\newblock In {\Bem Proceedings of the 2006 IEEE/ACM international conference on
  Computer-aided design}, \BPGS\ 836--843.

\bibitem[\protect\BCAY{Newton, Levine, Fox,\ \BBA\ Long}{Newton
  et~al.}{2007}]{newton2007learning}
Newton, M. A.~H., Levine, J., Fox, M., \BBA\ Long, D. \BBOP2007\BBCP.
\newblock \BBOQ Learning macro-actions for arbitrary planners and
  domains.\BBCQ\
\newblock In {\Bem ICAPS}, \lowercase{\BVOL}\ 2007, \BPGS\ 256--263.

\bibitem[\protect\BCAY{Riddle, Douglas, Barley,\ \BBA\ Franco}{Riddle
  et~al.}{2016}]{riddle2016improving}
Riddle, P., Douglas, J., Barley, M., \BBA\ Franco, S. \BBOP2016\BBCP.
\newblock \BBOQ Improving performance by reformulating pddl into a bagged
  representation\BBCQ\
\newblock In {\Bem Proceedings of the 8th Workshop on Heuristic Search for
  Domain-independent Planning (HSDIP@ ICAPS)}, \BPGS\ 28--36.

\bibitem[\protect\BCAY{Shoeeb\ \BBA\ McCluskey}{Shoeeb\ \BBA\
  McCluskey}{2011}]{shoeeb2011comparing}
Shoeeb, S.\BBACOMMA\  \BBA\ McCluskey, T. \BBOP2011\BBCP.
\newblock \BBOQ On comparing planning domain models\BBCQ\
\newblock In {\Bem Proceedings of the Annual PLANSIG Conference}, \BPGS\
  92--94. UK PLANNING AND SCHEDULING Special Interest Group.

\bibitem[\protect\BCAY{Sreedharan, Hernandez, Mishra,\ \BBA\
  Kambhampati}{Sreedharan et~al.}{2019}]{ijcai2019p83}
Sreedharan, S., Hernandez, A.~O., Mishra, A.~P., \BBA\ Kambhampati, S.
  \BBOP2019\BBCP.
\newblock \BBOQ Model-free model reconciliation\BBCQ\
\newblock In {\Bem Proceedings of the Twenty-Eighth International Joint
  Conference on Artificial Intelligence, {IJCAI-19}}, \BPGS\ 587--594.
  International Joint Conferences on Artificial Intelligence Organization.

\bibitem[\protect\BCAY{Vallati, Chrpa,\ \BBA\ Cerutti}{Vallati
  et~al.}{2013}]{Vallati2013TowardsAP}
Vallati, M., Chrpa, L., \BBA\ Cerutti, F. \BBOP2013\BBCP.
\newblock \BBOQ Towards automated planning domain models generation\BBCQ\
\newblock In {\Bem The 5th Italian Workshop on Planning and Scheduling, 4th
  December 2013, Turin, Italy.}

\bibitem[\protect\BCAY{Vallati, Hutter, Chrpa,\ \BBA\ McCluskey}{Vallati
  et~al.}{2015}]{vallati2015effective}
Vallati, M., Hutter, F., Chrpa, L., \BBA\ McCluskey, T.~L. \BBOP2015\BBCP.
\newblock \BBOQ On the effective configuration of planning domain models\BBCQ\
\newblock In {\Bem International Joint Conference on Artificial Intelligence
  (IJCAI)}. AAAI press.

\bibitem[\protect\BCAY{Van~Eijk}{Van~Eijk}{2000}]{van2000sequential}
Van~Eijk, C. \BBOP2000\BBCP.
\newblock \BBOQ Sequential equivalence checking based on structural
  similarities\BBCQ\
\newblock {\Bem IEEE Transactions on Computer-Aided Design of Integrated
  Circuits and Systems}, {\Bem 19\/}(7), 814--819.

\bibitem[\protect\BCAY{Zhuo\ \BBA\ Kambhampati}{Zhuo\ \BBA\
  Kambhampati}{2013}]{zhuo2013action}
Zhuo, H.~H.\BBACOMMA\  \BBA\ Kambhampati, S. \BBOP2013\BBCP.
\newblock \BBOQ Action-model acquisition from noisy plan traces\BBCQ\
\newblock In {\Bem Twenty-Third International Joint Conference on Artificial
  Intelligence}.

\end{thebibliography}
\bibliographystyle{theapa}

\end{document}